\documentclass{article}
\usepackage{arxiv}
\usepackage[utf8]{inputenc}
\usepackage[T1]{fontenc}
\usepackage{times}
\usepackage{xspace}
\usepackage[dvipsnames]{xcolor}
\usepackage{graphicx}
\usepackage{amsmath}
\usepackage{amssymb}
\usepackage{booktabs}
\usepackage[numbers,sort&compress]{natbib}
\usepackage[format=plain,labelformat=simple,labelsep=period,font=small,compatibility=false]{caption}
\usepackage[font=footnotesize,skip=3pt,subrefformat=parens]{subcaption}
\usepackage[hyphens]{url}
\usepackage[shortlabels,inline]{enumitem}
\usepackage{array}
\usepackage{longtable}
\usepackage{placeins}
\usepackage{algpseudocode}
\usepackage{pifont}
\usepackage{multirow}
\usepackage{colortbl}
\usepackage{microtype}
\usepackage{authblk}
\AtBeginDocument{\newgeometry{textheight=9in,textwidth=6.875in,top=1in,headheight=14pt,headsep=25pt,footskip=30pt}}
\definecolor{oursbg}{HTML}{EEF3FA}
\newcommand{\yes}{\ding{51}}
\newcommand{\no}{\textcolor{gray}{\ding{55}}}
\makeatletter
\DeclareRobustCommand\onedot{\futurelet\@let@token\@onedot}
\def\@onedot{\ifx\@let@token.\else.\null\fi\xspace}
\def\etal{\emph{et al}\onedot}
\makeatother
\definecolor{linkblue}{rgb}{0.21,0.49,0.74}
\usepackage[breaklinks,colorlinks,allcolors=linkblue]{hyperref}
\renewcommand{\headeright}{Preprint}
\renewcommand{\undertitle}{Preprint}
\renewcommand{\shorttitle}{What Matters for Cross-Modal Alignment}

\title{What Matters for Cross-Modal Alignment}
\author[1]{Yipeng Sun\thanks{Corresponding author: \texttt{yipeng.sun@fau.de}}}
\author[1]{Jiajun Wang}
\author[1]{Siyuan Mei}
\author[2]{Yan Xia}
\author[1]{Siming Bayer}
\author[1]{Vincent Christlein}
\author[1]{Andreas Maier}
\affil[1]{Pattern Recognition Lab, Friedrich-Alexander-Universit\"at Erlangen-N\"urnberg, Erlangen, Germany}
\affil[2]{Department of Orthodontics and Orofacial Orthopedics, Friedrich-Alexander-Universit\"at Erlangen-N\"urnberg, Erlangen, Germany}
\date{}
\begin{document}
\maketitle
\begin{abstract}
The Platonic Representation Hypothesis posits that models trained on different modalities converge toward a shared representation. Related work aligns frozen image and text encoders on natural images without paired data. We test zero-pair alignment on five datasets across four modalities: natural images, environmental audio, chest radiographs, and 3D computed tomography (CT). It succeeds only on COCO and fails on the other four. We analyze the failures and find that alignment must first locate the subspace where the modalities co-vary and then orient the mapping within it. Consistent with previous studies, medical imaging is the hardest case: its shared subspace carries the smallest fraction of the variance. When this fraction is small, alignment needs image--text pairs, and random-matrix theory predicts how many pairs are needed. These findings motivate \textit{LightAlign}: a few pairs let a ridge regression between two frozen encoders locate and orient the mapping simultaneously, and a classifier fitted only on report text then predicts findings from the mapped images. On 3D CT, \textit{LightAlign} surpasses models trained end to end on image--report pairs without finding-level labels: with only 200 pairs, \textit{LightAlign} outperforms CT-CLIP on CT-RATE (0.754 vs.\ 0.733 AUROC), and with only 100 CT-RATE pairs, it does so on RAD-ChestCT, an out-of-distribution dataset from a different hospital. On CheXpert, it reaches 0.862 mean AUROC with only about 5\,\% as many pairs as models trained on MIMIC-CXR.
\end{abstract}

\begin{center}
\small Code and weights: \href{https://github.com/sypsyp97/lightalign}{github.com/sypsyp97/lightalign}\qquad
Project page: \href{https://sypsyp97.github.io/lightalign/}{sypsyp97.github.io/lightalign}
\end{center}
\begin{figure}[!t]
\centering
\includegraphics[width=0.62\textwidth]{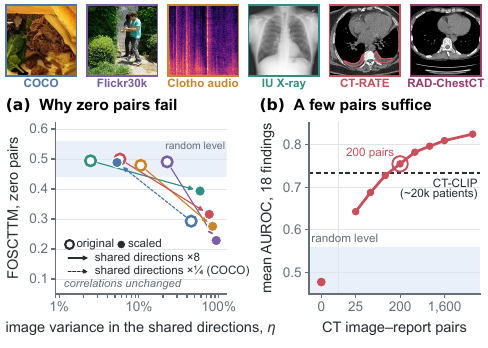}%
{\phantomsubcaption\label{fig:teaser:a}}{\phantomsubcaption\label{fig:teaser:b}}%
\caption{A large variance share is necessary for zero-pair alignment, and 200 pairs outperform an end-to-end model. Top: one held-out example per dataset, framed in its marker color (CT: one axial slice). \textbf{(a)}~Each marker is one dataset (open: before, filled: after scaling the shared directions eight-fold at unchanged canonical correlations); horizontal: mean fraction $\eta$ of image variance in the shared directions; vertical: FOSCTTM, lower is better (shaded: random level). Scaling the shared directions aligns all four failing datasets (solid arrows, after sign correction; each combination in Fig.~\ref{fig:locate}); shrinking them on COCO to a quarter breaks the alignment (dashed arrow). \textbf{(b)}~Mean AUROC over 18 findings on the 3,002 test scans vs.\ pair count (red curve; ring: 200 pairs, both encoders frozen); dashed line: CT-CLIP, trained end to end on the image--report pairs of about 20,000 patients without finding labels.}
\label{fig:teaser}
\end{figure}

\begin{figure*}[t]
\centering
\includegraphics[width=\textwidth]{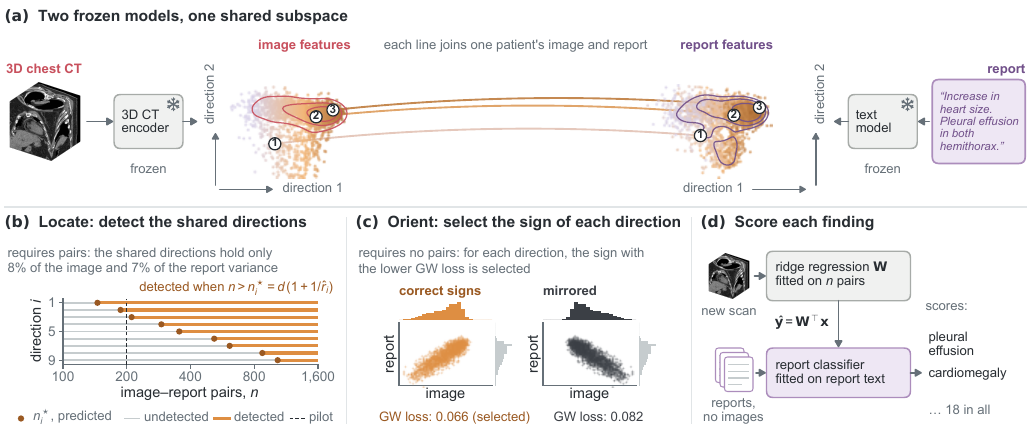}%
{\phantomsubcaption\label{fig:overview:a}}{\phantomsubcaption\label{fig:overview:b}}{\phantomsubcaption\label{fig:overview:c}}{\phantomsubcaption\label{fig:overview:d}}%
\caption{Locate the shared subspace with paired data, orient it without paired data, predict findings with a report classifier.
\textbf{(a)}~On 2,025 validation patients of CT-RATE, the image and report features match on the first two shared directions (correlations 0.87 and 0.73). Color: image score on direction~1; numbered lines join image and report of three patients.
\textbf{(b)}~Shared directions hold 8\,\% of image and 7\,\% of report variance. Rows: directions~1 to 9; horizontal: pairs~$n$. Orange: detected; gray: undetected; dashed line: 200-pair pilot. Direction~$i$ is detected when $n>n^\star_i$ (Eq.~\eqref{eq:law}); dots: $n^\star_i$ predicted from $\hat r_i$ estimated on 2,024 patients.
\textbf{(c)}~Direction~1 and its mirror (image side negated). The mirror has the higher Gromov--Wasserstein (GW) loss (0.082 vs.\ 0.066, disjoint halves of the validation patients). Orange: correct signs; black: mirror.
\textbf{(d)}~Ridge regression maps a new scan to the text feature space, where a report classifier predicts each finding (Algorithm~2 in Fig.~\ref{alg:few}). All panels: VoxelFM block~11 and Qwen3.5-9B on CT-RATE.}
\label{fig:overview}
\end{figure*}

\section{Introduction}
\label{sec:intro}

The Platonic Representation Hypothesis posits that independently trained unimodal models converge to a shared geometric structure~\cite{arXiv:2405.07987}. Schnaus~\etal~\cite{arXiv:2610.09411} show that Wasserstein Procrustes~\cite{grave2019wp} coarsely aligns a frozen image encoder with a frozen language model using zero pairs on natural images, but fails on medical images. We test Wasserstein Procrustes on five datasets across four modalities: natural images (COCO~\cite{coco2014}, Flickr30k~\cite{flickr30k2014}), environmental audio (Clotho~\cite{clotho2020}), 2D chest radiographs (IU X-ray~\cite{iuxray2016}), and 3D computed tomography (CT) from CT-RATE~\cite{arXiv:2403.17834}. In our experiments, zero-pair alignment fails on four of the five datasets and succeeds only on COCO (Fig.~\ref{fig:teaser:a}). We measure alignment by FOSCTTM~\cite{liu2019mmdma}, the fraction of samples closer than the true match (0 is perfect, 0.5 is random). On CT, all 56 tested encoder--text combinations stay at the random level (0.440--0.536), against 0.012 on COCO (Appendix~\ref{app:locate}).

The failure is surprising because CT features carry a strong signal: the top canonical correlation between the paired image and report features of validation patients is 0.87 (Fig.~\ref{fig:overview:a}). Similarity scores such as centered kernel alignment (CKA)~\cite{kornblith2019cka,arXiv:2610.09411} can predict whether zero-pair alignment succeeds, but not why it fails, how to fix it, which encoder to choose, or how many image--text pairs to collect.

To answer these questions, we analyze the failures and find that alignment needs two conditions in sequence. The first is \emph{localization} of the shared subspace (Fig.~\ref{fig:overview:b}), the directions along which the two modalities co-vary. Even when canonical correlations are high, localization depends on the variance share $\eta$, the fraction of image feature variance in the shared subspace. Distribution-matching objectives weight each direction by its variance, so when $\eta$ is small, private variation dominates the objective and zero-pair alignment fails. The variance share is smallest in the medical datasets: only 0.5--10.5\,\% on CT and 2--3.5\,\% on chest radiographs, consistent with the low agreement between medical image encoders~\cite{arXiv:2607.20274}. To test this causally, we scale the shared directions eight-fold, which raises $\eta$ at essentially unchanged canonical correlations and turns random matching into alignment on every failing dataset after sign correction. Shrinking the same directions on COCO breaks its alignment (Fig.~\ref{fig:teaser:a}).\looseness=-1

However, a large variance share is necessary but not sufficient. The second condition is \emph{orientation} of each located direction (Fig.~\ref{fig:overview:c}). After rescaling the shared directions, the zero-pair method finds the subspace on every seed but often returns a mirrored map on all four failing datasets. Because real features are skewed, the mirrored map has the higher Gromov--Wasserstein (GW)~\cite{memoli2011gw,peyre2016gw} loss on every seed. The GW loss selects the sign pattern that corrects every mirrored map on CT and chest radiographs without paired data, and on CT it outperforms oracle signs from paired correlations. Without skewness, sign correction largely fails.

When $\eta$ is small, the modalities may share strongly correlated directions that zero-pair methods miss, so we ask how many pairs are needed to recover them. With few pairs, sampling noise pushes spurious sample canonical correlations up to a bulk edge, masking weak shared directions. Random-matrix theory describes this high-dimensional regime: a direction with squared canonical correlation $r$ in $d$ principal components becomes detectable only above the threshold $n^\star\!=\!d(1{+}1/r)$~\cite{bao2019canonical,ma2023sample} (Fig.~\ref{fig:overview:b}). We find that it also holds for real frozen features, which are strongly non-Gaussian. It predicts the squared sample canonical correlations with a median error of 0.016 over 2,550 CT settings and of 0.008--0.030 on the four other datasets, seven further text models, and BIMCV-R CT.\looseness=-1

Based on these findings, we introduce \textit{LightAlign}, a ridge regression that maps features of a frozen image encoder to those of a frozen text model. With paired data, the regression locates and orients the map in one step. A logistic regression fitted on report text alone predicts each finding. One forward pass through each frozen encoder suffices, and the whole process takes about a minute on four CPU cores (Appendix~\ref{app:impl}). We evaluate it on chest radiographs (CheXpert) and 3D CT (CT-RATE), the two hardest cases: their shared subspaces have the smallest variance share (Table~\ref{tab:domains}) and paired data are scarce. To our knowledge, we are the first to align 3D CT volumes with clinical text from frozen encoders with few pairs. On CT-RATE, 200 pairs exceed the area under the receiver operating characteristic curve (AUROC) of CT-CLIP trained end to end on about 20,000 patients (0.754 vs.\ 0.733; Fig.~\ref{fig:teaser:b}), and all 5,994 training pairs exceed COLIPRI (0.824 vs.\ 0.816). On the out-of-distribution (OOD) hospital RAD-ChestCT, we use none of its data and apply the CT-RATE map and classifier unchanged. On RAD-ChestCT, 100 pairs exceed CT-CLIP and all pairs match HLIP. On CheXpert, \textit{LightAlign} reaches 0.862 mean AUROC with about 5\,\% as many pairs as models trained on MIMIC-CXR.

Our main contributions are:
\begin{enumerate}
\item We show causally on five datasets in four modalities that zero-pair alignment needs a large variance share $\eta$ of the shared subspace. A large share is not sufficient, because the map must also be oriented. Real features are skewed, so the GW loss selects the correct signs on CT and chest radiographs without paired data, and on CT it even outperforms oracle signs from paired correlations. \item We find that a small $\eta$ means that two encoders cannot be aligned with zero pairs. The detection threshold $n^\star\!=\!d(1{+}1/r)$ of random-matrix theory holds in every modality, and a 200-pair pilot uses it to select the best encoder combination and the number of pairs needed. \item We propose \textit{LightAlign}, to our knowledge the first method that can align 3D CT volumes with clinical text from frozen encoders with few pairs. With only 200 pairs it outperforms CT-CLIP trained end to end on image--report pairs, and with all pairs it outperforms COLIPRI. On CheXpert it reaches 0.862 mean AUROC with about 5\,\% as many pairs as models trained on MIMIC-CXR.
\end{enumerate}

\begin{table*}[t]
\centering
\caption{Multi-abnormality classification on 3D chest CT. CT-RATE: test set, i.e., the official validation set (3,002 scans, 18 findings), mean AUROC, weighted F1, and accuracy at the receiver operating characteristic (ROC) corner threshold of CT-CLIP~\cite{arXiv:2403.17834}. RAD-ChestCT: OOD hospital, 3,630 volumes, 16 findings; ours applies the CT-RATE map and classifier unchanged. Labels: finding-label supervision; --: no text encoder, not reported, or not evaluated. Published models: numbers from the cited papers; SOTAlign and Hadgi~\etal: fitted by us. Ours: frozen VoxelFM block~11 + Qwen3.5-9B middle layer, selected on validation patients; ridge regression on $n$ pairs, report classifier, five seeds; 0 pairs uses the Wasserstein Procrustes method~\cite{arXiv:2610.09411}. COLIPRI (CRM): contrastive + report generation + masked autoencoder; (CM): without report generation. NLST: National Lung Screening Trial. VocabFine: CT-CLIP fine-tuned on 18 labels. Other few-pair maps: Table~\ref{tab:supp_fewpair}. \textbf{Bold}: best without image labels; \underline{underline}: second best.}
\label{tab:main}
\setlength{\tabcolsep}{2.5pt}
\renewcommand{\arraystretch}{0.93}
\small
\begin{tabular}{lccclcccc}
\toprule
& \multicolumn{3}{c}{Trained (\yes) or not (\no)} & & \multicolumn{3}{c}{CT-RATE} & RAD-ChestCT \\
\cmidrule(lr){2-4}\cmidrule(lr){6-8}\cmidrule(l){9-9}
Method & Image enc. & Text enc. & Labels & Training data & AUROC$\uparrow$ & F1$\uparrow$ & Acc.$\uparrow$ & AUROC$\uparrow$ \\
\midrule
\multicolumn{9}{l}{\emph{Trained end to end on image--report pairs, no finding labels}}\\
CT-CLIP~\cite{arXiv:2403.17834} & \yes & \yes & \no & CT-RATE, 20k patients & 0.733$^{a}$ & 0.707$^{a}$ & 0.667$^{a}$ & 0.629 \\
Merlin~\cite{arXiv:2406.06512} & \yes & \yes & \no$^{f}$ & abdominal CT + EHR codes & 0.662$^{b}$ & -- & -- & 0.644$^{c}$ \\
Jolia~\cite{arXiv:2606.24570} & \yes & \no & \no & CT-RATE, INSPECT, Merlin$^{m}$ & 0.699$^{b}$ & -- & -- & -- \\
fVLM~\cite{arXiv:2501.14548} & \yes & \yes & \no & CT-RATE + organ masks & 0.778$^{d}$ & 0.751$^{d}$ & 0.718$^{d}$ & 0.680 \\
HLIP~\cite{arXiv:2505.21862} & \yes & \yes & \no & CT-RATE & 0.777$^{d}$ & 0.747$^{d}$ & 0.714$^{d}$ & 0.723$^{d}$ \\
COLIPRI (CRM)~\cite{arXiv:2510.15042} & \yes & \yes & \no & CT-RATE + NLST images & \underline{0.816}$^{b}$ & -- & -- & \textbf{0.732}$^{e}$ \\
\midrule
\multicolumn{9}{l}{\emph{Trained with finding labels}}\\
CT-CLIP + head~\cite{arXiv:2604.04133} & \yes$^{k}$ & -- & \yes & CT-RATE labels & 0.574$^{g}$ & -- & -- & -- \\
M3D-LaMed + head~\cite{arXiv:2404.00578,arXiv:2604.04133} & \no & -- & \yes & CT-RATE labels & 0.740$^{g}$ & -- & -- & -- \\
VocabFine~\cite{arXiv:2403.17834,arXiv:2609.27511} & \yes & \yes & \yes & CT-RATE labels & 0.758 & 0.738$^{l}$ & 0.705$^{l}$ & 0.650 \\
RadFM + head~\cite{arXiv:2308.02463,arXiv:2604.04133} & \no & -- & \yes & CT-RATE labels & 0.759$^{g}$ & -- & -- & -- \\
Merlin + head~\cite{arXiv:2406.06512,arXiv:2604.04133} & \no & -- & \yes & CT-RATE labels & 0.798$^{g}$ & -- & -- & -- \\
CT-SSG~\cite{dipiazza2026ctssg} & \yes & -- & \yes & CT-RATE labels & 0.836$^{h}$ & -- & -- & 0.746$^{h}$ \\
Pillar-0 + head~\cite{arXiv:2511.17803,arXiv:2604.10437} & \no & -- & \yes & CT-RATE labels & 0.861$^{g}$ & -- & -- & 0.796$^{n}$ \\
COLIPRI (CM) + head~\cite{arXiv:2510.15042} & \yes$^{k}$ & -- & \yes & CT-RATE labels & 0.862 & -- & -- & -- \\
VoxelFM + head~\cite{arXiv:2604.04133} & \no & -- & \yes & CT-RATE labels & 0.870$^{g}$ & -- & -- & -- \\
NV-Reason-CT~\cite{arXiv:2609.27511} & \yes & \yes & \yes & CT-RATE, NIH, CancerVerse$^{i}$ & 0.871 & -- & -- & 0.796 \\
\midrule
\multicolumn{9}{l}{\emph{Same frozen VoxelFM and Qwen3.5-9B, other few-pair maps, logistic regression fitted on report text}}\\
SOTAlign~\cite{arXiv:2602.23353}, 100 pairs & \no & \no & \no$^{j}$ & CT-RATE & 0.644 & 0.659 & 0.611 & 0.573 \\
SOTAlign~\cite{arXiv:2602.23353}, 400 pairs & \no & \no & \no$^{j}$ & CT-RATE & 0.670 & 0.678 & 0.633 & 0.597 \\
SOTAlign~\cite{arXiv:2602.23353}, 1,600 pairs & \no & \no & \no$^{j}$ & CT-RATE & 0.702 & 0.696 & 0.655 & 0.622 \\
Hadgi~\etal~\cite{arXiv:2503.05283}, 100 pairs & \no & \no & \no$^{j}$ & CT-RATE & 0.685 & 0.683 & 0.638 & 0.603 \\
Hadgi~\etal~\cite{arXiv:2503.05283}, 400 pairs & \no & \no & \no$^{j}$ & CT-RATE & 0.752 & 0.731 & 0.695 & 0.652 \\
Hadgi~\etal~\cite{arXiv:2503.05283}, 1,600 pairs & \no & \no & \no$^{j}$ & CT-RATE & 0.790 & 0.757 & 0.725 & 0.686 \\
\midrule
\rowcolor{oursbg}\multicolumn{9}{l}{\emph{Ours: frozen VoxelFM + frozen Qwen3.5-9B, ridge regression on $n$ pairs, a logistic regression fitted on report text}}\\
\rowcolor{oursbg} Ours, 0 pairs &  &  &  &  & 0.478 & 0.547 & 0.487 & 0.512 \\
\rowcolor{oursbg} Ours, 50 pairs &  &  &  &  & 0.687 & 0.685 & 0.642 & 0.610 \\
\rowcolor{oursbg} Ours, 100 pairs &  &  &  &  & 0.727 & 0.709 & 0.670 & 0.636 \\
\rowcolor{oursbg} Ours, 200 pairs &  &  &  &  & 0.754 & 0.726 & 0.690 & 0.658 \\
\rowcolor{oursbg} Ours, 400 pairs &  &  &  &  & 0.782 & 0.750 & 0.718 & 0.674 \\
\rowcolor{oursbg} Ours, 1,600 pairs &  &  &  &  & 0.809 & \underline{0.769} & \underline{0.740} & 0.705 \\
\rowcolor{oursbg} Ours, 5,994 pairs & \multirow{-7}{*}{\no} & \multirow{-7}{*}{\no} & \multirow{-7}{*}{\no$^{j}$} & \multirow{-7}{*}{CT-RATE} & \textbf{0.824} & \textbf{0.777} & \textbf{0.750} & \underline{0.724} \\
\bottomrule
\end{tabular}
\par\vspace{2pt}\begin{minipage}{\textwidth}\footnotesize $^{a}$reproduction of the released checkpoint (0.7328 in~\cite{arXiv:2609.27511}); $^{b}$re-evaluated on the same scans in~\cite{arXiv:2609.27511}; $^{c}$from~\cite{arXiv:2501.14548} (also 0.644 in~\cite{arXiv:2505.21862}); $^{d}$16 findings (14 for HLIP on RAD-ChestCT), unit not stated, accuracy is balanced; $^{e}$label subset, 0.730 on all; $^{f}$electronic health record codes; $^{g}$3,039-case validation set from~\cite{arXiv:2609.27511}; $^{h}$five-fold cross-validation, not official validation set; $^{i}$70,111 volumes, supervised fine-tuning then reinforcement learning on 59 chest and abdominal findings, text side a language model; $^{j}$report-derived labels fit only the report classifier; no image is labeled; $^{k}$image encoder trained by its authors on CT-RATE scans and reports, then frozen; $^{l}$from~\cite{arXiv:2403.17834}, 1,564 validation scans; $^{m}$Merlin-Abd-CT dataset; $^{n}$from Table~7 of~\cite{arXiv:2609.27511}, frozen Pillar-0 features.\end{minipage}
\end{table*}

\section{Related Work}

\noindent\textbf{Representation convergence.}~Following the Platonic Representation Hypothesis~\cite{arXiv:2405.07987}, recent work studies linear equivalence~\cite{arXiv:2507.01098,arXiv:2605.28870,arXiv:2610.07168}, which structure converges~\cite{arXiv:2602.14486,arXiv:2609.27252,arXiv:2610.20286}, capacity~\cite{lobashev2025infogeo}, concept transfer~\cite{arXiv:2610.36269}, and frozen-encoder comparison in medical imaging~\cite{arXiv:2607.20274,arXiv:2607.22771}. These studies ask whether encoders converge. We show that aligning them without paired data also needs a large variance share.
Balestriero and LeCun~\cite{arXiv:2610.39302} prove that joint-embedding predictive architectures learn the same embedding up to a rotation that the first two moments of unpaired data cannot identify, and independently we show on frozen encoders in four modalities that skewness separates the correct map from its mirror and that the map can be oriented without paired data.

\noindent\textbf{Zero- and few-pair alignment.}~Motivated by this convergence, recent work aligns embedding spaces with few or no pairs. Shared cross-modal components are identifiable under distributional assumptions~\cite{arXiv:2409.19422} and recoverable by spectral methods~\cite{arXiv:2505.21524} or quadratic assignment~\cite{arXiv:2610.09530}. Unsupervised alignment also succeeds on natural images~\cite{arXiv:2610.09411} and across text spaces~\cite{arXiv:2505.12540}. RankAlign matches class-level prototypes without paired examples~\cite{arXiv:2610.31145}, and PANDA uses class prototypes across partial- and zero-pair regimes~\cite{arXiv:2608.25970}. Skewness fixes the sign of linear translation axes~\cite{arXiv:2007.12568}. When zero pairs fail, few-pair methods use optimal transport with paired anchors~\cite{arXiv:2610.12781,arXiv:2602.23353}, lightweight modules~\cite{arXiv:2210.01738,arXiv:2312.10144,arXiv:2409.19425,arXiv:2412.04616,arXiv:2506.16895}, or an affine canonical correlation analysis map on 30,000 pairs~\cite{arXiv:2503.05283}. These methods show where alignment works, but not why it fails on real frozen features or how many pairs it needs. We answer both questions for frozen encoders.

\noindent\textbf{Canonical correlations in high dimensions.}~Counting image--text pairs reduces to counting sample canonical correlations in high dimensions. Random-matrix theory characterizes their phase transition for Gaussian~\cite{bao2019canonical} and non-Gaussian vectors~\cite{ma2023sample}, the distribution of sample canonical directions~\cite{arXiv:2606.09153}, and when multimodal learning recovers shared factors~\cite{arXiv:2407.03522,arXiv:2606.11190}. These results assume idealized random vectors. We find that the detection threshold also holds for real frozen features and use it to count image--text pairs.

\noindent\textbf{Vision--language models.}~CLIP~\cite{radford2021clip} trains on 400 million image--text pairs with a contrastive loss and SigLIP~\cite{zhai2023siglip} with a sigmoid loss on each pair. Our questions matter most in medical imaging, where paired data are scarce. CT-CLIP~\cite{arXiv:2403.17834}, Merlin~\cite{arXiv:2406.06512}, fVLM~\cite{arXiv:2501.14548}, HLIP~\cite{arXiv:2505.21862}, COLIPRI~\cite{arXiv:2510.15042}, Jolia~\cite{arXiv:2606.24570}, and NV-Reason-CT~\cite{arXiv:2609.27511} train volumetric encoders or projectors on thousands of paired scans and reports. SpeechCT-CLIP~\cite{arXiv:2510.02322} distills one into speech. Three radiograph methods avoid image--report pairs~\cite{arXiv:2111.04318,arXiv:2403.13444,arXiv:2407.03919} but need knowledge graphs or disease labels and train their encoders. Benchmarks evaluate frozen CT encoders with supervised heads~\cite{arXiv:2607.20993,arXiv:2608.05960,arXiv:2608.25948}, and text-trained classifiers transfer to radiographs~\cite{arXiv:2409.02729,arXiv:2602.03098}. Linear or affine maps~\cite{arXiv:2209.15162,arXiv:2503.04478} and a Transformer~\cite{arXiv:2606.17213} bridge frozen encoders and language models, but none predicts how many pairs suffice or explains why fewer fail. \textit{LightAlign} keeps both encoders frozen, needs no image labels, and outperforms end-to-end 3D CT models trained on image--report pairs without finding labels (Table~\ref{tab:main}).\looseness=-1

\section{Method}
\label{sec:method}

This section diagnoses the failure in three steps and derives \textit{LightAlign} from them: locate the shared subspace, orient its directions, and predict how many pairs alignment needs.

\subsection{Problem formulation}
A frozen encoder maps a photograph, audio clip, chest radiograph, or 3D CT scan to $\mathbf{x}\in\mathbb{R}^{p}$, while a frozen text model maps the caption or report to $\mathbf{y}\in\mathbb{R}^{q}$. For brevity, we call the non-text side the image side. For a language model, we average the token states at a middle layer. For a text embedding model, we use the output embedding. We use a middle layer because the last layers specialize in next-token prediction whereas middle layers encode more general features~\cite{arXiv:2502.02013}, and our layer ablation (Sec.~\ref{sec:ablation}) confirms this choice. We normalize each feature vector to unit norm, subtract the unpaired training-set mean, and renormalize it. We seek a linear map $\mathbf{W}\in\mathbb{R}^{p\times q}$ such that $\mathbf{W}^{\top}\mathbf{x}$ approximates $\mathbf{y}$. With $n$ image--text pairs, we can fit $\mathbf{W}$ directly. With no pairs, we can only match the two feature distributions.

\subsection{Step 1: locate the shared subspace}\label{sec:locate}
We find that zero-pair methods can locate the shared subspace only when it has enough variance. We first project each modality onto its leading $d=64$ principal components, computed without paired data. A shared direction $i$ is a pair of unit vectors, one per modality, whose projections have population squared canonical correlation $r_i$ (Fig.~\ref{fig:overview:a}).

We estimate these directions by canonical correlation analysis (CCA) on 2,024 training patients (Sec.~\ref{sec:setup}).

Let $\mathbf{U}_x\in\mathbb{R}^{p\times k}$ and $\mathbf{U}_y\in\mathbb{R}^{q\times k}$ be orthonormal bases for the $k$ canonical directions that exceed both a permutation null and the noise bulk edge (Sec.~\ref{sec:count}). The variance share is $\eta=\lVert(\mathbf{X}-\bar{\mathbf{X}})\mathbf{U}_x\rVert_F^2/\lVert\mathbf{X}-\bar{\mathbf{X}}\rVert_F^2$, where $\lVert\cdot\rVert_F$ is the Frobenius norm, the rows of $\mathbf{X}$ are the training image features, and $\bar{\mathbf{X}}$ repeats their mean.\looseness=-1

A zero-pair method compares distances between items within each modality. For independent items $\mathbf{x},\mathbf{x}'$ with covariance $\boldsymbol{\Sigma}_x$, the expected squared distance $\mathbb{E}\lVert\mathbf{x}-\mathbf{x}'\rVert^2=2\,\mathrm{tr}\,\boldsymbol{\Sigma}_x$ is twice the sum of the variances of all directions, so the shared subspace contributes a fraction $\eta$ of this distance, regardless of its correlation. When $\eta$ is small, private directions dominate the geometry that the method matches. To test this effect causally, we rescale the shared directions as \begin{equation} \tilde{\mathbf{x}}=\mathbf{x}+(\beta-1)\,\mathbf{U}_x\mathbf{U}_x^{\top}(\mathbf{x}-\bar{\mathbf{x}}),\qquad \beta>0, \label{eq:amp} \end{equation} where $\tilde{\mathbf{x}}$ is the rescaled feature, $\bar{\mathbf{x}}$ the modality mean, and $\beta$ the scale factor, a power of two from 0.25 to 32. Rescaling changes $\eta$ but not the $r_i$, since canonical correlations are invariant under invertible linear maps. As a control, we rescale $k$ random directions with the same share.\looseness=-1

\subsection{Step 2: orient each shared direction}\label{sec:orient}
Once the subspace is located, the sign of each direction is still unknown. Without paired data, an objective that uses only the covariance of each modality cannot determine this sign, because reflecting the mapped image features along a principal axis leaves their covariance unchanged. A map that locates the subspace can therefore be a \emph{mirror} of the correct one. We find that real frozen features are skewed (Sec.~\ref{sec:orientres}). The third moment $\mathbb{E}[z^3]$ of the projection $z$ of a centered feature onto a shared direction changes sign when the direction is negated, so the mirrored map has the higher Gromov--Wasserstein (GW) loss (Fig.~\ref{fig:overview:c}). Without paired data, the canonical directions are unknown. Let the columns of $\mathbf{Q}$ be the top $t=10$ right singular vectors of the centered mapped image features, $\mathbf{s}\in\{\pm1\}^t$ a sign pattern, $\mathbf{Q}_{\mathbf{s}}$ the columns with $s_j=-1$, $\mathbf{R}_{\mathbf{s}}=\mathbf{I}-2\mathbf{Q}_{\mathbf{s}}\mathbf{Q}_{\mathbf{s}}^{\top}$ the reflection along them, and $\pi_{\mathbf{s}}$ the one-to-one matching of the $M$ reflected image features to $M$ reports by the Hungarian algorithm. With $C^X_{ab}$ and $C^Y_{ab}$ the cosine distances between items $a$ and $b$ within each modality, \emph{GW sign selection} keeps the sign pattern with the lowest GW loss,
\begin{equation}
\mathbf{s}^\star=\arg\min_{\mathbf{s}\in\{\pm1\}^t}\frac{1}{M^2}\sum_{a,b=1}^{M}\big(C^X_{ab}-C^Y_{\pi_{\mathbf{s}}(a)\,\pi_{\mathbf{s}}(b)}\big)^2.
\label{eq:sign}
\end{equation} The Wasserstein Procrustes method of Schnaus~\etal~\cite{arXiv:2610.09411} locates the shared subspace, and GW sign selection orients it (Algorithm~1 in Fig.~\ref{alg:zero}). For the zero-pair baseline, we run Wasserstein Procrustes alone with default hyperparameters on two disjoint halves of the training set.\looseness=-1

\subsection{Step 3: predict how many pairs are needed}\label{sec:count}
When $\eta$ is small, localization needs paired data, which remove the dependence on $\eta$. With few pairs, sampling noise lifts the squared sample canonical correlations up to a bulk edge and hides any weak shared direction below it (Fig.~\ref{fig:overview:b}). For projected dimension $d$ on both sides and $n$ pairs with $c=d/n<1/2$, random-matrix theory~\cite{bao2019canonical} shows that as $d$ and $n$ grow with $c$ fixed, direction $i$ separates from the bulk once $r_i>c/(1-c)$, i.e.
\begin{equation}
n > n^\star_i = d\left(1+\frac{1}{r_i}\right).
\label{eq:law}
\end{equation}
Above this threshold, the squared sample canonical correlation $\lambda_i$ converges to
\begin{equation}
\lambda_i \to \gamma(r_i)= r_i\left(1-c+\frac{c}{r_i}\right)^{2},
\label{eq:outlier}
\end{equation}
whereas below $n^\star_i$ it stays at the bulk edge $\lambda_+=4c(1-c)$. Bao~\etal~\cite{bao2019canonical} prove Eqs.~\eqref{eq:law}--\eqref{eq:outlier} for Gaussian data. Ma and Yang~\cite{ma2023sample} extend Eq.~\eqref{eq:law} to non-Gaussian data with independent entries or factor structure and finite fourth moment.

In Algorithm~2 (Fig.~\ref{alg:few}), a pilot of $n_0$ pairs ($n_0=200$ in our experiments) tests each $\lambda_i$ against the bulk edge $\lambda_+$ and a permutation null of the top canonical correlation (100 permutations, $p\le0.05$). For each detected direction, we invert Eq.~\eqref{eq:outlier} to get $\hat r_i=\gamma^{-1}(\lambda_i)$, and Eq.~\eqref{eq:law} gives $n^\star_i$. We rank encoders by $S(n_0)=\sum_{i:\,n_0>n^\star_i}\hat r_i$, choose the one with the highest $S$, and fit its map on the pilot pairs.

\subsection{\texorpdfstring{\textit{LightAlign}}{LightAlign}: image-to-text ridge regression}\label{sec:lightalign}
Let the rows of $\mathbf{X}_n\in\mathbb{R}^{n\times p}$ and $\mathbf{Y}_n\in\mathbb{R}^{n\times q}$ be the paired image and text features. The cross-covariance $\mathbf{X}_n^{\top}\mathbf{Y}_n$ locates and orients the shared directions, so the first two steps collapse into one ridge regression $\mathbf{W}=(\mathbf{X}_n^{\top}\mathbf{X}_n+\alpha\mathbf{I})^{-1}\mathbf{X}_n^{\top}\mathbf{Y}_n$ at full dimension. We select $\alpha\in\{10^{-3},10^{-2.5},\dots,10^{3}\}$ by leave-one-out cross-validation.

A test scan $\mathbf{x}$ is mapped to $\hat{\mathbf{y}}=\mathbf{W}^{\top}\mathbf{x}$ (Fig.~\ref{fig:overview:d}). We score each finding $\ell$ in two ways. The \emph{zero-shot prompt} embeds ``\texttt{IMPRESSION: <finding> is present.}'' and ``\texttt{IMPRESSION: No <finding>.}'' with the text model (embeddings $\mathbf{e}^{+}_\ell$ and $\mathbf{e}^{-}_\ell$) and scores $g^{\mathrm{P}}_\ell=\hat{\mathbf{y}}^{\top}(\mathbf{e}^{+}_\ell-\mathbf{e}^{-}_\ell)/\lVert\hat{\mathbf{y}}\rVert$. The \emph{report classifier} fits one logistic regression per finding on the frozen features of unpaired training reports with report-derived labels~\cite{arXiv:2403.17834} and scores the logit $g^{\mathrm{R}}_\ell=\mathbf{w}_\ell^{\top}\hat{\mathbf{y}}+b_\ell$ without using any image labels.\looseness=-1

\begin{figure*}[t]
\centering
\includegraphics[width=\textwidth]{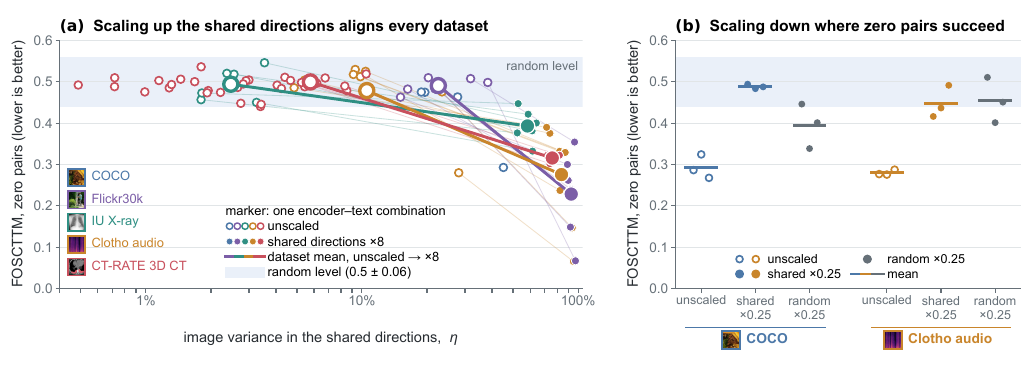}%
{\phantomsubcaption\label{fig:locate:a}}{\phantomsubcaption\label{fig:locate:b}}%
\caption{Without paired data, the zero-pair method finds the shared subspace only when it holds a large share of the variance.
\textbf{(a)}~Horizontal: share $\eta$ of image variance in the shared directions (log scale). Vertical: FOSCTTM at zero pairs. Color: dataset. Each encoder--text combination (three-seed mean): open: original; filled: shared directions scaled eight-fold at unchanged canonical correlations (sign correction: oracle signs on Flickr30k, Clotho, IU X-ray; GW sign selection on CT). Large markers: dataset means. Band: random level (0.44--0.56).
\textbf{(b)}~Where zero pairs work, shrinking the shared directions to a quarter breaks alignment on COCO (DINOv2 mean $\times$ MPNet; Clotho: AST $\times$ MPNet); shrinking random directions of equal variance (gray) breaks COCO less (FOSCTTM 0.395 vs.\ 0.488) and Clotho about as much (0.454 vs.\ 0.448). Dots: single seeds; lines: three-seed mean.}
\label{fig:locate}
\end{figure*}

\section{Experiments}
\label{sec:experiments}

\subsection{Setup}\label{sec:setup}
On CT-RATE~\cite{arXiv:2403.17834}, we first evaluate 15 frozen CT encoders (3D, video, 2D) and 13 text models (110M to 32B; Appendix~\ref{app:data}) with paired data and select VoxelFM. We then ablate its block (2 to 12) and the text-model layer (Sec.~\ref{sec:ablation}) and use VoxelFM block 11 with the middle layer of Qwen3.5-9B. No CT encoder was trained on scans paired with reports. Of the 5,994 CT training patients, we use 2,024 to estimate the shared directions, 1,945 for the zero-pair method, and 2,025 for validation. The four other datasets use a 1:1:4 estimation, validation, and training split. We select the configuration on the validation patients and test it once on the 3,002 scans of the official CT-RATE validation set (18 findings). On chest radiographs, we use the official CheXpert test set~\cite{chexpert2019} (500 studies, five findings) and CheXpert Plus pairs~\cite{chexpertplus2024}. On the OOD hospital RAD-ChestCT, we apply the CT-RATE map and classifier unchanged to its 3,630 volumes. We report FOSCTTM on validation pairs and mean AUROC where labels exist, and compare with baselines by paired bootstrap.

\begin{table}[t]
\centering
\caption{Scaling the shared directions aligns all four failing datasets. Combinations: encoder--text combinations of Fig.~\ref{fig:locate:a}, three seeds each. Share $\eta$: fraction of image variance in the shared directions (range). FOSCTTM: dataset mean / best combination. $\times 8$: shared directions scaled eight-fold at fixed correlations, signs corrected (oracle signs from paired test correlations; GW sign selection on CT). CT-RATE $\times 8$ and mirror columns: the three combinations of Fig.~\ref{fig:orient:a} (nine seeds). Mirrored seeds: maps on scaled features whose FOSCTTM improves by more than 0.1 under the best of the $2^{10}$ sign patterns on paired validation data, as mirrored / all seeds. Corrected without paired data: mirrored seeds where GW sign selection (Eq.~\eqref{eq:sign}; variants: Table~\ref{tab:supp_sign_methods}) recovers over half of that improvement. --: not applicable.}
\label{tab:domains}
\setlength{\tabcolsep}{3pt}
\small
{%
\begin{tabular}{@{}lccccc@{}}
\toprule
& \multicolumn{2}{c}{Unscaled} & $\times 8$ & \multicolumn{2}{c}{Mirror} \\
\cmidrule(lr){2-3}\cmidrule(lr){4-4}\cmidrule(l){5-6}
Dataset & Share & FOSCTTM$\downarrow$ & FOSCTTM$\downarrow$ & Mirrored & Corrected \\
(combinations) & $\eta$ & mean/best & mean/best & seeds & w/o paired data \\
\midrule
COCO (4) & 0.19--0.45 & 0.43/0.29 & -- & -- & -- \\
Flickr30k (6) & 0.15--0.38 & 0.49/0.46 & 0.23/0.07 & 3/18 & 0/3 \\
Clotho (8) & 0.05--0.28 & 0.48/0.28 & 0.28/0.07 & 12/24 & 1/12 \\
IU X-ray (6) & 0.02--0.035 & 0.49/0.45 & 0.39/0.33 & 3/18 & 3/3 \\
CT-RATE (31) & 0.005--0.105 & 0.49/0.44 & 0.32/0.30 & 4/9 & 4/4 \\
\bottomrule
\end{tabular}}
\end{table}

\subsection{Localization requires a large variance share}
On CT, all 56 tested encoder--text combinations stay at the random level with the Wasserstein Procrustes method of Schnaus~\etal~\cite{arXiv:2610.09411} (FOSCTTM 0.440--0.536), and mini-vec2vec~\cite{arXiv:2510.02348} fails on every encoder (Table~\ref{tab:supp_zeropair}). The encoders do share information: on the paired validation patients, the similarity between the image and report features is significant for all 15 encoders (permutation test, $p\le 0.005$). Over the whole feature space, however, linear CKA is only 0.01--0.09 on CT against 0.21--0.57 on COCO, because CKA weights directions by variance and the shared directions carry little variance (Table~\ref{tab:supp_loc}).

In unscaled features, the shared directions explain 0.5--10.5\,\% of the image variance on CT (Fig.~\ref{fig:locate:a}, Table~\ref{tab:domains}). Every unscaled CT combination fails without paired data, whereas every dataset aligns once $\eta \ge 0.5$. To test whether $\eta$ causes the failure, we scale the shared directions eight-fold (Eq.~\eqref{eq:amp}), which raises the measured $\eta$ to 0.52--0.97 (0.72--0.82 on CT) while the canonical correlations change by at most 0.059 after unit normalization. Every failing dataset then aligns (Table~\ref{tab:domains}), and on the three CT combinations FOSCTTM falls from 0.50 to 0.32. On Flickr30k and Clotho, single combinations reach 0.07 (Table~\ref{tab:supp_locate_cells}).

Shrinking the shared directions on COCO to a quarter breaks alignment (FOSCTTM 0.293 to 0.488), even though the top canonical correlation changes little (0.94 to 0.93, Fig.~\ref{fig:locate:b}). Shrinking 38 random directions with the same share degrades it much less (0.395). Whitening, independent component analysis (ICA), and ICA on the most skewed components leave CT at the random level and break COCO at 64 or more dimensions. Locating a small shared subspace thus requires paired data (Appendix~\ref{app:locate}, Table~\ref{tab:supp_pairfree}).

\begin{figure}[t]
\centering
\includegraphics[width=0.62\textwidth]{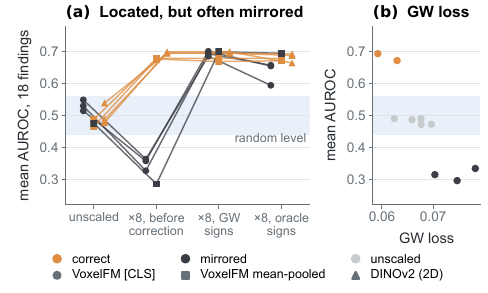}%
{\phantomsubcaption\label{fig:orient:a}}{\phantomsubcaption\label{fig:orient:b}}%
\caption{A located subspace is often mirrored, and the GW loss distinguishes the mirror without paired data.
\textbf{(a)}~Mean AUROC over 18 findings per seed on CT for three encoder--text combinations (circle: VoxelFM [CLS] token, block 11, $\times$ Qwen3.5-9B; square: VoxelFM mean-pooled patch tokens $\times$ Qwen3.5-4B; triangle: DINOv2 on 2D slices $\times$ Qwen3.5-4B): original features, shared directions scaled eight-fold before sign correction, and signs chosen by the GW loss (Eq.~\eqref{eq:sign}) or by oracle signs. Each line is one seed, colored by the orientation of its map before sign correction (orange: correct; black: mirrored). Band: random level.
\textbf{(b)}~GW loss against mean AUROC (VoxelFM patch $\times$ Qwen3.5-4B, five seeds). Gray: original; orange: correct; black: mirrored.}
\label{fig:orient}
\end{figure}

\subsection{Orientation: skewness corrects mirrored maps}\label{sec:orientres}
Scaling the shared directions locates the subspace but does not orient it. For VoxelFM patch $\times$ Qwen3.5-4B, it raises the subspace overlap (the fraction of the paired shared subspace that the map captures) from 0.035--0.044 to 0.691--0.767 on every seed, yet the mean AUROC is still 0.462 because three of five seeds are mirrored (Fig.~\ref{fig:orient:b}; Table~\ref{tab:supp_orient_seeds}). Projected onto the top shared direction, the mapped image and report features correlate at $+0.70$ and $+0.71$ on the correct seeds and at $-0.71$ to $-0.77$ on the mirrored ones.

We find that frozen features are strongly non-Gaussian: 75--100\,\% of the top image principal components (41--95\,\% of the text features) lie outside the Gaussian 99\,\% band of excess kurtosis, and most shared directions are significantly
skewed (Appendix~\ref{app:orient}, Fig.~\ref{fig:supp_extra:a}). On every seed, the mirror has the higher GW loss of
its hard matching (0.070--0.078, against 0.059--0.063 for the correct map; Fig.~\ref{fig:orient:b}).

On the final configuration, all three seeds are mirrored, and GW sign selection raises their mean AUROC from 0.35 to 0.69 (Fig.~\ref{fig:orient:a}), above the oracle signs from the paired test correlations (0.636). On seeds that are already correctly oriented, it preserves the orientation (Table~\ref{tab:coco_signs}).

If skewness drives this correction, negating both centered features on a random half of the items should remove it. This makes the feature distribution symmetric in expectation while leaving the covariance largely unchanged (top canonical correlation 0.880 vs.\ 0.863). Sign correction then drops from 4 of 4 mirrored seeds to 1 of 3, so skewness drives it.

\begin{figure}[t]
\centering
\includegraphics[width=0.62\textwidth]{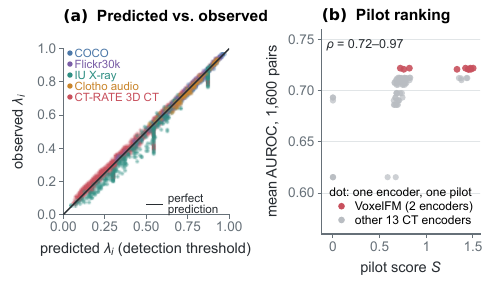}%
{\phantomsubcaption\label{fig:price:a}}{\phantomsubcaption\label{fig:price:b}}%
\caption{The random-matrix detection threshold predicts how many pairs localization requires, and a 200-pair pilot ranks the encoders.
\textbf{(a)}~Top three squared sample canonical correlations $\lambda_i$ of every setting, predicted by Eqs.~\eqref{eq:law}--\eqref{eq:outlier} against observed (CT-RATE: 2,550 settings); line: perfect prediction; color: dataset.
\textbf{(b)}~A 200-pair pilot score (Qwen3.5-4B middle layer, $d=64$) ranks 15 CT encoders by AUROC at 1,600 pairs. Horizontal: pilot score $S$, the summed estimated strengths of the detected shared directions. Each dot: one encoder in one of five independent 200-pair pilots; red: the two VoxelFM encoders; light gray: the other 13. $\rho$: Spearman rank correlation per pilot, 0.72--0.97.}
\label{fig:price}
\end{figure}

\subsection{Number of pairs and encoder choice}
Over 2,550 CT settings (15 encoders, five Qwen3.5 sizes, two layers, $d$ from 16 to 128, $n$ from 100 to 1,600; Appendix~\ref{app:price}), the random-matrix results of Eqs.~\eqref{eq:law}--\eqref{eq:outlier} predict the top three squared canonical correlations with a median error of 0.016, and predict the number of detected directions within one in 71\,\% of settings (Fig.~\ref{fig:price:a}; Fig.~\ref{fig:supp_extra:b}). The error is 0.008--0.030 on the four other datasets, seven further text models, and BIMCV-R CT (Appendix~\ref{app:price}). Counts predicted on CT-RATE transfer to BIMCV-R within one in all 60 matched settings (Table~\ref{tab:bimcv_transfer}). The random-matrix detection threshold holds at every language-model layer and every VoxelFM block (median error 0.006--0.023; Table~\ref{tab:law_depth}). For our configuration, it predicts 145 to 1,032 pairs for the nine shared directions (Fig.~\ref{fig:payoff:a}, triangles), and the AUROC rises from 0.727 at 100 pairs to 0.809 at 1,600 pairs.

The same threshold lets a 200-pair pilot choose the encoder. Over five pilots, its score $S$ (Sec.~\ref{sec:count}) ranks the 15 CT encoders by AUROC at 1,600 pairs with a median Spearman $\rho$ of 0.81 (Fig.~\ref{fig:price:b}; Table~\ref{tab:supp_pilot}). Four pilots select VoxelFM, at most 0.0011 AUROC below the best.

On CT, the image encoder decides the result: it explains 99.0\,\% of the AUROC variance, against 0.7\,\% for the text model, whereas the text model dominates on captions (98.5\,\% on Flickr30k). Scaling Qwen3.5 from 0.8B to 27B helps on captions, but a 110M text embedding model (CXR-BERT on CT, MPNet elsewhere) matches or outperforms \mbox{Qwen3.5-27B} in every dataset (Appendix~\ref{app:ablation}).

\subsection{Results with a few hundred pairs}
\looseness=-1 \textit{LightAlign} outperforms end-to-end models trained without finding labels (Table~\ref{tab:main}; Figs.~\ref{fig:payoff},~\ref{fig:qual}, Table~\ref{tab:supp_payoff_ct}). It surpasses Merlin from 50 and Jolia from 100 pairs. From 200 it outperforms CT-CLIP on AUROC (0.754 vs.\ 0.733), weighted F1, and every evaluation subset. With our report classifier, CT-CLIP drops to 0.682 (gap $[+0.064, +0.082]$; Appendix~\ref{app:payoff}). With 400 pairs it exceeds VocabFine~\cite{arXiv:2403.17834,arXiv:2609.27511}, HLIP~\cite{arXiv:2505.21862}, and fVLM~\cite{arXiv:2501.14548}; with all pairs, COLIPRI (0.824 vs.\ 0.816~\cite{arXiv:2510.15042,arXiv:2609.27511}). All five models above 0.824 use finding labels, up to NV-Reason-CT (0.871, about 70,000 volumes~\cite{arXiv:2609.27511}; Table~\ref{tab:supp_vlm}). On RAD-ChestCT, it rises from 0.636 to 0.724 (100 to all pairs), outperforms CT-CLIP (0.629) and Merlin (0.644), and matches HLIP (0.723).

On the same frozen CT features and pairs, we apply six general few-pair alignment methods: Procrustes, CCA, ASIF-style anchors~\cite{arXiv:2210.01738}, SOTAlign~\cite{arXiv:2602.23353}, Hadgi~\etal~\cite{arXiv:2503.05283}, and a multilayer perceptron (MLP) trained with InfoNCE~\cite{arXiv:1807.03748}. All use the better of the prompt and our report classifier. The baselines reach 0.680--0.790 AUROC at 1,600 pairs against 0.809 for \textit{LightAlign}, which leads all six at 100, 400, and 1,600 pairs (Appendix~\ref{app:payoff}; Sec.~\ref{sec:ablation}). \textit{LightAlign} also leads SOTAlign and Hadgi~\etal on F1, accuracy, and RAD-ChestCT at all pair counts (Table~\ref{tab:main}). At 1,600 pairs with the report classifier, the shared subspace is 0.055 AUROC above a principal-component subspace of the same size and 0.070 above a random one (Appendix~\ref{app:payoff}).

On CheXpert chest radiographs, RAD-DINO~\cite{raddino2025} $\times$ Qwen3.5-4B (middle layer) selected on validation studies reaches 0.862 mean AUROC with 19,858 CheXpert Plus pairs and 0.876 with all 124,769 training pairs. With zero pairs, it remains near the random level (0.502), as on CT (Tables~\ref{tab:cxr} and~\ref{tab:supp_chexpert}). Models trained on MIMIC-CXR~\cite{mimiccxr2019} reach 0.889--0.923 mean AUROC. Our 19,858 pairs are about 5\,\% of CheXzero's and KAD's 377,000.

\subsection{Ablation study}
\label{sec:ablation}
\begin{table}[t]
\centering
\caption{Ablation of \textit{LightAlign} on CT-RATE. Entries: mean AUROC over 18 findings, averaged over five seeds. Columns: number $n$ of image--report pairs. Frozen VoxelFM and Qwen3.5-9B features. Map and readout rows: test set (3,002 scans). Layer rows: the 2,025 validation patients. Bold: best per column and group.}
\label{tab:ablation}
\setlength{\tabcolsep}{4pt}
\small
\begin{tabular}{@{}lccc@{}}
\toprule
& \multicolumn{3}{c}{Image--report pairs $n$} \\
\cmidrule(l){2-4}
Variant & 100 & 400 & 1,600 \\
\midrule
\multicolumn{4}{@{}l}{\emph{Map and readout (test set)}}\\
Ours: ridge regression, report classifier & \textbf{0.727} & \textbf{0.782} & \textbf{0.809} \\
Map: MLP trained with InfoNCE & 0.647 & 0.685 & 0.731 \\
Readout: zero-shot prompt & 0.709 & 0.727 & 0.734 \\
No map: image classifier on the $n$ pairs & 0.698 & 0.767 & 0.806 \\
\midrule
\multicolumn{4}{@{}l}{\emph{Layer choice (validation patients)}}\\
Block 11, middle text layer (ours) & \textbf{0.729} & \textbf{0.783} & \textbf{0.812} \\
Block 10, middle text layer & 0.726 & 0.778 & 0.805 \\
Block 12 (last), middle text layer & 0.726 & 0.779 & 0.805 \\
Block 12 (last), last text layer & 0.722 & 0.776 & 0.801 \\
\bottomrule
\end{tabular}
\end{table}

At 1,600 pairs in Table~\ref{tab:ablation}, \textit{LightAlign} reaches 0.809, an MLP trained with InfoNCE 0.731, and the zero-shot prompt 0.734. From 100 to 1,600 pairs, the report classifier gains 0.082 and the prompt only 0.025. The prompt concentrates its scores on one direction (effective rank 1.04; report classifier 3.07; median over 15 encoders). The report classifier outperforms the prompt in all 120 combinations (Appendix~\ref{app:ablation}). With only the $n$ pairs and their own reports (Table~\ref{tab:supp_reports}), \textit{LightAlign} outperforms a tuned image classifier fitted without the ridge regression on the same $n$ scans and their report-derived labels (Table~\ref{tab:ablation}; Table~\ref{tab:supp_probe_cv}) at 100, 200, 400, and 1,600 pairs: 0.734 vs.\ 0.698, 0.759 vs.\ 0.735, 0.785 vs.\ 0.767, and 0.811 vs.\ 0.806. VoxelFM block 10 or 12 instead of 11, or the last instead of the middle layer of Qwen3.5-9B, performs worse (Table~\ref{tab:supp_layerchoice}).

\section{Conclusion}
\label{sec:conclusion}

Following the Platonic Representation Hypothesis, recent work aligns image and text encoders with zero paired data on natural images. We find that zero-pair alignment fails on four of five datasets in four modalities. Causal tests show that zero-pair alignment needs a large variance share $\eta$ in the shared subspace, but this is still not sufficient: the map must also be oriented, and feature skewness lets the GW loss correct the orientation on CT and chest radiographs without paired data. When $\eta$ is small, the modalities can still share strongly correlated directions, but zero-pair methods cannot locate them. The random-matrix detection threshold holds in every modality and lets a pilot select the best encoder combination and the number of pairs. These findings motivate \textit{LightAlign}, a ridge regression between frozen encoders. We validate it on the two hardest modalities: 3D CT and chest radiographs have the smallest variance share and few paired data. On CT-RATE, 200 pairs suffice to outperform CT-CLIP and all pairs outperform COLIPRI. On the OOD hospital RAD-ChestCT, 100 CT-RATE pairs suffice to outperform CT-CLIP. On CheXpert, \textit{LightAlign} reaches 0.862 mean AUROC with about 5\,\% as many pairs as models trained on MIMIC-CXR. In every dataset, a 110M text embedding model matches or outperforms Qwen3.5-27B.

\section*{Acknowledgments}
The authors gratefully acknowledge the scientific support and HPC resources provided by the Erlangen National High Performance Computing Center (NHR@FAU) of the Friedrich-Alexander-Universit\"at Erlangen-N\"urnberg (FAU). The hardware is partially funded by the German Research Foundation (DFG).

\paragraph{LLM usage.} We used Claude Opus 5.5 by Anthropic to assist with proofreading, coding, and generating plotting scripts. The authors designed all experiments, reviewed every result and figure, and take full responsibility for all methods, results, and claims presented in this paper.
{\small
\bibliographystyle{ieeenat_fullname}
\bibliography{refs}
}
\clearpage
\appendix
\setcounter{table}{0}\setcounter{figure}{0}
\renewcommand{\thetable}{S\arabic{table}}\renewcommand{\thefigure}{S\arabic{figure}}
\section*{Appendix}
This appendix extends Secs.~\ref{sec:method} and~\ref{sec:experiments}, in order of importance.
Section~\ref{app:data} defines the notation, data, and evaluation that the other sections use. Section~\ref{app:payoff} gives the few-pair results behind Table~\ref{tab:main} (Fig.~\ref{fig:payoff}, Tables~\ref{tab:cxr}, \ref{tab:supp_fewpair}, and~\ref{tab:supp_vlm}).
Sections~\ref{app:locate}, \ref{app:orient}, and~\ref{app:price} derive and test the locate, orient, and pair-count steps (Tables~\ref{tab:supp_zeropair}, \ref{tab:supp_orient_seeds}, and~\ref{tab:supp_law_domains}; Algorithm 1 in Fig.~\ref{alg:zero} and Algorithm 2 in Fig.~\ref{alg:few}).
Section~\ref{app:ablation} contains the ablations (Sec.~\ref{sec:ablation}), Section~\ref{app:qual} the qualitative examples (Fig.~\ref{fig:qual}), and Section~\ref{app:impl} the implementation details.
Section~\ref{app:tables} holds all result tables.

\section{Notation, data, and evaluation}
\label{app:data}
This section defines the notation, the datasets and splits, the feature extraction, and the metrics that the other sections use.
\paragraph{Symbols.}\leavevmode\\
{\small\begin{tabular}{@{}l>{\raggedright\arraybackslash}p{0.72\columnwidth}@{}}
$\mathbf{x},\mathbf{y}$ & frozen image and text features, centered, unit norm\\
$n$ & number of image--report pairs\\
$d$ & principal components kept per side before canonical correlation analysis (CCA) (64)\\
$c=d/n$ & aspect ratio\\
$k$ & number of shared directions\\
$\eta$ & fraction of image feature variance in the shared directions (Sec.~\ref{sec:locate})\\
$r_i$ & population squared canonical correlation of shared direction $i$\\
$\lambda_i$ & sample squared canonical correlation\\
$\lambda_+$ & noise bulk edge, $4c(1-c)$\\
$\gamma(r)$ & limit of $\lambda_i$ above the threshold, $r(1-c+c/r)^2$\\
$n^\star_i$ & pairs that detect direction $i$, $d(1+1/r_i)$\\
$\hat r_i$ & estimate of $r_i$ from a pilot\\
$S$ & pilot score, $\sum_{i\,\text{detected}}\hat r_i$\\
$\beta$ & scale factor of the shared-direction scaling (Eq.~1)\\
$\mathbf{s},\,t$ & sign pattern over the top $t=10$ directions (Eq.~2)\\
$M$ & items per side in the GW sign selection\\
$\mathbf{X}_n,\mathbf{Y}_n$ & image and text features of the $n$ pairs\\
$\alpha$ & ridge penalty\\
$g^{\mathrm{P}}_\ell,g^{\mathrm{R}}_\ell$ & zero-shot prompt and report classifier scores of finding $\ell$\\
\end{tabular}}
\paragraph{Stratified area under the receiver operating characteristic curve (AUROC).} Many CT findings co-occur with the overall disease burden and with report length, so on CT we also compute a stratified AUROC. We group patients by finding count (up to 5, or 6 and above) and by report-length tercile, compute an AUROC inside each group, and pool the groups weighted by positives times negatives. A score that only ranks patients by overall sickness obtains about 0.5.
\paragraph{Models.} CT encoders (15): VoxelFM~\cite{arXiv:2604.04133} (class token and patch mean), CT-FM~\cite{arXiv:2501.09001}, DALE-CT~\cite{arXiv:2606.07775} (two variants), V-JEPA 2~\cite{arXiv:2506.09985} (L, H, and g),
Wan~\cite{arXiv:2503.20314} variational autoencoder (VAE), Wan diffusion transformer (DiT; a text-to-video model, run with the empty prompt), LTX VAE~\cite{arXiv:2601.03233} (LTX-2.5), VideoMAE-B~\cite{arXiv:2203.12602}, DINOv3~\cite{arXiv:2508.10104} (2D and 2.5D), and DINOv2~\cite{arXiv:2304.07193} 2D; VoxelFM features from each of blocks 2 to 12.
Text models (13): Qwen3.5~\cite{qwen35} 0.8B, 2B, 4B, 9B, and 27B, all-mpnet-base-v2 (MPNet)~\cite{arXiv:2004.09297,arXiv:1908.10084}, CXR-BERT~\cite{arXiv:2204.09817}, MedGemma 1.5 4B~\cite{arXiv:2604.05081}, Granite 4.2 8B~\cite{granite42},
Qwen3-VL-Embedding 8B~\cite{arXiv:2601.04720} (Qwen3-VL-Emb.\ in tables), Ministral 3 14B~\cite{arXiv:2601.08584}, Gemma 4 31B~\cite{arXiv:2607.02770}, and OLMo 3.1 32B~\cite{arXiv:2512.13961}. Audio encoders: AST~\cite{arXiv:2104.01778}, EAT~\cite{arXiv:2401.03497}, and Dasheng~\cite{arXiv:2406.06992}.
VoxelFM was pre-trained without reports. Its authors used only the official training split of CT-RATE and excluded the validation set, which is our test set~\cite{arXiv:2604.04133}. RAD-ChestCT is not among its pre-training datasets.

\paragraph{Datasets and splits.}
\emph{CT-RATE}~\cite{arXiv:2403.17834} (v2 release, voxels in Hounsfield units, HU). Each patient contributes one reconstruction (scan a,
reconstruction 1), after removing the volumes on the dataset's own non-chest lists. We call one such reconstruction a scan. Reports are the English findings
followed by the impression; the 18 finding labels are the dataset's report-derived labels. The training set is a hashed
30\,\% of the training patients, 5,994 patients, split by a hash of the patient id into an estimation split (2,024 patients, used to estimate the shared directions), a
pair split (1,945; its two halves of 972 and 973 items feed the zero-pair aligner), and validation patients (2,025), which
we use for every selection before the official set is read. The official validation set, our test set, has 3,039 volumes. We
evaluate four subsets: the 3,002 chest volumes after removing the 37 non-chest volumes (the headline subset), all 3,039 volumes, one reconstruction per
study (1,564), and the original a\_1 reconstructions (1,293). No slice-thickness or HU-range exclusion is applied; all volumes are
resampled to 1\,mm.
\emph{BIMCV-R}~\cite{bimcvr2024} (OOD CT dataset). From 8,069 metadata rows we drop the rows flagged as radiograph reports (224), with disagreeing file and report identifiers (1,333), as untranslated (551), or as sharing a report with other rows (1,001); the flags overlap. We then sample 24\,\% of the remaining patients by hash, one volume per patient: 1,090 patients (880 train, 210 test). During extraction 141
volumes are rejected (122 implausible voxel spacing, 1 implausible field of view, 1 not HU-like, 17 without lung in
the TotalSegmentator~\cite{totalseg2023} mask), leaving 949 (773 / 176). Fourteen of the 18 CT-RATE labels have a counterpart in the
dataset's labels, each with at least 20 positives.
\emph{RAD-ChestCT}~\cite{draelos2021radchestct} (OOD hospital). The public release: 3,630 non-contrast chest CT
volumes, stored in HU clipped to $[-1000,1000]$ at 0.8\,mm isotropic voxels, with 84 abnormality labels per lung or
organ location. All 3,630 volumes are evaluated; none is used for fitting. A finding is positive when it is positive at
any location. Sixteen of the 18 CT-RATE labels are mapped as in CT-CLIP~\cite{arXiv:2403.17834}: mosaic attenuation
pattern has no counterpart, and arterial and coronary artery wall calcification form one label, scored by the larger
of the two classifier scores. The slice axis is reversed to the CT-RATE orientation, and volumes are resampled to
1\,mm.
\emph{COCO}~\cite{coco2014} (2017 train): 12,000 images with their first caption; \emph{Flickr30k}~\cite{flickr30k2014}: 12,000 images, first caption;
\emph{IU X-ray}~\cite{iuxray2016}: 3,826 studies (frontal and lateral images averaged; report = findings and impression);
\emph{Clotho}~\cite{clotho2020} v2.1: 4,884 clips, first caption. These four are split by a hash of the item id into one sixth
estimation, one sixth validation, and four sixths training (COCO 2,049 / 2,020 / 7,931; Flickr30k 1,995 / 1,990 / 8,015;
IU X-ray 649 / 644 / 2,533; Clotho 861 / 773 / 3,250). \emph{CheXpert}~\cite{chexpert2019}: 30,000 training studies of CheXpert Plus~\cite{chexpertplus2024}, drawn uniformly at random,
one frontal image per study, split by patient into estimation (5,124), pairs (19,858),
and validation (5,018); reports are findings and impression; evaluation uses the official test set (500 studies) with the
radiologist labels of the five competition findings, and the official validation set (200 studies) only to set the
thresholds of the threshold metrics. The encoder choice and the zero-pair row use the first draw, 6,000 training studies plus the 200 validation studies (the 6,200-image subset).
\paragraph{Feature extraction.}
CT volumes are resampled to 1\,mm isotropic voxels. \emph{VoxelFM} (3D vision transformer, ViT-B) takes HU values
clipped to $[-1000,1900]$ and normalized by the model's training statistics; we extract the class token after the final
layer norm (864-d), the mean of the patch tokens (VoxelFM patch), or the class token after the transformer blocks 2--12.
\emph{CT-FM} (3D SegResEncoder) clips HU to $[-1024,2048]$, scales them to $[0,1]$, and uses 3\,mm slices; we average its last stage (512-d). Slice-based models
take the body-cropped volume: \emph{DALE-CT} (0 and 1S) and \emph{DINOv2/DINOv3} take an axial slice every 4\,mm and
average the per-slice class tokens (1024-d; DINOv3 2.5D stacks three neighboring slices); the video models \emph{V-JEPA 2}
(L/H/g, 64 frames, 256\,px), \emph{VideoMAE-B} (16 frames), and the \emph{Wan} and \emph{LTX} VAEs (65 frames) take the axial
slices as frames, with the lung, soft-tissue, and bone windows as color channels, and we average their tokens (Wan VAE: latent
pooled to $2\times4\times4$, 512-d; Wan DiT: tokens of the first 15 of 30 blocks at flow time 0.2). For the other datasets,
images are resized to 224\,px: DINOv2-L and DINOv3-L give the class token (1024-d; also the token mean for DINOv2),
RAD-DINO~\cite{raddino2025} the class token (768-d). Audio is resampled to 16\,kHz and split into 10.24\,s windows: AST (pooled output),
EAT (class token or token mean), and Dasheng (time mean), all 768-d, averaged over windows. Text models: decoder language
models (Qwen3.5 0.8B--27B and the other models of the main text) read the report without a chat template, truncated at
1,024 tokens, and we average the hidden states of the middle entry of the hidden-state stack (index 16 of 33 for
Qwen3.5-4B) over tokens; MPNet gives its mean-pooled output embedding (384 tokens); CXR-BERT reads 512 tokens. CXR-BERT and Qwen3-VL-Embedding also give the token mean of their middle hidden layer, not their output embedding.\looseness=-1
\paragraph{Metrics.} FOSCTTM (fraction of samples closer than the true match)~\cite{liu2019mmdma}: for each validation image, the fraction
of wrong reports that the map places closer to it than its own report (ties count one half), averaged over images and over both directions (image to report and report to image); 0 is a perfect match and
0.5 the random level. AUROC: mean over findings of the area under the ROC curve; on CT-RATE the 18 findings on the
3,002 chest scans of the official validation set, on CheXpert the five competition findings on the 500
test studies. Intervals: paired bootstrap over patients with 1,000 resamples.

\section{Few-pair results}
\label{app:payoff}
\begin{figure*}[tp]
\centering
\includegraphics[width=\textwidth]{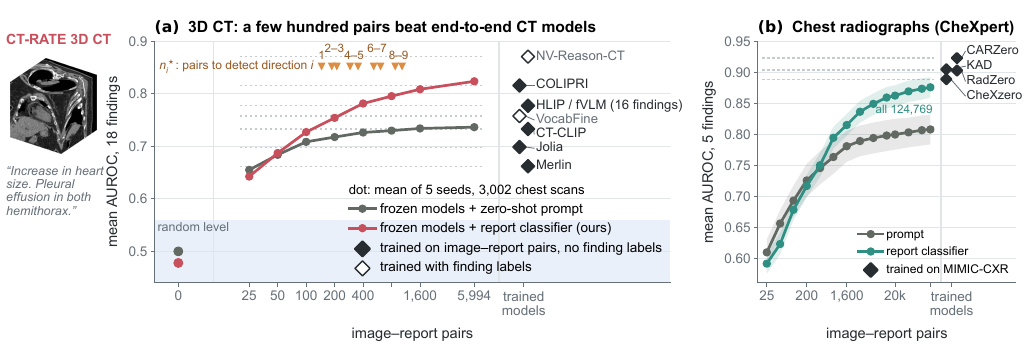}%
{\phantomsubcaption\label{fig:payoff:a}}{\phantomsubcaption\label{fig:payoff:b}}%
\caption{Few-pair curves behind Table~\ref{tab:main}. \textbf{(a)}~3D CT: mean AUROC over 18 findings on the
3,002 scans of the CT-RATE validation set (VoxelFM block 11, Qwen3.5-9B middle layer; dots: mean of five seeds). Gray line: frozen models with the zero-shot prompt; red line: frozen models with the report classifier (ours); shaded band: random level.
Diamonds and dotted guide lines: end-to-end models of Table~1 (filled: trained on image--report pairs without finding labels; open: trained with finding labels, VocabFine and NV-Reason-CT; HLIP and fVLM: 16 findings). Orange triangles, numbered 1 to 9, mark the pairs $n^\star_i$ that each of the nine shared directions needs, estimated on 2,024 patients. CT example: a cube cut from the volume; its faces are axial, coronal, and sagittal slices.
\textbf{(b)}~CheXpert test set, mean AUROC over five findings (RAD-DINO, Qwen3.5-4B). Gray line: prompt; teal line: report classifier; band: study-level bootstrap 95\,\% interval; diamonds, each named, with dotted guide lines: models trained on MIMIC-CXR (CheXzero, RadZero, KAD, CARZero). Both panels: logarithmic pair axis.}
\label{fig:payoff}
\end{figure*}

This section reports the few-pair results behind Table~\ref{tab:main}: the curves over the number of pairs (Fig.~\ref{fig:payoff}), classification on chest radiographs (Table~\ref{tab:cxr}), and the few-pair results and more CT baselines (Tables~\ref{tab:supp_fewpair} and~\ref{tab:supp_vlm}).

\begin{table}[tp]
\centering
\caption{Classification of five findings on chest radiographs. CheXpert official test set, 500 studies, five competition findings, radiologist consensus labels; mean AUROC; Table~\ref{tab:supp_thresh} reports threshold metrics. Pairs: image--report pairs used for training or fitting.
Baselines train on MIMIC-CXR. Ours: frozen RAD-DINO and Qwen3.5-4B middle layer (selected on validation studies), ridge regression on $n$ CheXpert Plus pairs, report classifier; classifier fitted on the 19,858 reports of the first rows; $n=0$ on the 6,200-image subset.}
\label{tab:cxr}
\setlength{\tabcolsep}{3.5pt}
\small
\begin{tabular}{lccrc}
\toprule
& \multicolumn{2}{c}{Trained?} & & \\
\cmidrule(lr){2-3}
Method & Image enc. & Text enc. & Pairs & AUROC$\uparrow$ \\
\midrule
CheXzero~\cite{tiu2022chexzero} & \yes & \yes & 377k & 0.889$^{b}$ \\
RadZero~\cite{arXiv:2504.07416} & \no$^{a}$ & \yes & 353k$^{c}$ & 0.903 \\
KAD~\cite{zhang2023kad} & \yes & \yes & 377k & 0.905 \\
CARZero~\cite{arXiv:2402.17417} & \yes & \yes & --$^{d}$ & 0.923 \\
\midrule
\rowcolor{oursbg}\multicolumn{5}{l}{\emph{Ours: frozen RAD-DINO + Qwen3.5-4B, ridge regression,}}\\
\rowcolor{oursbg}\multicolumn{5}{l}{\emph{report classifier, no image labels}}\\
\rowcolor{oursbg} Ours &  &  & 0 & 0.502 \\
\rowcolor{oursbg} Ours &  &  & 200 & 0.717 \\
\rowcolor{oursbg} Ours &  &  & 400 & 0.750 \\
\rowcolor{oursbg} Ours &  &  & 1,600 & 0.815 \\
\rowcolor{oursbg} Ours &  &  & 6,400 & 0.849 \\
\rowcolor{oursbg} Ours &  &  & 19,858 & 0.862 \\
\rowcolor{oursbg} Ours &  &  & 40,000 & 0.869 \\
\rowcolor{oursbg} Ours &  &  & 80,000 & 0.874 \\
\rowcolor{oursbg} Ours & \multirow{-9}{*}{\no} & \multirow{-9}{*}{\no} & 124,769 & 0.876 \\
\bottomrule
\end{tabular}
\par\vspace{2pt}\begin{minipage}{\linewidth}\footnotesize $^{a}$frozen image encoder with trained layers on top; $^{b}$ensemble of ten checkpoints, best single checkpoint 0.878; $^{c}$training images, 224\,px variant; $^{d}$frontal views only, count not stated.\end{minipage}
\end{table}

\begin{table*}[tp]
\centering
\caption{MLP: multilayer perceptron. Each entry is the mean AUROC over seeds for one alignment method, one scoring method, and one pair count (the sample standard deviation, or the bootstrap interval of the difference from CT-CLIP). The validation block is VoxelFM with Qwen3.5-4B (middle layer), five seeds, mean and stratified AUROC (the mean of the AUROCs within strata of finding count and report length) at 100, 400, and 1,600 pairs. The official block is VoxelFM block 11 with Qwen3.5-9B (middle layer), 3,002 chest scans, 18 findings.}
\label{tab:supp_fewpair}
\setlength{\tabcolsep}{3pt}
\footnotesize
\begin{tabular*}{\textwidth}{@{\extracolsep{\fill}}lccccccc@{}}
\toprule
\multicolumn{8}{l}{\emph{Validation split}}\\
\midrule
Method & scoring method & \multicolumn{3}{c}{mean AUROC} & \multicolumn{3}{c}{stratified AUROC} \\
\cmidrule(lr){3-5}\cmidrule(lr){6-8}
 & & 100$\uparrow$ & 400$\uparrow$ & 1,600$\uparrow$ & 100$\uparrow$ & 400$\uparrow$ & 1,600$\uparrow$ \\
\midrule
Procrustes & prompt & 0.669{\scriptsize$\pm$0.018} & 0.693{\scriptsize$\pm$0.007} & 0.687{\scriptsize$\pm$0.003} & 0.529{\scriptsize$\pm$0.004} & 0.536{\scriptsize$\pm$0.005} & 0.536{\scriptsize$\pm$0.001} \\
Procrustes & classifier & 0.632{\scriptsize$\pm$0.008} & 0.671{\scriptsize$\pm$0.006} & 0.698{\scriptsize$\pm$0.002} & 0.592{\scriptsize$\pm$0.005} & 0.636{\scriptsize$\pm$0.004} & 0.669{\scriptsize$\pm$0.002} \\
CCA & prompt & 0.654{\scriptsize$\pm$0.019} & 0.673{\scriptsize$\pm$0.009} & 0.662{\scriptsize$\pm$0.006} & 0.529{\scriptsize$\pm$0.004} & 0.532{\scriptsize$\pm$0.006} & 0.531{\scriptsize$\pm$0.001} \\
CCA & classifier & 0.552{\scriptsize$\pm$0.012} & 0.608{\scriptsize$\pm$0.005} & 0.682{\scriptsize$\pm$0.004} & 0.538{\scriptsize$\pm$0.007} & 0.582{\scriptsize$\pm$0.005} & 0.636{\scriptsize$\pm$0.002} \\
Anchors & prompt & 0.661{\scriptsize$\pm$0.027} & 0.691{\scriptsize$\pm$0.009} & 0.700{\scriptsize$\pm$0.002} & 0.531{\scriptsize$\pm$0.004} & 0.540{\scriptsize$\pm$0.002} & 0.542{\scriptsize$\pm$0.000} \\
Anchors & classifier & 0.603{\scriptsize$\pm$0.017} & 0.637{\scriptsize$\pm$0.015} & 0.660{\scriptsize$\pm$0.004} & 0.574{\scriptsize$\pm$0.014} & 0.603{\scriptsize$\pm$0.008} & 0.631{\scriptsize$\pm$0.005} \\
SOTAlign & prompt & 0.566{\scriptsize$\pm$0.029} & 0.552{\scriptsize$\pm$0.015} & 0.510{\scriptsize$\pm$0.015} & 0.520{\scriptsize$\pm$0.007} & 0.516{\scriptsize$\pm$0.012} & 0.511{\scriptsize$\pm$0.004} \\
SOTAlign & classifier & 0.665{\scriptsize$\pm$0.004} & 0.697{\scriptsize$\pm$0.002} & 0.718{\scriptsize$\pm$0.003} & 0.587{\scriptsize$\pm$0.007} & 0.624{\scriptsize$\pm$0.006} & 0.652{\scriptsize$\pm$0.003} \\
Hadgi & prompt & 0.669{\scriptsize$\pm$0.016} & 0.694{\scriptsize$\pm$0.002} & 0.709{\scriptsize$\pm$0.001} & 0.522{\scriptsize$\pm$0.006} & 0.525{\scriptsize$\pm$0.010} & 0.527{\scriptsize$\pm$0.002} \\
Hadgi & classifier & 0.692{\scriptsize$\pm$0.011} & 0.753{\scriptsize$\pm$0.005} & 0.790{\scriptsize$\pm$0.003} & 0.605{\scriptsize$\pm$0.009} & 0.657{\scriptsize$\pm$0.006} & 0.695{\scriptsize$\pm$0.005} \\
Image probe & image classifier & 0.693{\scriptsize$\pm$0.008} & 0.729{\scriptsize$\pm$0.001} & 0.742{\scriptsize$\pm$0.002} & 0.613{\scriptsize$\pm$0.007} & 0.653{\scriptsize$\pm$0.005} & 0.664{\scriptsize$\pm$0.003} \\
MLP + InfoNCE & prompt & 0.575{\scriptsize$\pm$0.041} & 0.635{\scriptsize$\pm$0.070} & 0.691{\scriptsize$\pm$0.030} & 0.518{\scriptsize$\pm$0.017} & 0.521{\scriptsize$\pm$0.011} & 0.537{\scriptsize$\pm$0.007} \\
MLP + InfoNCE & classifier & 0.663{\scriptsize$\pm$0.011} & 0.683{\scriptsize$\pm$0.050} & 0.718{\scriptsize$\pm$0.041} & 0.588{\scriptsize$\pm$0.008} & 0.626{\scriptsize$\pm$0.026} & 0.659{\scriptsize$\pm$0.023} \\
Ours (ridge) & prompt & 0.707{\scriptsize$\pm$0.009} & 0.718{\scriptsize$\pm$0.003} & 0.721{\scriptsize$\pm$0.002} & 0.531{\scriptsize$\pm$0.006} & 0.535{\scriptsize$\pm$0.006} & 0.535{\scriptsize$\pm$0.001} \\
Ours (ridge) & classifier & 0.727{\scriptsize$\pm$0.008} & 0.779{\scriptsize$\pm$0.002} & 0.802{\scriptsize$\pm$0.003} & 0.618{\scriptsize$\pm$0.006} & 0.669{\scriptsize$\pm$0.002} & 0.696{\scriptsize$\pm$0.006} \\
\bottomrule
\end{tabular*}
\par\medskip
\setlength{\tabcolsep}{3pt}
\footnotesize
\begin{tabular*}{\textwidth}{@{\extracolsep{\fill}}llccc@{}}
\toprule
\multicolumn{5}{l}{\emph{Official validation, chest, mean AUROC, 18 findings}}\\
\midrule
Method & scoring method & 100$\uparrow$ & 400$\uparrow$ & 1,600$\uparrow$ \\
\midrule
Procrustes & prompt & 0.662{\scriptsize[-0.080,\,-0.064]} & 0.675{\scriptsize[-0.065,\,-0.051]} & 0.677{\scriptsize[-0.063,\,-0.048]} \\
Procrustes & classifier & 0.632{\scriptsize[-0.111,\,-0.091]} & 0.666{\scriptsize[-0.077,\,-0.056]} & 0.680{\scriptsize[-0.065,\,-0.040]} \\
CCA & prompt & 0.637{\scriptsize[-0.106,\,-0.088]} & 0.658{\scriptsize[-0.083,\,-0.068]} & 0.658{\scriptsize[-0.084,\,-0.068]} \\
CCA & classifier & 0.549{\scriptsize[-0.194,\,-0.174]} & 0.610{\scriptsize[-0.133,\,-0.113]} & 0.681{\scriptsize[-0.061,\,-0.043]} \\
Anchors & prompt & 0.662{\scriptsize[-0.080,\,-0.062]} & 0.684{\scriptsize[-0.057,\,-0.041]} & 0.698{\scriptsize[-0.044,\,-0.028]} \\
Anchors & classifier & 0.598{\scriptsize[-0.145,\,-0.125]} & 0.639{\scriptsize[-0.105,\,-0.084]} & 0.669{\scriptsize[-0.075,\,-0.053]} \\
SOTAlign & prompt & 0.543{\scriptsize[-0.201,\,-0.179]} & 0.555{\scriptsize[-0.190,\,-0.167]} & 0.549{\scriptsize[-0.196,\,-0.171]} \\
SOTAlign & classifier & 0.644{\scriptsize[-0.098,\,-0.080]} & 0.670{\scriptsize[-0.072,\,-0.053]} & 0.702{\scriptsize[-0.041,\,-0.022]} \\
Hadgi & prompt & 0.646{\scriptsize[-0.095,\,-0.079]} & 0.684{\scriptsize[-0.056,\,-0.043]} & 0.706{\scriptsize[-0.034,\,-0.020]} \\
Hadgi & classifier & 0.685{\scriptsize[-0.056,\,-0.040]} & 0.752{\scriptsize[0.011,\,0.027]} & 0.790{\scriptsize[0.050,\,0.066]} \\
Schnaus & prompt & 0.613{\scriptsize[-0.128,\,-0.112]} & 0.634{\scriptsize[-0.107,\,-0.091]} & 0.644{\scriptsize[-0.099,\,-0.080]} \\
Schnaus & classifier & 0.607{\scriptsize[-0.135,\,-0.117]} & 0.650{\scriptsize[-0.094,\,-0.073]} & 0.666{\scriptsize[-0.078,\,-0.055]} \\
MLP + InfoNCE & prompt & 0.582{\scriptsize[-0.162,\,-0.142]} & 0.651{\scriptsize[-0.089,\,-0.074]} & 0.687{\scriptsize[-0.054,\,-0.038]} \\
MLP + InfoNCE & classifier & 0.647{\scriptsize[-0.096,\,-0.078]} & 0.685{\scriptsize[-0.056,\,-0.039]} & 0.731{\scriptsize[-0.011,\,0.007]} \\
\bottomrule
\end{tabular*}
\end{table*}

\begin{table*}[tp]
\centering
\caption{Further CT-RATE baselines, copied from Table~3 of NV-Reason-CT~\cite{arXiv:2609.27511} (18 findings). Its re-evaluations use the corrected
validation set of 3,002 chest reconstructions; several imported results use the earlier 3,039-case set. Generative vision--language models (VLMs) answer Yes/No per finding; we report macro-F1 from these answers, and AUROC where the paper reports a score; NV-Reason-CT is trained with supervised fine-tuning (SFT) and reinforcement learning (RL); macro-F1 uses each
paper's own threshold or answer rule and is not comparable to the weighted F1 of Table~\ref{tab:main}.}
\label{tab:supp_vlm}
\small
\begin{tabular}{@{}llcc@{}}
\toprule
Model & Evaluation & Macro-F1$\uparrow$ & AUROC$\uparrow$ \\
\midrule
\multicolumn{4}{@{}l}{\emph{Generative VLMs, zero-shot Yes/No answers}}\\
RadFM~\cite{arXiv:2308.02463} & native Yes/No & 0.2153 & -- \\
MedGemma 1.5~\cite{arXiv:2604.05081} & Yes/No per finding & 0.3033 & -- \\
M3D-LaMed~\cite{arXiv:2404.00578} & one-token Yes/No & 0.3046 & -- \\
ClinFusion-32B~\cite{arXiv:2607.24743} & single Yes/No prompt & 0.3741 & -- \\
ClinFusion-8B~\cite{arXiv:2607.24743} & single Yes/No prompt & 0.4419 & -- \\
NV-Reason-CT~\cite{arXiv:2609.27511} & Yes/No, SFT + RL & 0.6140 & 0.8710 \\
\midrule
\multicolumn{4}{@{}l}{\emph{Contrastive models, zero-shot prompts}}\\
CT-CLIP VocabFine~\cite{arXiv:2403.17834} & Pos+Neg prompts & 0.2662 & 0.7583 \\
Merlin~\cite{arXiv:2406.06512} & Pos+Neg prompts & 0.3579 & 0.6617 \\
Jolia~\cite{arXiv:2606.24570} & 8 prompt pairs & 0.3589 & 0.6986 \\
CT-CLIP~\cite{arXiv:2403.17834} & Pos+Neg prompts & 0.3976 & 0.7328 \\
\midrule
\multicolumn{4}{@{}l}{\emph{Heads trained with finding labels}}\\
CT-CLIP + head & retrained head & 0.2760 & 0.5740 \\
M3D-LaMed + head & retrained head & 0.4257 & 0.7403 \\
RadFM + head & retrained head & 0.4325 & 0.7592 \\
Merlin + head & retrained head & 0.4955 & 0.7983 \\
Pillar-0 + head~\cite{arXiv:2511.17803,arXiv:2604.10437} & linear head (DCP-1) & 0.5440 & 0.8610 \\
CT-SSG~\cite{dipiazza2026ctssg} & authors' eval & 0.5718 & 0.8364 \\
VoxelFM + Q-Former~\cite{arXiv:2604.04133} & authors' eval & 0.5807 & 0.8701 \\
\bottomrule
\end{tabular}
\end{table*}

\paragraph{Against CT-CLIP.} The paired bootstrap interval of the AUROC gain over CT-CLIP is $[+0.014,+0.028]$ at 200 pairs. The gain also holds for weighted F1 (Table~\ref{tab:main}), on all four evaluation subsets, and on the 16-label set. With the zero-shot prompt in place of the report classifier, neither model uses report-derived labels. \textit{LightAlign} then matches CT-CLIP from 800 pairs: the interval of the AUROC difference is $[-0.009,+0.003]$ at 800 pairs, $[-0.005,+0.007]$ at 1,600 pairs (0.734 vs.\ 0.733), and $[-0.003,+0.010]$ with all pairs (0.737).\looseness=-1

\paragraph{Threshold metrics.} The F1 and accuracy values we compute on CT-RATE in Table~\ref{tab:main}, including the zero-pair, SOTAlign, and Hadgi~\etal rows, use the ROC-corner criterion of CT-CLIP~\cite{arXiv:2403.17834}. For each finding, we min-max scale the scores to $[0,1]$, and pick the threshold among 100 evenly spaced points in $[0,1]$ that is closest to the top-left corner of the ROC curve on the evaluated volumes. We then compute the weighted F1 and the accuracy, average over the 18 findings, and average over seeds. For CT-CLIP we use its released probabilities without scaling. Table~\ref{tab:supp_thresh} lists the weighted F1 and accuracy for 50, 100, 200, 400, 1,600, and 5,994 pairs and every evaluation subset. On CheXpert, the threshold maximizes the Matthews correlation coefficient (MCC) on the validation set, as in CheXzero and KAD. With all 19,858 pairs, the report classifier reaches MCC 0.431 and F1 0.513, against CheXzero's 0.523 and 0.606 and KAD's 0.589 and 0.646, both trained on about 377k MIMIC-CXR~\cite{mimiccxr2019} pairs.
\paragraph{Few-pair baselines and the shared subspace.} With 1,600 pairs, the CCA map of Hadgi~\etal\ with the report classifier reaches 0.790, lower than \textit{LightAlign} (0.809). The paired bootstrap interval of the AUROC gain of \textit{LightAlign} over Hadgi~\etal\ is $[+0.038,+0.046]$ at 100 pairs, $[+0.027,+0.033]$ at 400 pairs, and $[+0.016,+0.021]$ at 1,600 pairs. We also run the Wasserstein Procrustes method of Schnaus~\etal~\cite{arXiv:2610.09411} with the same pairs and the authors' default configuration. With the report classifier, it reaches 0.666 with 1,600 pairs and lags behind CT-CLIP at 100, 400, and 1,600 pairs. We list all baselines with both scoring methods in Table~\ref{tab:supp_fewpair}. The shared subspace improves the AUROC over a principal-component subspace by $+0.055$ and over a random subspace of the same size by $+0.070$.

\paragraph{Prediction versus alignment.} Kamai~\etal~\cite{arXiv:2606.11190} study linear maps in a spiked model. Cross-modal alignment is equivalent to CCA, which whitens both modalities; it fails when nuisance variation is correlated across modalities. Cross-modal prediction is equivalent to reduced-rank regression, which whitens only the source modality; its recovery depends on the quality of the source modality. Ridge regression from images to reports is a regularized prediction map. In our experiments, it outperforms the two CCA-based maps (CCA and Hadgi~\etal) at 100, 400, and 1,600 pairs (Table~\ref{tab:supp_fewpair}). Kamai~\etal\ also note that the regime where both objectives fail likely requires structure beyond the pairwise cross-covariance. GW sign selection (Sec.~\ref{sec:method}) uses such structure and requires no pairs: the third moment of the projections onto the shared directions.
\paragraph{Algorithm 2 details.} We choose the ridge penalty $\alpha$ from 13 log-spaced values between $10^{-3}$ and $10^{3}$, by leave-one-out cross-validation on the $n$ pairs, which can be done in closed form. The report classifier is one logistic regression per finding on the frozen text features of the unpaired training reports, with the report-derived labels of the dataset (standardized features, L2 penalty with $C=1$, L-BFGS solver, the scikit-learn defaults). The all-pairs map uses all 5,994 training patients, including the validation patients of the configuration choice; the test set is disjoint from them. A new scan $\mathbf{x}$ is scored by the logit $\mathbf{w}_\ell^{\top}\mathbf{W}^{\top}\mathbf{x}+b_\ell$.
\paragraph{Fixed classifier, different encoders.} The report classifier trained on CT-CLIP's own text embeddings of the 5,994 training reports, applied to CT-CLIP's image embeddings, scores 0.682 on the 3,002 chest scans (all 3,039: 0.679; study subset 0.679; a\_1: 0.685), below CT-CLIP's prompts on the same rerun (0.733). Our 200-pair configuration outperforms this CT-CLIP variant by $[+0.064, +0.082]$ (paired bootstrap). A logistic probe trained on the VoxelFM block-11 image features with the report-derived labels of all 5,994 training images, with $C$ tuned by cross-validation, scores 0.824 (Table~\ref{tab:supp_probe_cv}). We did not run COLIPRI; we quote its zero-shot 0.816 as re-evaluated in~\cite{arXiv:2609.27511} and its supervised probe 0.862 from its authors~\cite{arXiv:2510.15042}.
\paragraph{Variance decomposition and score rank.} On validation CT patients we fit two-way sums of squares to
the $15\times8$ grid of combination means (15 CT encoders, 8 text models, report classifier, 1,600 pairs):
the encoder explains 99.0\,\% of the mean-AUROC variance, the text model 0.7\,\% and the remainder 0.3\,\% (stratified
AUROC: 93.2\,\% / 5.7\,\%). For the zero-shot prompt the order reverses (text model 89.6\,\%). The effective rank of the score matrix (findings by scans; exponential of the entropy of
its normalized squared singular values) is 1.04 for the prompt and 3.07 for the report classifier (median over 15
encoders and five seeds, Qwen3.5-4B, 1,600 pairs); the prompt's median weight on the top singular direction is 0.90. The report classifier keeps finding-specific information beyond the axis common to all findings. When we regress each finding's scores on the mean score across the 18 findings and retain the residuals, \textit{LightAlign} remains above CT-CLIP at all five pair counts of Table~\ref{tab:supp_resid}. At 200 pairs, its residual AUROC is 0.641 versus 0.595 for CT-CLIP (interval $[+0.036,+0.058]$), and the gap widens to $[+0.106,+0.130]$ at 1,600 pairs. The effective rank of its scores is 4.26 at 200 pairs versus 1.63 for CT-CLIP. Di Folco~\etal~\cite{arXiv:2503.04478} map a frozen chest X-ray encoder to a frozen medical text encoder via a transformation fitted on anchor pairs. They report that zero-shot classification through prompt similarity remains challenging for all models. This supports our finding that the bottleneck lies in the prompt readout, not the map.
With the singular value decomposition $\mathbf{W}=\sum_j \sigma_j\mathbf{a}_j\mathbf{b}_j^{\top}$, the weight of a scoring direction $\mathbf{v}_\ell$ on the top singular direction is $(\sigma_1\mathbf{b}_1^{\top}\mathbf{v}_\ell)^2/\sum_j (\sigma_j\mathbf{b}_j^{\top}\mathbf{v}_\ell)^2$ (median over findings).
\paragraph{External hospital (RAD-ChestCT).} We apply the map and the report classifier fitted on CT-RATE, without any changes, to the 3,630 RAD-ChestCT volumes. The 18 CT-RATE findings map to the 16 RAD-ChestCT labels as in CT-CLIP~\cite{arXiv:2403.17834}: the two scores for arterial and coronary wall calcification are merged by taking their maximum, and mosaic attenuation pattern is dropped. The mean AUROC over five seeds ($\pm$ sample standard deviation) is 0.512$\pm$0.016 using 0 pairs, 0.610$\pm$0.013 using 50, 0.636$\pm$0.010 using 100, 0.658$\pm$0.012 using 200, 0.674$\pm$0.003 using 400, 0.690$\pm$0.007 using 800, 0.705$\pm$0.005 using 1,600, and 0.724 using all 5,994 pairs. Over 1,000 bootstrap resamples of the 3,630 volumes, the 95\,\% interval of the mean AUROC using 100 pairs is $[0.631, 0.641]$, above the 0.629 that CT-CLIP reports~\cite{arXiv:2403.17834}. With the same features, pairs, and report classifier, Hadgi~\etal\ reach 0.603, 0.652, and 0.686, and SOTAlign 0.573, 0.597, and 0.622 at 100, 400, and 1,600 pairs (Table~\ref{tab:main}). On the 14 labels that HLIP~\cite{arXiv:2505.21862} reports, we reach 0.727 with all 5,994 pairs and 0.711 with 1,600 using two frozen encoders and one ridge regression, whereas HLIP trains its image encoder end to end and reports 0.723. The ranking of the 15 CT encoders also transfers: with ridge regression on 100 pairs (Qwen3.5-4B, zero-shot prompt), the Spearman correlation between their AUROCs on RAD-ChestCT and on CT-RATE is 0.89, and with 100 random pairs every encoder stays at 0.467--0.511 on RAD-ChestCT.

\section{Localization: derivation and every experiment}
\label{app:locate}
This section derives why distribution matching weights directions by variance (notation in Section~\ref{app:data}) and lists the zero-pair experiments.
Table~\ref{tab:supp_zeropair} compares all zero-pair methods on CT-RATE.
\begin{table*}[tp]
\centering
\caption{Zero-pair methods on CT-RATE. Result columns are numbered after the encoder column. Column 1 is the zero-pair Wasserstein Procrustes aligner~\cite{arXiv:2610.09411}: the minimum and maximum FOSCTTM over the text models of that encoder, with the number of text models. Column 2 is mini-vec2vec~\cite{arXiv:2510.02348} with Qwen3.5-4B at the middle layer: the mean FOSCTTM over seeds $\pm$ the sample standard deviation. Columns 3--6 use the zero-shot prompt on the validation patients. Column 3 is the mini-vec2vec prompt AUROC. Column 4 is the AUROC of ridge regression fitted on a random matching (random pseudo pairs). Column 5 uses no true pair: image anchors come from an external segmentation model (TotalSegmentator), report anchors come from zero-shot prompts on four findings, unpaired halves are matched by the Hungarian algorithm, and ridge regression is fitted. Column 6 is ridge regression on 100 true pairs with the same zero-shot prompt. The last column is the FOSCTTM of the anatomy-anchor map on the same held-out patients. The 15 rows cover 55 of the 56 encoder--text combinations. The 56th, VoxelFM block 11 with Qwen3.5-9B (middle layer), has FOSCTTM 0.490 (Table~\ref{tab:supp_locate_cells}).}
\label{tab:supp_zeropair}
\setlength{\tabcolsep}{3pt}
\footnotesize
\begin{tabular*}{\textwidth}{@{\extracolsep{\fill}}lccccccc@{}}
\toprule
Image encoder & \shortstack{Wasserstein--\\Procrustes\\FOSCTTM$\downarrow$\\min--max ($n$)} & \shortstack{mini-vec2vec\\FOSCTTM$\downarrow$} & \shortstack{mini-vec2vec\\AUROC$\uparrow$} & \shortstack{Random pairs\\AUROC$\uparrow$} & \shortstack{Anatomy anchors\\AUROC$\uparrow$} & \shortstack{100 true pairs\\AUROC$\uparrow$} & \shortstack{Anatomy anchors\\FOSCTTM$\downarrow$} \\
\midrule
VoxelFM & 0.453--0.519 (9) & 0.517{\scriptsize$\pm$0.010} & 0.487 & 0.430 & 0.671 & 0.698 & 0.396 \\
VoxelFM patch & 0.468--0.519 (9) & 0.499{\scriptsize$\pm$0.018} & 0.480 & 0.437 & 0.666 & 0.693 & 0.400 \\
CT-FM & 0.485--0.515 (9) & 0.497{\scriptsize$\pm$0.006} & 0.517 & 0.465 & 0.636 & 0.598 & 0.423 \\
V-JEPA 2-L & 0.447--0.474 (6) & 0.446{\scriptsize$\pm$0.011} & 0.599 & 0.491 & 0.665 & 0.659 & 0.398 \\
V-JEPA 2-H & 0.445--0.495 (2) & 0.522{\scriptsize$\pm$0.046} & 0.485 & 0.498 & 0.669 & 0.656 & 0.394 \\
V-JEPA 2-g & 0.440--0.503 (2) & 0.427{\scriptsize$\pm$0.005} & 0.625 & 0.479 & 0.666 & 0.664 & 0.396 \\
Wan VAE & 0.493--0.536 (2) & 0.530{\scriptsize$\pm$0.004} & 0.455 & 0.500 & 0.627 & 0.577 & 0.424 \\
\bottomrule
\end{tabular*}
\par\medskip
\setlength{\tabcolsep}{3pt}
\footnotesize
\begin{tabular*}{\textwidth}{@{\extracolsep{\fill}}lccccccc@{}}
\toprule
Image encoder & \shortstack{Wasserstein--\\Procrustes\\FOSCTTM$\downarrow$\\min--max ($n$)} & \shortstack{mini-vec2vec\\FOSCTTM$\downarrow$} & \shortstack{mini-vec2vec\\AUROC$\uparrow$} & \shortstack{Random pairs\\AUROC$\uparrow$} & \shortstack{Anatomy anchors\\AUROC$\uparrow$} & \shortstack{100 true pairs\\AUROC$\uparrow$} & \shortstack{Anatomy anchors\\FOSCTTM$\downarrow$} \\
\midrule
Wan DiT & 0.475--0.488 (2) & 0.489{\scriptsize$\pm$0.016} & 0.521 & 0.508 & 0.637 & 0.589 & 0.419 \\
LTX VAE & 0.492--0.510 (2) & 0.485{\scriptsize$\pm$0.022} & 0.526 & 0.495 & 0.554 & 0.541 & 0.466 \\
DINOv3 2D & 0.488--0.509 (2) & 0.512{\scriptsize$\pm$0.010} & 0.488 & 0.457 & 0.665 & 0.668 & 0.397 \\
DINOv3 2.5D & 0.497--0.503 (2) & 0.500{\scriptsize$\pm$0.005} & 0.501 & 0.459 & 0.669 & 0.671 & 0.398 \\
DALECT-0 & 0.486--0.504 (2) & 0.492{\scriptsize$\pm$0.020} & 0.497 & 0.479 & 0.653 & 0.648 & 0.415 \\
DALECT-1s & 0.479--0.506 (2) & 0.490{\scriptsize$\pm$0.021} & 0.523 & 0.490 & 0.651 & 0.601 & 0.414 \\
VideoMAE-B & 0.473--0.500 (2) & 0.523{\scriptsize$\pm$0.032} & 0.486 & 0.499 & 0.651 & 0.588 & 0.406 \\
DINOv2 2D & 0.492--0.511 (2) & 0.514{\scriptsize$\pm$0.012} & 0.472 & 0.466 & 0.670 & 0.661 & 0.395 \\
\bottomrule
\end{tabular*}
\end{table*}

\begin{figure}[tp]
\centering
\includegraphics[width=.6\textwidth]{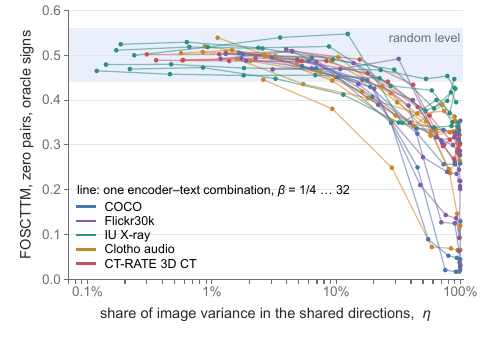}
\caption{FOSCTTM at zero pairs against the share $\eta$ of image variance in the shared directions (log scale), as Eq.~\eqref{eq:amp} scales the shared directions by $\beta\in\{1/4,1/2,\dots,32\}$ at unchanged canonical correlations. Line: one encoder--text combination; point: three-seed mean at one $\beta$. Signs are corrected with the correlations on the paired test set (oracle; on CT, the 2,025 validation patients), so the vertical axis measures localization alone. Band: random level (0.44--0.56).}
\label{fig:supp_phase}
\end{figure}

\paragraph{Why distribution matching weights directions by variance.}
Write a centered, unit-norm feature as $\mathbf{x}=\mathbf{x}^{\parallel}+\mathbf{x}^{\perp}$, with $\mathbf{x}^{\parallel}=\mathbf{U}_x\mathbf{U}_x^{\top}\mathbf{x}$
in the shared subspace and $\mathbf{x}^{\perp}$ in its orthogonal complement, so that $\mathbb{E}\lVert\mathbf{x}^{\parallel}\rVert^2=\eta$ and
$\mathbb{E}\lVert\mathbf{x}^{\perp}\rVert^2=1-\eta$. Split the text features $\mathbf{y}=\mathbf{y}^{\parallel}+\mathbf{y}^{\perp}$ in the same way, with $\mathbb{E}\lVert\mathbf{y}^{\parallel}\rVert^2=\eta_y$. For an orthogonal map $\mathbf{W}$ and a coupling $\pi$ of $N$ items, the Wasserstein Procrustes cost splits over the two
orthogonal parts, shared and private,
\begin{equation*}
\begin{aligned}
\textstyle\frac{1}{N}\sum_{i}\lVert\mathbf{W}^{\top}\mathbf{x}_i-\mathbf{y}_{\pi(i)}\rVert^2
&=\textstyle\frac{1}{N}\sum_{i}\lVert\mathbf{W}^{\top}\mathbf{x}^{\parallel}_i-\mathbf{y}^{\parallel}_{\pi(i)}\rVert^2\\
&\quad+\textstyle\frac{1}{N}\sum_{i}\lVert\mathbf{W}^{\top}\mathbf{x}^{\perp}_i-\mathbf{y}^{\perp}_{\pi(i)}\rVert^2
\end{aligned}
\end{equation*}
whenever $\mathbf{W}$ maps the shared subspace of one side onto that of the other. Only the cross terms of the two sums depend on $\mathbf{W}$ and $\pi$. By the Cauchy--Schwarz inequality, the shared cross term $\frac{2}{N}\sum_i\langle\mathbf{W}^{\top}\mathbf{x}^{\parallel}_i,\mathbf{y}^{\parallel}_{\pi(i)}\rangle$ is at most $2\sqrt{\eta\,\eta_y}$ in magnitude, and the private one at most $2\sqrt{(1-\eta)(1-\eta_y)}$.
The private directions have no partner on the other side, but the objective still lowers the second sum by matching
private directions of similar variance to each other. Recovering the true shared alignment lowers the first sum by at most $4\sqrt{\eta\,\eta_y}$. The Gromov--Wasserstein (GW) objective behaves in a similar way:
the inner product $\langle\mathbf{x}_i,\mathbf{x}_j\rangle=\langle\mathbf{x}^{\parallel}_i,\mathbf{x}^{\parallel}_j\rangle+\langle\mathbf{x}^{\perp}_i,\mathbf{x}^{\perp}_j\rangle$
enters the distance matrix additively, and the variance of each part across pairs is
$\sum_{l}\mu_l^2$ over the eigenvalues $\mu_l$ of that part. With the shared variance $\eta$ spread evenly over $k$ directions and the private variance $1-\eta$ over $m$ directions, the two parts contribute $\eta^2/k$ and $(1-\eta)^2/m$. With $\eta$ of a few percent, both objectives are therefore determined by
the private directions unless the shared variance is concentrated in far fewer directions than the private variance.
\paragraph{The intervention keeps the canonical correlations.} The map $\mathbf{A}=\mathbf{I}+(\beta-1)\mathbf{U}_x\mathbf{U}_x^{\top}$ has eigenvalues of $\beta$ (on the shared subspace) and 1, and thus is invertible for $\beta>0$. The canonical correlations of $(\mathbf{A}\mathbf{x},\mathbf{y})$ equal those of $(\mathbf{x},\mathbf{y})$, because the whitening of $\mathbf{A}\mathbf{x}$ removes $\mathbf{A}$. The share becomes $\tilde \eta=\beta^2\eta/(\beta^2\eta+1-\eta )$, so $\beta=8$ turns $\eta=0.05$ into $0.77$. We renormalize every vector after scaling, which is not a linear operation; we therefore measured the change on the data, where the top canonical correlation changes by at most 0.059. The control rescales $k$ random orthonormal directions chosen to hold the same share of variance.
\paragraph{Results.}
With the original features, the share acts as a threshold. We estimate the shared directions on the estimation split and measure $\eta$ on the pair split (1,945 patients on CT; the training split elsewhere). Of the 55 combinations with original features and a measured $\eta$ (31 on CT), 52 have $\eta <0.3$, and 51 of these stay at the random level (FOSCTTM 0.44--0.55); the exception, Clotho AST $\times$ MPNet, has $\eta=0.28$ and aligns (0.28). The same combinations together with their eight-fold scaled versions, which extend
$\eta$ to 0.97, give Spearman $-0.65$ ($p<10^{-10}$, the points of Fig.~\ref{fig:locate:a}), and every one of the 23 scaled combinations moves
closer to a perfect match than its unscaled version (median FOSCTTM gain 0.18).
To test whether the small CT training set explains the failure, we subsample COCO to the CT training-set size. The Wasserstein Procrustes aligner reaches FOSCTTM 0.012 on the authors' COCO features and 0.166 on the subsampled COCO, while every CT combination stays at 0.440--0.536. The COCO shrinking experiment uses DINOv2 token mean
$\times$ MPNet with $k=38$ shared directions.
Between $\eta=0.15$ and $0.45$ some combinations align and some do not. COCO aligns at $\eta=0.45$; Flickr30k fails at $\eta=0.38$. An independent team reports the same failure on Flickr30k. Kordes and Dorent~\cite{arXiv:2610.12781} test Gromov--Wasserstein, Wasserstein Procrustes, and Iterative Closest Point on their own frozen image and text encoders and find that all three collapse to near-random recall. This matches our Flickr30k result at $\eta=0.38$. Schnaus~\etal~\cite{arXiv:2610.09411} report the opposite on COCO, where we reproduce their zero-pair success with a FOSCTTM of $0.012$ on their features. Both results are consistent with the shares we measure on the two datasets.
For single combinations (Table~\ref{tab:supp_locate_cells}), eight-fold scaling moves FOSCTTM as follows: Flickr30k DINOv2 $\times$ MPNet from 0.51 to 0.13--0.16, Clotho AST $\times$
MPNet from 0.28 to 0.07, IU X-ray RAD-DINO $\times$ MPNet from 0.45 to 0.32--0.35.
\paragraph{Sweeping the scale.}
Figure~\ref{fig:supp_phase} scales the shared directions of 27 encoder--text combinations by eight values of $\beta$ from 1/4 to 32, with three seeds each. The combinations are the 20 scaled combinations of Fig.~\ref{fig:locate:a} outside CT, the three CT-RATE combinations of Fig.~\ref{fig:orient:a} and four COCO combinations. With oracle signs, all 51 points with $\eta<0.05$ have FOSCTTM 0.445 or higher, at the random level, and 102 of the 105 points with $\eta>0.5$ fall below the random level (FOSCTTM below 0.44). The level at large $\eta$ depends on the dataset. The best FOSCTTM per combination is 0.017--0.303 on COCO, 0.028--0.287 on Flickr30k, 0.065--0.321 on Clotho, 0.269--0.300 on CT-RATE and 0.339--0.427 on IU X-ray. A large share is thus necessary for zero-pair localization in every dataset, but it does not set the final level.
The shared signal exists without pairs: calibrated mutual nearest neighbors over the whole feature space reach 0.002--0.014 on CT against 0.09--0.25 on COCO (Table~\ref{tab:supp_loc}).
\paragraph{Whitening and independent component analysis (ICA).}
Whitening, ICA, and ICA on the most skewed components leave CT at FOSCTTM 0.42--0.52, and at 64 or more dimensions each breaks COCO (0.29 to 0.44--0.51; Table~\ref{tab:supp_pairfree}). Within one modality, a shared direction resembles a private one.
\paragraph{Zero-pair localization.}
Table~\ref{tab:supp_locate_cells} lists every encoder--text combination that we ran without pairs. Table~\ref{tab:supp_zeropair} compares all zero-pair methods that we ran on CT-RATE. The Wasserstein Procrustes aligner and mini-vec2vec stay at the random level in retrieval. Pseudo pairs from four anatomy anchors (on the image, the pleural and pericardial effusion volumes, the lung-nodule count, and the heart-to-lung volume ratio, which TotalSegmentator measures; on the report, the zero-shot prompts of these four findings) reach a prompt AUROC of 0.55--0.67, about the level of 100 true pairs, while random pseudo pairs stay at 0.43--0.51. An external measurement of shared anatomy can replace true pairs when locating a few shared directions. Table~\ref{tab:supp_pairfree} scales the directions that the anatomy measurements find.

\section{Orientation: derivation and the diagnosis}
\label{app:orient}
\begin{figure*}[tp]
\centering
\includegraphics[width=\textwidth]{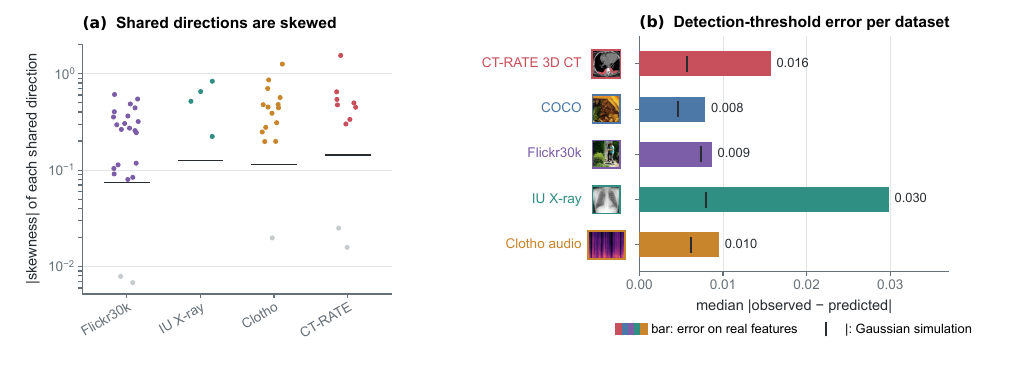}%
{\phantomsubcaption\label{fig:supp_extra:a}}{\phantomsubcaption\label{fig:supp_extra:b}}%
\caption{Skewness in the shared directions, and the detection threshold per dataset. \textbf{(a)}~$|$skewness$|$ of each image-side shared
direction (dot, log axis, color = dataset, sideways jitter; DINOv2 $\times$ Qwen3.5-4B on Flickr30k and IU X-ray, AST $\times$ Qwen3.5-4B on Clotho, VoxelFM block~11 $\times$ Qwen3.5-9B on CT) against the 99\,\% level of a Gaussian of the same size (line); gray dots lie below it.
\textbf{(b)}~Median error of the detection threshold per dataset on real features (bars) and on Gaussian data of the same $d$
and $n$ (ticks). The error is the absolute difference between the observed and predicted top three squared canonical correlations. Table~\ref{tab:supp_law_domains} gives the number of settings per dataset.}
\label{fig:supp_extra}
\end{figure*}

This section derives why a symmetric distribution leaves the sign of a shared direction undetermined (notation in Section~\ref{app:data}) and tests the skewness that fixes it.
Figs.~\ref{alg:zero} and~\ref{alg:few} show Algorithms 1 and 2, Fig.~\ref{fig:supp_extra} the skewness in the shared directions, and Table~\ref{tab:supp_orient_seeds} every seed.
\definecolor{hlloc}{HTML}{FFF0DD}
\definecolor{hlori}{HTML}{F8E1E4}
\definecolor{hlpair}{HTML}{EAF0F9}
\definecolor{hlrep}{HTML}{EEE7F4}
\newcommand{\hlb}[2]{{\setlength{\fboxsep}{1.2pt}\colorbox{#1}{\ensuremath{\displaystyle #2}}}}
\newcommand{\algobox}[3]{%
  \begin{minipage}[t]{#1}\small
  \hrule height .8pt\vspace{3pt}
  \textbf{Algorithm #2}\vspace{2pt}\hrule height .4pt\vspace{2pt}
  #3
  \vspace{1pt}\hrule height .8pt
  \end{minipage}}
\algrenewcommand\algorithmicrequire{\textbf{In:}}
\algrenewcommand\algorithmicensure{\textbf{Out:}}
\makeatletter
\newcounter{alghref}
\gdef\theHALG@line{\arabic{alghref}.\arabic{ALG@line}}
\AtBeginEnvironment{algorithmic}{\stepcounter{alghref}}
\makeatother

\newcommand{\algoboxh}[4]{%
  \begin{minipage}[t][#2][t]{#1}\small
  \hrule height .8pt\vspace{3pt}
  \textbf{Algorithm #3}\vspace{2pt}\hrule height .4pt\vspace{2pt}
  #4
  \vfill\hrule height .8pt
  \end{minipage}}
\newsavebox{\algsaveA}\newsavebox{\algsaveB}\newlength{\alghA}\newlength{\alghB}
\newcommand{\algAtitle}{1\ \ Zero pairs: locate, then orient}
\newcommand{\algBtitle}{2\ \ Few pairs: predict how many pairs are needed for alignment}
\newcommand{\algAbody}{%
\begin{algorithmic}[1]
\Require $\mathbf{X}\in\mathbb{R}^{N_x\times p}$, $\mathbf{Y}\in\mathbb{R}^{N_y\times q}$ unpaired; $t=10$; row sets $B,B'$ of size $M=\min(1{,}500,N_x,N_y)$
\Ensure map $\mathbf{W}$ from image to report space
\State \hlb{hlloc}{\mathbf{W}\gets\textsc{WassersteinProcrustes}(\mathbf{X},\mathbf{Y})} \hfill locate~\cite{arXiv:2610.09411}
\State $\mathbf{H}\gets\mathrm{rownorm}(\mathbf{X}_B\mathbf{W})$,\ $\mathbf{G}\gets\mathrm{rownorm}(\mathbf{Y}_{B'}-\bar{\mathbf{y}})$
\State $\mathbf{C}^X\gets\mathbf{1}-\mathbf{H}\mathbf{H}^{\top}$,\ \ $\mathbf{C}^Y\gets\mathbf{1}-\mathbf{G}\mathbf{G}^{\top}$
\State $\mathbf{Q}\gets$ top-$t$ right singular vectors of $\mathbf{H}-\bar{\mathbf{H}}$
\For{each sign pattern $\mathbf{s}\in\{\pm1\}^{t}$}
  \State $\mathbf{R}_{\mathbf{s}}\gets\mathbf{I}-2\,\mathbf{Q}_{\mathbf{s}}\mathbf{Q}_{\mathbf{s}}^{\top}$ \hfill $\mathbf{Q}_{\mathbf{s}}$: columns with $s_j=-1$
  \State $\pi_{\mathbf{s}}\gets\textsc{hungarian}(\mathbf{H}\mathbf{R}_{\mathbf{s}}\mathbf{G}^{\top})$
  \State \hlb{hlori}{L_{\mathbf{s}}\gets\tfrac{1}{M^2}\textstyle\sum_{a,b}\big(C^X_{ab}-C^Y_{\pi_{\mathbf{s}}(a)\pi_{\mathbf{s}}(b)}\big)^2}
\EndFor
\State \hlb{hlori}{\mathbf{s}^\star\gets\arg\min_{\mathbf{s}}L_{\mathbf{s}}} \hfill orient, no pairs
\State \Return $\mathbf{W}\mathbf{R}_{\mathbf{s}^\star}$
\end{algorithmic}}
\newcommand{\algBbody}{%
\begin{algorithmic}[1]
\Require encoders $\mathcal{E}$, language model; \hlb{hlpair}{n_0=200\ \text{pilot pairs}}; unpaired reports with report-derived labels
\Ensure encoder $e^\star$, map $\mathbf{W}$, pairs $n^\star_i$ each detected direction needs
\For{each encoder $e\in\mathcal{E}$}
  \State project each side onto its top-$d$ principal components (no pairs)
  \State $\lambda_{1..d}\gets$ squared canonical correlations of the pilot pairs
  \State $\hat k\gets\#\{\lambda_i>$ permutation null and $>\lambda_+\}$;\ \ $\hat r_i\gets\gamma^{-1}(\lambda_i)$
  \State $S_e\gets\sum_{i\le\hat k}\hat r_i$
\EndFor
\State $e^\star\gets\arg\max_e S_e$
\State \hlb{hlloc}{n^\star_i\gets d\,(1+1/\hat r_i)} for $e^\star$ \hfill Eq.~\eqref{eq:law}
\State \hlb{hlpair}{\mathbf{W}\gets\textsc{ridge}(\mathbf{X}_{n_0},\mathbf{Y}_{n_0})} of $e^\star$ on the pilot pairs
\State \hlb{hlrep}{g_\ell\gets\textsc{logistic}(\text{report features},\ \text{labels}_\ell)} \hfill text only
\State \Return $e^\star$, $\mathbf{W}$, $\{n^\star_i\}$; finding $\ell$ of a new scan $\mathbf{x}$: $g_\ell(\mathbf{W}^{\top}\mathbf{x})$
\end{algorithmic}}

\newcommand{\algboth}{%
\begin{figure*}[tp]
\sbox{\algsaveA}{\algobox{.485\textwidth}{\algAtitle}{\algAbody}}%
\sbox{\algsaveB}{\algobox{.485\textwidth}{\algBtitle}{\algBbody}}%
\setlength{\alghA}{\ht\algsaveA}\addtolength{\alghA}{\dp\algsaveA}%
\setlength{\alghB}{\ht\algsaveB}\addtolength{\alghB}{\dp\algsaveB}%
\ifdim\alghB>\alghA\setlength{\alghA}{\alghB}\fi
\begin{minipage}[t]{.485\textwidth}
\algoboxh{\linewidth}{\alghA}{\algAtitle}{\algAbody}
\caption{Zero-pair alignment. The distribution-matching aligner locates the shared subspace, and the
Gromov--Wasserstein (GW) loss of the hard matching selects the sign of each leading direction without paired supervision. Orange: locate. Red: orient. $t$ is the number of leading directions whose signs the search covers.}
\label{alg:zero}
\end{minipage}\hfill
\begin{minipage}[t]{.485\textwidth}
\algoboxh{\linewidth}{\alghA}{\algBtitle}{\algBbody}
\caption{Few-pair alignment. A 200-pair pilot estimates the canonical correlation strengths, gives the
pair count $n^\star_i$ of each detected direction, and fits the final linear map; the report classifier operates only in
the text space. Orange: detection threshold. Blue: steps that use pairs. Purple: report side only. $S_e$ is the sum of the estimated strengths $\hat r_i$ of the detected directions.}
\label{alg:few}
\end{minipage}
\end{figure*}}

\algboth

\begin{table*}[tp]
\centering
\caption{Each row is one seed: prompt AUROC of the map on the original features, after scaling the shared directions eight-fold, after GW signs (no pairs), and after oracle signs (needs pairs). The overlap block (seeds 0 to 4) is the fraction of the paired shared subspace the map captures. The paired map uses all training pairs; its seed changes only the centering means.}
\label{tab:supp_orient_seeds}
\setlength{\tabcolsep}{3pt}
\footnotesize
\begin{tabular*}{\textwidth}{@{\extracolsep{\fill}}llccccc@{}}
\toprule
\multicolumn{7}{l}{\emph{CT combinations, prompt AUROC per seed}}\\
\midrule
Image encoder & Text model & seed & unscaled$\uparrow$ & $\times 8$$\uparrow$ & GW signs$\uparrow$ & oracle signs (needs pairs)$\uparrow$ \\
\midrule
VoxelFM block 11 & Qwen3.5-9B (middle layer) & 2197670066 & 0.530 & 0.327 & 0.694 & 0.658 \\
VoxelFM block 11 & Qwen3.5-9B (middle layer) & 576165995 & 0.549 & 0.363 & 0.687 & 0.595 \\
VoxelFM block 11 & Qwen3.5-9B (middle layer) & 734796314 & 0.515 & 0.358 & 0.702 & 0.655 \\
VoxelFM patch & Qwen3.5-4B (middle layer) & 2197670066 & 0.475 & 0.285 & 0.700 & 0.695 \\
VoxelFM patch & Qwen3.5-4B (middle layer) & 576165995 & 0.488 & 0.679 & 0.679 & 0.679 \\
VoxelFM patch & Qwen3.5-4B (middle layer) & 734796314 & 0.466 & 0.677 & 0.670 & 0.672 \\
DINOv2 2D & Qwen3.5-4B (middle layer) & 2197670066 & 0.487 & 0.695 & 0.695 & 0.689 \\
DINOv2 2D & Qwen3.5-4B (middle layer) & 576165995 & 0.478 & 0.700 & 0.700 & 0.665 \\
DINOv2 2D & Qwen3.5-4B (middle layer) & 734796314 & 0.539 & 0.696 & 0.696 & 0.693 \\
\midrule
\multicolumn{7}{l}{\emph{Subspace overlap (seeds 0--4, separate from the rows above)}}\\
\midrule
Image encoder & scale & map & seed & overlap$\uparrow$ & AUROC$\uparrow$ & FOSCTTM$\downarrow$ \\
\midrule
VoxelFM patch & unscaled & zero-pair aligner & 0 & 0.035 & 0.473 & 0.500 \\
VoxelFM patch & unscaled & zero-pair aligner & 1 & 0.041 & 0.487 & 0.498 \\
VoxelFM patch & unscaled & zero-pair aligner & 2 & 0.044 & 0.490 & 0.485 \\
VoxelFM patch & unscaled & zero-pair aligner & 3 & 0.042 & 0.490 & 0.505 \\
VoxelFM patch & unscaled & zero-pair aligner & 4 & 0.044 & 0.472 & 0.491 \\
VoxelFM patch & unscaled & paired map & 0 & 0.149 & 0.684 & 0.285 \\
VoxelFM patch & unscaled & paired map & 1 & 0.149 & 0.684 & 0.288 \\
VoxelFM patch & unscaled & paired map & 2 & 0.149 & 0.684 & 0.289 \\
VoxelFM patch & unscaled & paired map & 3 & 0.149 & 0.684 & 0.287 \\
VoxelFM patch & unscaled & paired map & 4 & 0.149 & 0.684 & 0.287 \\
\bottomrule
\end{tabular*}
\par\medskip
\setlength{\tabcolsep}{3pt}
\footnotesize
\begin{tabular*}{\textwidth}{@{\extracolsep{\fill}}llccccc@{}}
\toprule
\multicolumn{7}{l}{\emph{Subspace overlap (seeds 0--4, separate from the rows above)}}\\
\midrule
Image encoder & scale & map & seed & overlap$\uparrow$ & AUROC$\uparrow$ & FOSCTTM$\downarrow$ \\
\midrule
VoxelFM patch & unscaled & entropic GW & 0 & 0.030 & 0.489 & 0.517 \\
VoxelFM patch & unscaled & entropic GW & 1 & 0.044 & 0.539 & 0.487 \\
VoxelFM patch & unscaled & entropic GW & 2 & 0.047 & 0.496 & 0.505 \\
VoxelFM patch & unscaled & entropic GW & 3 & 0.044 & 0.533 & 0.519 \\
VoxelFM patch & unscaled & entropic GW & 4 & 0.054 & 0.552 & 0.493 \\
VoxelFM patch & 8.0 & zero-pair aligner & 0 & 0.729 & 0.672 & 0.311 \\
VoxelFM patch & 8.0 & zero-pair aligner & 1 & 0.691 & 0.694 & 0.317 \\
VoxelFM patch & 8.0 & zero-pair aligner & 2 & 0.732 & 0.296 & 0.680 \\
VoxelFM patch & 8.0 & zero-pair aligner & 3 & 0.739 & 0.314 & 0.638 \\
VoxelFM patch & 8.0 & zero-pair aligner & 4 & 0.767 & 0.334 & 0.622 \\
VoxelFM patch & 8.0 & paired map & 0 & 0.759 & 0.696 & 0.169 \\
VoxelFM patch & 8.0 & paired map & 1 & 0.761 & 0.695 & 0.169 \\
VoxelFM patch & 8.0 & paired map & 2 & 0.759 & 0.695 & 0.169 \\
VoxelFM patch & 8.0 & paired map & 3 & 0.760 & 0.695 & 0.169 \\
VoxelFM patch & 8.0 & paired map & 4 & 0.759 & 0.696 & 0.169 \\
VoxelFM patch & 8.0 & entropic GW & 0 & 0.661 & 0.704 & 0.322 \\
VoxelFM patch & 8.0 & entropic GW & 1 & 0.670 & 0.697 & 0.286 \\
VoxelFM patch & 8.0 & entropic GW & 2 & 0.642 & 0.709 & 0.349 \\
VoxelFM patch & 8.0 & entropic GW & 3 & 0.700 & 0.700 & 0.319 \\
VoxelFM patch & 8.0 & entropic GW & 4 & 0.693 & 0.696 & 0.305 \\
\bottomrule
\end{tabular*}
\end{table*}

\paragraph{Why a symmetric distribution leaves the sign undetermined.} Consider $\mathbf{R}=\mathbf{I}-2\mathbf{b}\mathbf{b}^{\top}$ that reflects a shared direction $\mathbf{b}$. If the score $z=\mathbf{b}^{\top}\mathbf{x}$ has a symmetric distribution and is independent of the rest, then $\mathbf{R}\mathbf{x}$ has the same distribution as $\mathbf{x}$, and any objective that depends only on the two distributions (Wasserstein, Gromov--Wasserstein, any second-order statistic) has the same value at $\mathbf{W}$ and at $\mathbf{R}\mathbf{W}$: the sign of $\mathbf{b}$ is not identifiable.

A reflection keeps every pairwise
distance on its own side, so with a fixed coupling the GW loss cannot see it either. The loss sees the sign only through the
coupling, because Algorithm~1 recomputes the hard matching $\pi_{\mathbf{s}}$ from $\mathbf{H}\mathbf{R}_{\mathbf{s}}\mathbf{G}^{\top}$. If
$z$ is skewed, its third moment changes sign under $\mathbf{R}$. The long tail of the mapped cloud is then matched with the short
tail of the target, the assignment pairs items with dissimilar neighborhoods, and the distortion grows. Only odd-order moments change sign under $\mathbf{R}$; even-order moments are unchanged.
\paragraph{The same maps before and after scaling the shared directions.} Along the strongest shared direction, the mapped image score correlates with the true report score at $+0.70$ and $+0.71$ on the correct seeds, and at $-0.71$ to $-0.77$ on the mirrored ones. On VoxelFM patch, the sign correction lifts the mirrored seed from 0.285 to 0.700 and keeps the correct seeds correct. \looseness=-1
\paragraph{GW sign selection (Algorithm 1).} The subsample takes $M=\min(1{,}500,\text{size of each unpaired half})$ items per side without replacement ($M=972$ on CT). The $2^{10}=1{,}024$ candidates are the sign patterns of the top 10 right singular vectors of the centered mapped images; the 0-th candidate keeps all signs. The hard matching is the Hungarian assignment (\texttt{scipy.optimize.linear\_sum\_assignment}) that maximizes the cosine similarity between the reflected mapped images and the reports, and the loss is the mean squared difference of the two cosine-distance matrices under this permutation. Ties are broken by the earlier candidate. With the candidates run in parallel threads, selection takes 27\,s on one CPU node.
\paragraph{Oracle signs.} The oracle signs use the shared report directions that we estimate on the estimation split (10 directions for the final configuration). On the paired evaluation items of the zero-pair experiments, which we call the paired test set, we project the mapped image features and the report features onto each direction. We reflect the mapped image features along every direction whose two projections have a negative Pearson correlation. On CT, these items are the 2,025 validation patients, not the official test set of 3,039 volumes. The oracle sets each sign from one correlation and does not search for the best AUROC, so it is not an upper bound. On the final configuration, GW sign selection reaches a prompt AUROC of 0.687--0.702 over three seeds, against 0.595--0.658 for the oracle signs (Table~\ref{tab:supp_orient_seeds}). The best sign pattern of Tables~\ref{tab:domains} and~\ref{tab:supp_sign_methods} is a different reference. It has the lowest FOSCTTM on paired data among the $2^{10}$ candidates of Algorithm~1.
\paragraph{Skewness-removal control.} We randomly sample one sign per item and flip the item's image and its report together through the training means. The cross-covariance preserves its expectation and all odd moments are zeroed in expectation, so a sign cannot be read from the skewness. In the sample, the top canonical correlation changes from 0.880 to 0.863 and no shared correlation changes by more than 0.022. On the final configuration (eight-fold scaling), all three seeds are mirrored and all are corrected without the control, and the spread of the GW loss over sign patterns is 0.024; with it, one seed is mirrored, the spread falls to 0.006, and this seed is not corrected. Over both configurations, selection corrects 4 of 4 mirrored seeds without the control and 1 of 3 with it. Correction is measured by FOSCTTM.
\paragraph{Original features and other datasets.}
With the original features, even the best sign pattern (chosen with test pairs) has FOSCTTM 0.39--0.47 outside COCO, so signs matter only after localization succeeds.
On the scaled combinations of the three other datasets, GW sign selection over the top 10 directions corrects 4 of 18 mirrored
seeds (3 of 3 on IU X-ray), against 4 of 4 on CT (Table~\ref{tab:domains}). Table~\ref{tab:supp_sign_methods} compares enumeration with greedy search and the closed-form skew rule.
\paragraph{How Gaussian are the features.} Next, we project every feature set (all items of a dataset) onto its top 64 principal components. We compute the skewness and excess kurtosis of each component. The null distribution is 200 draws of standard normal data of the same size. A component is considered non-Gaussian if its excess kurtosis falls outside the central 99\,\% of the null distribution (CT: $[-0.159, 0.178]$), or if its $|$skewness$|$ exceeds the 99th percentile (CT: 0.082). Of the 29 image feature sets, 75--100\,\% of the components fall outside this kurtosis band; of the text feature sets, 41--95\,\% do. Of the 358 shared directions, 293 are significantly skewed on the image side and 278 on the report side. Mean pooling over tokens makes the features more Gaussian: the median excess kurtosis of Flickr30k DINOv2 falls from 0.72 (class token) to 0.24 (token mean), and of Clotho EAT from 1.87 to 0.68.

\section{How many pairs localization needs: derivation and tests}
\label{app:price}
This section derives the detection threshold from random-matrix theory (notation in Section~\ref{app:data}) and tests it on more datasets and 200-pair pilots. Table~\ref{tab:supp_law_domains} lists the detection threshold in every dataset and Table~\ref{tab:supp_pilot} lists every pilot.\looseness=-1
\begin{table}[tp]
\centering
\caption{Each row is one dataset: the median absolute error of the top three squared canonical correlations, the same median on Gaussian data of that size, and the number of settings in the median. The two medians are those of Fig.~\ref{fig:supp_extra:b}.}
\label{tab:supp_law_domains}
\setlength{\tabcolsep}{3pt}
\footnotesize
\begin{tabular*}{\linewidth}{@{\extracolsep{\fill}}lccc@{}}
\toprule
Dataset & $|$obs$-$pred$|$$\downarrow$ & Gaussian$\downarrow$ & settings \\
\midrule
CT-RATE 3D CT & 0.016 & 0.006 & 2,550 \\
COCO & 0.008 & 0.005 & 68 \\
Flickr30k & 0.009 & 0.007 & 102 \\
IU X-ray & 0.030 & 0.008 & 102 \\
Clotho audio & 0.010 & 0.006 & 34 \\
\bottomrule
\end{tabular*}
\end{table}

\begin{table}[tp]
\centering
\caption{Each row is one pilot (pilot size $n$ and seed) over the 15 image encoders: Spearman $\rho$ of the pilot score $S$ with AUROC at 1,600 pairs, the image encoder with the highest $S$, and how far that encoder's AUROC is from the best encoder. $S$ is the sum of the estimated strengths of the detected shared directions.}
\label{tab:supp_pilot}
\setlength{\tabcolsep}{3pt}
\footnotesize
\begin{tabular*}{\linewidth}{@{\extracolsep{\fill}}cccccc@{}}
\toprule
Pilot $n$ & seed & $\rho$$\uparrow$ & \shortstack{image encoder\\picked} & AUROC$\uparrow$ & gap$\downarrow$ \\
\midrule
200 & 0 & 0.968 & VoxelFM patch & 0.722 & 0.000 \\
200 & 1 & 0.806 & VoxelFM patch & 0.722 & 0.000 \\
200 & 2 & 0.724 & V-JEPA 2-g & 0.712 & 0.010 \\
200 & 3 & 0.786 & VoxelFM & 0.721 & 0.001 \\
200 & 4 & 0.846 & VoxelFM & 0.721 & 0.001 \\
400 & 0 & 0.911 & VoxelFM patch & 0.722 & 0.000 \\
400 & 1 & 0.843 & VoxelFM & 0.721 & 0.001 \\
400 & 2 & 0.639 & VoxelFM & 0.721 & 0.001 \\
400 & 3 & 0.718 & VoxelFM & 0.721 & 0.001 \\
400 & 4 & 0.725 & VoxelFM & 0.721 & 0.001 \\
\bottomrule
\end{tabular*}
\end{table}

\begin{table}[tp]
\centering
\caption{Each row is one score computed on the same pilot pairs as $S$ over the 15 image encoders, with 50 pilots per pilot size $n$: the median Spearman $\rho$ of the score with AUROC at 1,600 pairs, and how far the AUROC of the image encoder with the highest score is from the best encoder, averaged over the pilots. $\lambda_i$ are the sample squared canonical correlations of the pilot pairs on $d=64$ principal components. m$k$NN and CKA are computed on the pilot pairs as in Table~\ref{tab:supp_loc}.}
\label{tab:supp_pilot_scores}
\setlength{\tabcolsep}{3pt}
\footnotesize
\begin{tabular*}{\linewidth}{@{\extracolsep{\fill}}lcccc@{}}
\toprule
 & \multicolumn{2}{c}{Pilot $n=200$} & \multicolumn{2}{c}{Pilot $n=400$} \\
\cmidrule(lr){2-3}\cmidrule(l){4-5}
Score & $\rho$$\uparrow$ & gap$\downarrow$ & $\rho$$\uparrow$ & gap$\downarrow$ \\
\midrule
$S$ & 0.703 & 0.0039 & 0.816 & 0.0007 \\
$\lambda_1$ & 0.723 & 0.0025 & 0.909 & 0.0003 \\
$\lambda_1+\dots+\lambda_d$ & 0.641 & 0.0046 & 0.764 & 0.0008 \\
m$k$NN & 0.750 & 0.0039 & 0.758 & 0.0013 \\
CKA & 0.789 & 0.0087 & 0.811 & 0.0096 \\
\bottomrule
\end{tabular*}
\end{table}

\paragraph{The CT settings.} The 2,550 settings involve 15 encoders, five Qwen3.5 sizes, two layers, $d$ from 16 to 128, and $n$ from 100 to 1,600. We perform five 200-pair pilots, of which four select VoxelFM and the other selects V-JEPA 2-g, 0.0096 AUROC below the best one.
\paragraph{Other pilot scores.} Table~\ref{tab:supp_pilot_scores} compares $S$ with four other scores on the same pilot pairs, over 50 pilots per pilot size (seeds 0--49); the five pilots of Table~\ref{tab:supp_pilot} are seeds 0--4. CKA never picks the best image encoder. The top sample squared canonical correlation $\lambda_1$ ranks the encoders at least as well as $S$, because $S$ is computed from the same canonical correlations. But only $S$ also gives $n^\star_i$, the pairs that detect direction $i$.
\paragraph{The detection threshold in every dataset.} The median error of the largest squared canonical correlation is 0.008--0.010 in the seven other text models (Table~\ref{tab:supp_textmodels}). The median error of the top three squared canonical correlations is 0.024 in BIMCV-R. Table~\ref{tab:supp_law_domains} lists the other datasets and the Gaussian baseline. The detection threshold comes from random-matrix theory. Bao~\etal~\cite{bao2019canonical} study $n$ pairs of $p$- and $q$-dimensional Gaussian vectors with a finite number of population squared canonical correlations $r_1\ge r_2\ge\dots$, as $n,p,q\to\infty$ with $c_1=p/n$, $c_2=q/n$ fixed and $c_1+c_2<1$. Without signal, the sample squared canonical correlations fill a bulk with upper edge $\lambda_+=\big(\sqrt{c_1(1-c_2)}+\sqrt{c_2(1-c_1)}\big)^2$. A population squared canonical correlation $r_i$ produces an outlier above the bulk if and only if $r_i>\sqrt{c_1c_2/((1-c_1)(1-c_2))}$, and the outlier converges to $\gamma(r_i)=r_i\,(1-c_1+c_1/r_i)(1-c_2+c_2/r_i)$. We project both sides onto $d$ principal components, so $c_1=c_2=c=d/n$; then $\lambda_+=4c(1-c)$, the condition reads $r_i>c/(1-c)$, i.e.\ $n>d(1+1/r_i)$, and $\gamma(r)=r(1-c+c/r)^2$. Ma and Yang~\cite{ma2023sample} prove the same threshold for non-Gaussian data with independent entries or a factor structure and a finite fourth moment. To estimate each $r_i$ from the sample values $\lambda_1\ge\lambda_2\ge\dots$ on $n_0$ pairs, we take the leading directions in order and keep direction $i$ while $\lambda_i$ exceeds both $\lambda_+$ and the 95th percentile of the top sample value over 100 random re-pairings of the same items. Because the $\lambda_i$ are sorted and both thresholds are the same for every $i$, this equals the count in Algorithm~2. For a kept direction, $\gamma(r)=\lambda_i$ is the quadratic $(1-c)^2r^2+\big(2c(1-c)-\lambda_i\big)r+c^2=0$; we take its root with $r>c/(1-c)$ as $\hat r_i$.

\section{Ablations and robustness}
\label{app:ablation}
This section reports the ablations behind Sec.~\ref{sec:ablation} and the choice of the encoder and the text model.
\paragraph{A learned nonlinear map has lower AUROC.} Table~\ref{tab:ablation} ablates \textit{LightAlign}. We replace ridge regression with two MLPs trained with symmetric InfoNCE on the same $n$ pairs (architecture and cross-validation grid below). The MLP is 0.08 to 0.10 AUROC below ridge regression at every tested pair count, because with a few hundred pairs the extra capacity fits the pairs, not the shared subspace.

\paragraph{The report classifier gains more from additional pairs.} When we replace the report classifier with the zero-shot prompt and keep the same ridge regression, the AUROC improves only from 0.709 to 0.734 between 100 and 1,600 pairs, whereas the report classifier improves from 0.727 to 0.809 (Table~\ref{tab:ablation}). The prompt concentrates its weight along one axis, while the classifier spreads across the subspace (Section~\ref{app:payoff}). The report classifier outperforms the prompt on all 120 combinations of 15 CT encoders and 8 text models, with a bootstrap interval above zero for each combination (Table~\ref{tab:reader120}). The gain extends beyond disease burden: on the stratified AUROC, the report classifier reaches 0.696 against 0.535 for the prompt (VoxelFM, Qwen3.5-4B, 1,600 pairs, validation patients; Section~\ref{app:payoff}).

\begin{figure}[tp]
\centering
\includegraphics[width=0.62\textwidth]{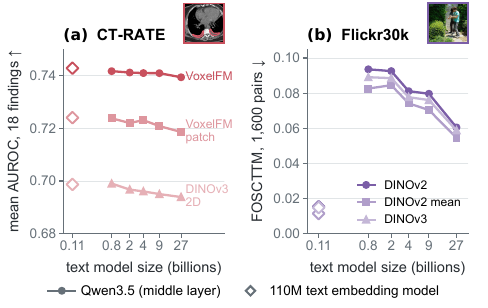}%
{\phantomsubcaption\label{fig:textscale:a}}{\phantomsubcaption\label{fig:textscale:b}}%
\caption{With ridge regression, the image encoder determines the result on CT, and the text model determines it on captions.
\textbf{(a)}~CT-RATE: mean AUROC of the report classifier over 18 findings on the 2,025 validation patients, averaged over seven pair counts (25 to 1,600) and five seeds. Each
line is one image encoder (dark to light: VoxelFM, VoxelFM patch, DINOv3 2D); filled marks: Qwen3.5 0.8B to 27B (middle layer); open diamond: CXR-BERT.
\textbf{(b)}~Flickr30k: FOSCTTM at 1,600 pairs, mean of three seeds; lines (dark to light): DINOv2, DINOv2 mean, DINOv3; open
diamond: MPNet. The x axis is the text model size in billions of parameters (log scale).}
\label{fig:textscale}
\end{figure}

\paragraph{Alignment outperforms direct supervision on the pairs.} We remove the map and train an image classifier on the $n$ paired
images with their report-derived labels. With the scikit-learn default $C=1$, its AUROC is 0.689--0.713 (Table~\ref{tab:supp_probe_cv}), lower than the map's. Table~\ref{tab:ablation} reports the tuned image classifier: we tune $C$ for each finding by stratified cross-validation on the $n$ pairs. The number of folds is 5, or the number of minority-class examples when this number is below 5. The grid has 13 log-spaced values of $C$ between $10^{-4}$ and $10^{2}$, and the selection criterion is the AUROC. A finding with fewer than two examples in the minority class keeps $C=1$. The tuned image classifier reaches 0.698, 0.735, and 0.767 at 100, 200, and 400 pairs. \textit{LightAlign} reaches 0.727, 0.754, and 0.782 at the same pair counts (Table~\ref{tab:supp_probe_cv}). At 1,600 pairs, \textit{LightAlign} reaches 0.809 and the tuned image classifier 0.806. \textit{LightAlign} thus outperforms direct supervision at every tested pair count below 5,994. The paired bootstrap interval of the AUROC gain of \textit{LightAlign} over the tuned image classifier is $[+0.026,+0.033]$ at 100 pairs, $[+0.017,+0.022]$ at 200 pairs, $[+0.013,+0.017]$ at 400 pairs, and $[+0.001,+0.005]$ at 1,600 pairs.The tuned image classifier catches up only when it is trained on the labels of all 5,994 training images (0.824 for both). \textit{LightAlign} trains no image classifier: two frozen encoders and one ridge regression reach this AUROC. The additional reports do not explain this difference: when the report classifier is trained on the reports of the $n$ pairs only, \textit{LightAlign} reaches 0.734, 0.785, and 0.811 at 100, 400, and 1,600 pairs (0.727, 0.782, and 0.809 with all 5,994 training reports; Table~\ref{tab:supp_reports}). With 200 pairs, the AUROC stays between 0.754 and 0.759 when the classifier is fitted on 200 to 5,994 reports. All three variants of Table~\ref{tab:ablation} perform worse than \textit{LightAlign} at every pair count. The rest of this section shows how the score variance divides between the image encoder and the text model in each dataset, and how the scale of the language model changes the result (Fig.~\ref{fig:textscale}).
\begin{table*}[tp]
\centering
\caption{The report classifier of \textit{LightAlign} with fewer reports on CT-RATE. Entries: mean AUROC over 18 findings on the test set (3,002 scans), averaged over five seeds. Frozen VoxelFM and Qwen3.5-9B features. Top: the classifier is fitted on all 5,994 training reports or on the reports of the $n$ pairs only. Bottom: 200 pairs, and the classifier is fitted on $m$ reports, which include the reports of the pairs. CT-CLIP reaches 0.733.}
\label{tab:supp_reports}
\setlength{\tabcolsep}{3pt}
\footnotesize
\begin{tabular*}{\linewidth}{@{\extracolsep{\fill}}lccccccc@{}}
\toprule
Image--report pairs $n$ & 25 & 50 & 100 & 200 & 400 & 800 & 1,600 \\
\midrule
All 5,994 reports & 0.643 & 0.687 & 0.727 & 0.754 & 0.782 & 0.796 & 0.809 \\
Reports of the $n$ pairs only & 0.652 & 0.694 & 0.734 & 0.759 & 0.785 & 0.798 & 0.811 \\
\midrule
Reports $m$ ($n=200$) & 200 & 500 & 1,000 & 2,000 & 4,000 & 5,994 & \\
\midrule
$m$ reports & 0.759 & 0.759 & 0.759 & 0.759 & 0.756 & 0.754 & \\
\bottomrule
\end{tabular*}
\end{table*}

\paragraph{Where model capacity belongs.}
We decompose the score variance over the encoder $\times$ text-model grid of each dataset. In the two medical datasets,
the image encoder determines the result. For the report classifier, it explains 99.0\,\% of the CT AUROC variance (text model 0.7\,\%) and 95\,\% of the
IU X-ray FOSCTTM variance at 1,600 pairs. On IU X-ray, the radiograph encoder RAD-DINO outperforms DINOv2 by 0.06 FOSCTTM,
whereas all text models lie within 0.02. In contrast, on natural captions the text model determines the result. It
explains 98.5\,\% of the variance on Flickr30k and 87\,\% on COCO, and on Clotho its share grows from 14\,\% at 100 pairs to
45\,\% at 1,600. The effect of language-model scale also differs across datasets (Fig.~\ref{fig:textscale}). Increasing Qwen3.5 from 0.8B to 27B improves captions (FOSCTTM on Flickr30k 0.094 to 0.061, on COCO 0.056 to 0.036, on Clotho 0.096 to 0.073)
but not reports (CT AUROC averaged over pair counts 0.742 to 0.739, IU X-ray 0.787 to 0.760). In every dataset, a
110M-parameter text embedding model matches or outperforms the 27B model (Table~\ref{tab:supp_textmodels}), and CXR-BERT achieves the best CT mean (0.743). In medical imaging, model capacity therefore belongs in the image encoder. This agrees with Balestriero and LeCun~\cite{arXiv:2610.39302}, whose 109M-parameter text joint-embedding predictive architecture (JEPA) aligns with image encoders better than every language model they evaluated, up to 72B parameters.
\paragraph{Encoder or text model, per dataset.} The FOSCTTM per combination at 1,600 pairs and the AUROC of RAD-DINO on IU X-ray are shown in Table~\ref{tab:supp_textmodels}. On Flickr30k, the text model explains 93.8\,\% of the variance (100 pairs), 93.7\,\% (400), and 98.5\,\% (1,600). On IU X-ray, the encoder explains 97.2\,\%, 98.2\,\%, and 95.1\,\% (FOSCTTM), and 86.9\,\%, 85.3\,\%, and 94.5\,\% (AUROC). On Clotho, the encoder explains 85.1\,\%, 72.6\,\%, and 53.8\,\%.
\paragraph{Choice of the encoder block and the text layer.} In Table~\ref{tab:supp_layerchoice}, we compare VoxelFM blocks 10, 11, and 12, as well as the middle and last layers of Qwen3.5-9B, on the 2,025 validation patients, using ridge regression and the report classifier. Block 11 with the middle layer is best at every pair count from 50 to 1,600 (0.812 at 1,600 pairs, against 0.805 for blocks 10 and 12, and 0.801 for block 12 with the last layer); only at 25 pairs is block 12 0.009 higher. VoxelFM patch features are below the class-token features at every pair count. We extend the comparison to blocks 2 to 12 and all language-model layers with Qwen3.5-4B and the prompt in Tables~\ref{tab:supp_encdepth} and~\ref{tab:supp_layers}.
\begin{table*}[t]
\centering
\caption{Choice of the VoxelFM block and the language-model layer. Each entry is the mean AUROC over 18 findings on the 2,025 validation patients (mean over five seeds $\pm$ sample standard deviation), with ridge regression on $n$ image--report pairs and the report classifier; the text model is Qwen3.5-9B. Columns: number of pairs $n$. Shaded: the final configuration. Bold: best per column.}
\label{tab:supp_layerchoice}
\setlength{\tabcolsep}{3pt}
\scriptsize
\begin{tabular*}{\textwidth}{@{\extracolsep{\fill}}llccccccc@{}}
\toprule
Image encoder & Qwen3.5-9B layer & 25$\uparrow$ & 50$\uparrow$ & 100$\uparrow$ & 200$\uparrow$ & 400$\uparrow$ & 800$\uparrow$ & 1,600$\uparrow$ \\
\midrule
VoxelFM block 10 & middle & 0.627{\scriptsize$\pm$0.005} & 0.680{\scriptsize$\pm$0.019} & 0.726{\scriptsize$\pm$0.008} & 0.755{\scriptsize$\pm$0.004} & 0.778{\scriptsize$\pm$0.003} & 0.795{\scriptsize$\pm$0.001} & 0.805{\scriptsize$\pm$0.001} \\
\rowcolor{oursbg} VoxelFM block 11 & middle & 0.629{\scriptsize$\pm$0.006} & \textbf{0.686}{\scriptsize$\pm$0.015} & \textbf{0.729}{\scriptsize$\pm$0.009} & \textbf{0.760}{\scriptsize$\pm$0.004} & \textbf{0.783}{\scriptsize$\pm$0.003} & \textbf{0.801}{\scriptsize$\pm$0.001} & \textbf{0.812}{\scriptsize$\pm$0.001} \\
VoxelFM block 12 (last) & middle & \textbf{0.638}{\scriptsize$\pm$0.006} & \textbf{0.686}{\scriptsize$\pm$0.014} & 0.726{\scriptsize$\pm$0.008} & 0.756{\scriptsize$\pm$0.004} & 0.779{\scriptsize$\pm$0.003} & 0.797{\scriptsize$\pm$0.001} & 0.805{\scriptsize$\pm$0.004} \\
VoxelFM block 12 (last) & last & \textbf{0.638}{\scriptsize$\pm$0.018} & 0.685{\scriptsize$\pm$0.013} & 0.722{\scriptsize$\pm$0.010} & 0.751{\scriptsize$\pm$0.004} & 0.776{\scriptsize$\pm$0.003} & 0.789{\scriptsize$\pm$0.001} & 0.801{\scriptsize$\pm$0.001} \\
VoxelFM patch & middle & 0.599{\scriptsize$\pm$0.013} & 0.660{\scriptsize$\pm$0.013} & 0.708{\scriptsize$\pm$0.010} & 0.738{\scriptsize$\pm$0.002} & 0.763{\scriptsize$\pm$0.003} & 0.782{\scriptsize$\pm$0.001} & 0.796{\scriptsize$\pm$0.001} \\
VoxelFM patch & last & 0.598{\scriptsize$\pm$0.015} & 0.653{\scriptsize$\pm$0.018} & 0.703{\scriptsize$\pm$0.010} & 0.730{\scriptsize$\pm$0.006} & 0.763{\scriptsize$\pm$0.003} & 0.782{\scriptsize$\pm$0.001} & 0.796{\scriptsize$\pm$0.001} \\
\bottomrule
\end{tabular*}
\end{table*}

\paragraph{Results without the block and layer choice.} We select block 11 of VoxelFM and the middle layer of Qwen3.5-9B on the 2,025 validation patients (Table~\ref{tab:supp_layerchoice}). These patients belong to the training set, and the test set is disjoint from them. Table~\ref{tab:supp_no_selection} repeats the test-set evaluation without the block choice, without the layer choice, and without both. The same pipeline then uses the final output of VoxelFM (block 12), the last layer of Qwen3.5-9B, or both. All four configurations use VoxelFM, which four of five 200-pair pilots select (Section~\ref{app:price}). With 200 pairs, all four configurations outperform CT-CLIP (0.733): the AUROC is 0.748--0.754, and every bootstrap interval of the difference lies above zero. The configuration without the block choice and without the layer choice reaches 0.748 at 200 pairs, with the interval $[+0.008,+0.021]$. It reaches 0.806 at 1,600 pairs and 0.824 with all 5,994 pairs. At each pair count, the four configurations differ by at most 0.007 AUROC. At 200, 1,600, and 5,994 pairs, the gain over CT-CLIP therefore does not depend on the block choice or the layer choice.
\paragraph{Language scaling on CT.} The input layer of the language model, where LLaVA-style models~\cite{arXiv:2304.08485} insert image features, is its worst layer (0.665 at 100 pairs against 0.707 at the middle layer). Zero pairs also fail at the last layer: Qwen3.5-4B last layer with all 15 CT encoders gives FOSCTTM 0.450--0.532. With VoxelFM (final block), Qwen3.5 0.8B / 2B / 4B / 9B / 27B give 0.7417 / 0.7411 / 0.7409 / 0.7409 /
0.7393 averaged over pair counts; at 1,600 pairs 0.7975 to 0.8075. With CXR-BERT, the two VoxelFM variants reach 0.743 and 0.724, and DINOv3 2D reaches 0.699.
The strength of the top shared direction depends on the encoder (0.335 to 0.787 across encoders, for every Qwen3.5 size); over VoxelFM blocks 2 to 11 the top
correlation rises from 0.505 to 0.783.
\paragraph{MLP ablation details.} The MLP ablation (Sec.~\ref{sec:ablation}) replaces ridge regression with two MLPs, one per modality, each with one hidden layer of 1,024 units and a 512-dimensional output. Both are trained with symmetric InfoNCE on the same $n$ pairs. We use five-fold cross-validation to select the weight decay ($10^{-4}$ to $10$) and the training length (5 to 1,600 epochs). For every pair count, the selected weight decay is 3 or 10, the two strongest regularization values.

\section{Qualitative results}
\label{app:qual}
\begin{figure*}[tp]
\centering
\includegraphics[width=\textwidth]{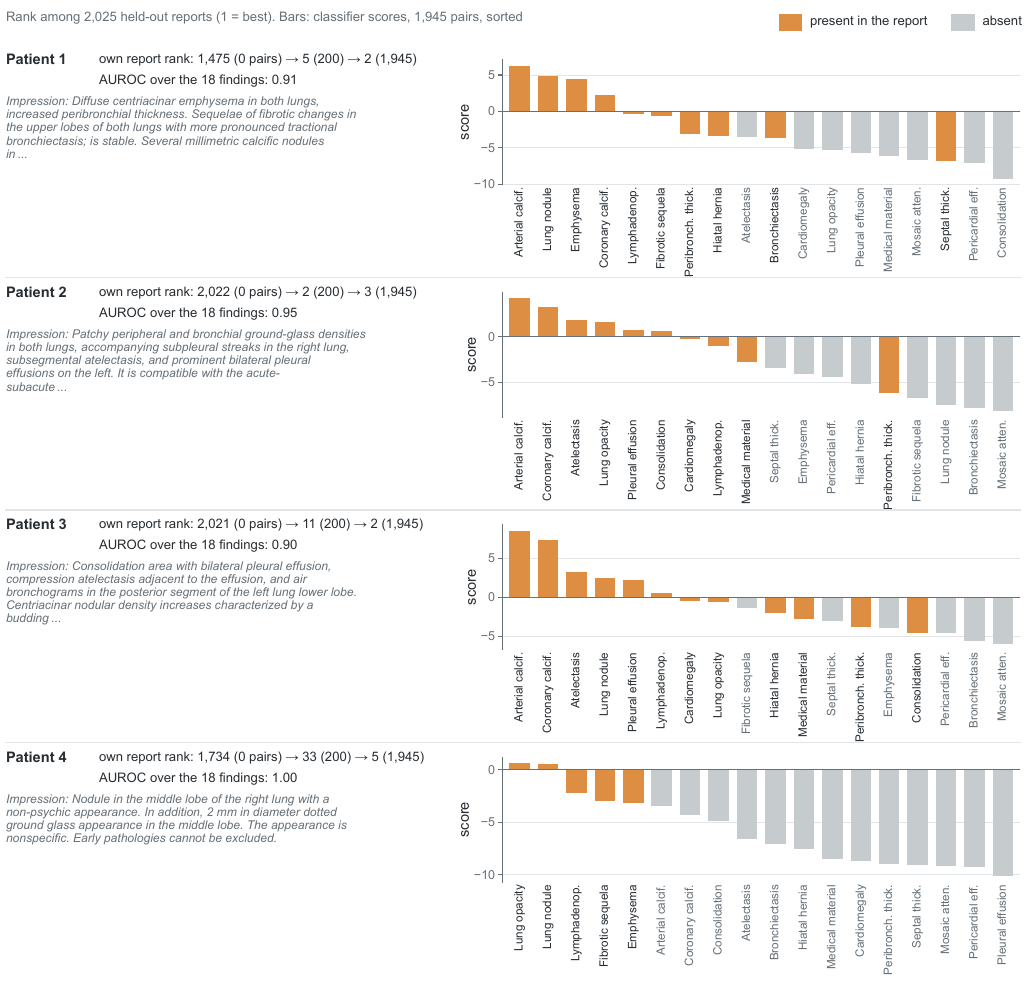}
\caption{Qualitative examples. Four validation patients of CT-RATE that pass the selection rule of
Section~\ref{app:qual}. Left: the impression of the patient's own report, the rank of that report among 2,025
validation reports for zero pairs, 200 pairs, and 1,945 pairs, and the AUROC over the patient's 18 findings.
Right: report classifier scores of the 18 findings on the 1,945-pair map, sorted by score; orange bars and dark
labels mark findings that the report states, gray bars and labels mark findings that it does not state; y axis: report-classifier score. Report text is quoted verbatim from CT-RATE, including its translation errors.}
\label{fig:qual}
\end{figure*}

The examples illustrate the few-pair results of Table~\ref{tab:main}.
Figure~\ref{fig:qual} shows four validation patients of CT-RATE for the final configuration (VoxelFM block 11 and
Qwen3.5-9B, middle layer), with ridge regression fitted on all 1,945 pairs of the pair split. We selected the
examples with a rule that we fixed before we looked at them: at least two findings in the report, the patient's own report
among the top 5 of the 2,025 validation reports with the full map, at rank 500 or worse with zero pairs, and an AUROC
of at least 0.9 over the patient's 18 findings. Eight patients pass this rule, and we show four of them.

Without pairs, the patient's own report has a random rank,
for example rank 2,022 of 2,025 for patient 2. With 200 pairs, it is in the top 33 for all four
patients. With the full map, it is in the top 5. The report classifier then reads the findings from the
mapped scan: when the 18 findings are sorted by their score, the findings that the report states rank near the top. Over all 2,025 validation patients, the median AUROC over a patient's findings is 0.81 with 200 pairs and 0.87 with the full map; Table~\ref{tab:main} measures each finding across patients.

\section{Implementation and reproducibility}
\label{app:impl}
This section gives the code, the seeds, and the hardware behind the experiments.
\paragraph{Zero-pair aligner.} We use the code that the authors of~\cite{arXiv:2610.09411} released
(repository \texttt{unpaired-rosetta}, commit \texttt{fdfad84}) with the authors' defaults: 30 clusters, 30
restarts, batch size 10,000, 100 iterations, and the authors' quadratic assignment problem (QAP) solver pin.
\paragraph{Seeds.} In the zero-pair analyses the aligner uses the authors' seed derivation (first three of five seeds from 42); the zero-pair rows of the AUROC tables pass seeds 0--4 to the same aligner, and our ridge regression
pair draws and pilots use seeds 0--4. Data splits hash the patient or item id. Bootstrap intervals use 1,000
patient resamples, one fixed set per evaluation subset for all combinations. Two runs of the aligner with the same seed on different node types can give maps that differ by up to 0.06 FOSCTTM before sign selection, because the aligner amplifies floating-point differences. The CT rows of Table~\ref{tab:supp_sign_methods} and three COCO combinations of Table~\ref{tab:coco_signs} come from such a second run, so they differ from Tables~\ref{tab:coco_signs} and~\ref{tab:supp_locate_cells}, respectively, for the same seeds. For DINOv2 token mean with Qwen3.5-4B on COCO, only the second run exists, so this combination has values in Table~\ref{tab:coco_signs} and a dash in Table~\ref{tab:supp_locate_cells}.
\paragraph{Few-pair baselines.} All baselines use the same frozen features, pairs, and seeds as \textit{LightAlign}. We score each baseline with the zero-shot prompt and with the report classifier of \textit{LightAlign} (Table~\ref{tab:supp_fewpair}). SOTAlign~\cite{arXiv:2602.23353} uses the authors' code (\texttt{ExplainableML/SOTAlign}, commit \texttt{64c89ee}) and configuration: a linear head to 1,024 dimensions per modality, the semi-supervised SigLIP loss on the $n$ pairs and the KLOT loss with the CCA teacher (regularization 0.1) on the unpaired remainder, Lion with learning rate $10^{-4}$ and weight decay $10^{-5}$, 2,000 steps with 10\,\% warm-up and a cosine schedule, logit scale 20 and bias $-10$, loss weights 1 and $10^{-3}$, and Sinkhorn $\varepsilon$ of 0.05 (student) and 0.01 (teacher) with 100 iterations. We use the last step, because a checkpoint choice on extra pairs would use more than the $n$ pairs. For Hadgi~\etal~\cite{arXiv:2503.05283}, we reimplement their map: the regularized CCA of SOTAlign on the $n$ pairs to $k=50$ dimensions ($k=n-1$ for $n\le50$), then an affine map from the projected images to the projected reports, fitted by least squares on the same pairs. Procrustes, CCA, and ASIF-style anchors use the same pairs without unpaired data.
\paragraph{Data.} All datasets are public and de-identified, and we use them under their licenses. We collect no new human data.
\paragraph{Hardware and runtime.} CT features were extracted on A100 GPUs (VoxelFM: 4.4 A100-hours for the 5,994 training volumes, 1.7 for the 3,039 validation volumes). Image, audio, and text features were extracted on A40 and A100 GPUs and a 96\,GB RTX PRO GPU partition. The aligner, sign selection, ridge regression, and classifiers are CPU-based: on four cores, ridge regression on 200 pairs takes 0.3\,s, the 18 report classifiers 59\,s, and scoring 3,039 scans 0.8\,s, about one minute in total. At 2.6 A100-seconds per volume, the 200 pilot volumes take about 9 A100-minutes. In our experiments, the full evaluation at 100 pairs on CT-RATE (report classifier and patient bootstrap included) takes 43 minutes with SOTAlign on one A40 GPU and 16 CPU cores, and 7 minutes with the ridge regression of \textit{LightAlign} on 16 CPU cores.

\clearpage
\setlength{\LTcapwidth}{\textwidth}\setlength{\LTpre}{2pt}\setlength{\LTpost}{6pt}
\section{Complete result tables}
\label{app:tables}
This section collects the complete tables cited above.
{\footnotesize\setlength{\tabcolsep}{3pt}\setlength{\LTleft}{0pt plus 1fil}\setlength{\LTright}{0pt plus 1fil}
\begin{longtable}{@{\extracolsep{\fill}}rlcccc@{}}
\caption{Each entry is the mean AUROC (top) and the 95\,\% patient-clustered bootstrap interval of that value minus CT-CLIP (bottom), for one pair count, one evaluation subset, and one scoring method. The model is VoxelFM block 11 with Qwen3.5-9B (middle layer). The subsets are all validation volumes (all), volumes that contain the chest (chest), the first reconstruction of each study (study), and reconstruction 1 of studies whose identifier ends in \_a, chest only (patient $a_1$). The second block averages the 16 findings that remain after dropping Lymphadenopathy and Medical material.}\label{tab:supp_payoff_ct}\\
\toprule
\multicolumn{6}{l}{\emph{mean AUROC, 18 findings}}\\
\midrule
$n$ & scoring method & all (3039)$\uparrow$ & chest (3002)$\uparrow$ & study (1564)$\uparrow$ & patient $a_1$ (1293)$\uparrow$ \\
\midrule
0 & prompt & 0.500{\scriptsize[-0.247,\,-0.215]} & 0.500{\scriptsize[-0.250,\,-0.217]} & 0.496{\scriptsize[-0.254,\,-0.217]} & 0.495{\scriptsize[-0.263,\,-0.224]} \\
0 & classifier & 0.478{\scriptsize[-0.265,\,-0.241]} & 0.478{\scriptsize[-0.267,\,-0.242]} & 0.473{\scriptsize[-0.270,\,-0.245]} & 0.474{\scriptsize[-0.280,\,-0.252]} \\
25 & prompt & 0.654{\scriptsize[-0.084,\,-0.071]} & 0.655{\scriptsize[-0.085,\,-0.071]} & 0.658{\scriptsize[-0.080,\,-0.066]} & 0.664{\scriptsize[-0.083,\,-0.068]} \\
25 & classifier & 0.641{\scriptsize[-0.097,\,-0.082]} & 0.643{\scriptsize[-0.098,\,-0.082]} & 0.645{\scriptsize[-0.094,\,-0.077]} & 0.649{\scriptsize[-0.099,\,-0.081]} \\
50 & prompt & 0.682{\scriptsize[-0.056,\,-0.042]} & 0.684{\scriptsize[-0.056,\,-0.043]} & 0.681{\scriptsize[-0.057,\,-0.043]} & 0.688{\scriptsize[-0.059,\,-0.044]} \\
\bottomrule
\end{longtable}}
{\footnotesize\setlength{\tabcolsep}{3pt}\setlength{\LTleft}{0pt plus 1fil}\setlength{\LTright}{0pt plus 1fil}
\addtocounter{table}{-1}\begin{longtable}{@{\extracolsep{\fill}}rlcccc@{}}
\toprule
\multicolumn{6}{l}{\emph{mean AUROC, 16 findings}}\\
\midrule
$n$ & scoring method & all (3039)$\uparrow$ & chest (3002)$\uparrow$ & study (1564)$\uparrow$ & patient $a_1$ (1293)$\uparrow$ \\
\midrule
0 & prompt & 0.499{\scriptsize[-0.252,\,-0.218]} & 0.499{\scriptsize[-0.254,\,-0.220]} & 0.495{\scriptsize[-0.257,\,-0.218]} & 0.495{\scriptsize[-0.264,\,-0.226]} \\
0 & classifier & 0.479{\scriptsize[-0.266,\,-0.241]} & 0.480{\scriptsize[-0.267,\,-0.243]} & 0.475{\scriptsize[-0.271,\,-0.244]} & 0.476{\scriptsize[-0.278,\,-0.250]} \\
25 & prompt & 0.655{\scriptsize[-0.086,\,-0.072]} & 0.656{\scriptsize[-0.087,\,-0.073]} & 0.659{\scriptsize[-0.081,\,-0.067]} & 0.663{\scriptsize[-0.084,\,-0.069]} \\
25 & classifier & 0.647{\scriptsize[-0.095,\,-0.079]} & 0.648{\scriptsize[-0.096,\,-0.079]} & 0.650{\scriptsize[-0.091,\,-0.074]} & 0.653{\scriptsize[-0.095,\,-0.077]} \\
50 & prompt & 0.684{\scriptsize[-0.057,\,-0.043]} & 0.685{\scriptsize[-0.057,\,-0.043]} & 0.682{\scriptsize[-0.058,\,-0.044]} & 0.687{\scriptsize[-0.060,\,-0.044]} \\
50 & classifier & 0.691{\scriptsize[-0.051,\,-0.034]} & 0.693{\scriptsize[-0.051,\,-0.034]} & 0.690{\scriptsize[-0.051,\,-0.034]} & 0.694{\scriptsize[-0.054,\,-0.036]} \\
\bottomrule
\end{longtable}}

{\footnotesize\setlength{\tabcolsep}{3pt}\setlength{\LTleft}{0pt plus 1fil}\setlength{\LTright}{0pt plus 1fil}
\begin{longtable}{@{\extracolsep{\fill}}lrcccccc@{}}
\caption{Each entry is the mean AUROC of the five competition findings (top, with its study-level bootstrap 95\,\% interval) or one finding's AUROC, on the official CheXpert test set. Rows are one pair count; the largest count is every training pair. Rows above 19,858 pairs use the same report classifier, fitted on the same 19,858 base reports.}\label{tab:supp_chexpert}\\
\toprule
split & $n$ & \multicolumn{6}{c}{prompt} \\
\cmidrule(lr){3-8}
 & & AUROC$\uparrow$ & Atelectasis$\uparrow$ & Cardiomegaly$\uparrow$ & Consolidation$\uparrow$ & Edema$\uparrow$ & \shortstack{Pleural\\Effusion}$\uparrow$ \\
\midrule
test & 25 & 0.609{\scriptsize[0.589,\,0.630]} & 0.603 & 0.589 & 0.606 & 0.638 & 0.610 \\
test & 50 & 0.656{\scriptsize[0.633,\,0.678]} & 0.647 & 0.614 & 0.675 & 0.670 & 0.672 \\
test & 100 & 0.693{\scriptsize[0.668,\,0.717]} & 0.689 & 0.652 & 0.702 & 0.721 & 0.701 \\
test & 200 & 0.725{\scriptsize[0.699,\,0.752]} & 0.724 & 0.684 & 0.731 & 0.757 & 0.731 \\
test & 400 & 0.746{\scriptsize[0.719,\,0.773]} & 0.750 & 0.692 & 0.751 & 0.773 & 0.761 \\
test & 800 & 0.763{\scriptsize[0.736,\,0.790]} & 0.772 & 0.713 & 0.754 & 0.800 & 0.777 \\
test & 1,600 & 0.781{\scriptsize[0.756,\,0.806]} & 0.782 & 0.729 & 0.779 & 0.816 & 0.796 \\
test & 3,200 & 0.789{\scriptsize[0.765,\,0.814]} & 0.785 & 0.736 & 0.796 & 0.824 & 0.805 \\
test & 6,400 & 0.794{\scriptsize[0.769,\,0.819]} & 0.787 & 0.745 & 0.804 & 0.830 & 0.803 \\
test & 12,800 & 0.798{\scriptsize[0.773,\,0.823]} & 0.788 & 0.746 & 0.814 & 0.835 & 0.807 \\
test & 19,858 & 0.800{\scriptsize[0.775,\,0.825]} & 0.787 & 0.748 & 0.817 & 0.838 & 0.808 \\
test & 40,000 & 0.803{\scriptsize[0.779,\,0.827]} & 0.793 & 0.746 & 0.819 & 0.840 & 0.818 \\
test & 80,000 & 0.807{\scriptsize[0.782,\,0.830]} & 0.798 & 0.746 & 0.820 & 0.843 & 0.826 \\
test & 124,769 & 0.808{\scriptsize[0.784,\,0.831]} & 0.801 & 0.746 & 0.822 & 0.843 & 0.827 \\
\bottomrule
\end{longtable}}
{\footnotesize\setlength{\tabcolsep}{3pt}\setlength{\LTleft}{0pt plus 1fil}\setlength{\LTright}{0pt plus 1fil}
\addtocounter{table}{-1}\begin{longtable}{@{\extracolsep{\fill}}lrcccccc@{}}
\toprule
split & $n$ & \multicolumn{6}{c}{classifier} \\
\cmidrule(lr){3-8}
 & & AUROC$\uparrow$ & Atelectasis$\uparrow$ & Cardiomegaly$\uparrow$ & Consolidation$\uparrow$ & Edema$\uparrow$ & \shortstack{Pleural\\Effusion}$\uparrow$ \\
\midrule
test & 25 & 0.591{\scriptsize[0.575,\,0.607]} & 0.527 & 0.587 & 0.610 & 0.565 & 0.667 \\
test & 50 & 0.623{\scriptsize[0.606,\,0.641]} & 0.513 & 0.624 & 0.616 & 0.637 & 0.724 \\
test & 100 & 0.678{\scriptsize[0.661,\,0.695]} & 0.567 & 0.657 & 0.676 & 0.721 & 0.769 \\
test & 200 & 0.717{\scriptsize[0.698,\,0.735]} & 0.592 & 0.702 & 0.716 & 0.766 & 0.807 \\
test & 400 & 0.750{\scriptsize[0.731,\,0.770]} & 0.626 & 0.742 & 0.764 & 0.785 & 0.833 \\
test & 800 & 0.794{\scriptsize[0.777,\,0.812]} & 0.676 & 0.795 & 0.823 & 0.815 & 0.863 \\
test & 1,600 & 0.815{\scriptsize[0.798,\,0.833]} & 0.707 & 0.817 & 0.828 & 0.841 & 0.882 \\
test & 3,200 & 0.836{\scriptsize[0.819,\,0.854]} & 0.741 & 0.836 & 0.859 & 0.850 & 0.896 \\
test & 6,400 & 0.849{\scriptsize[0.831,\,0.866]} & 0.764 & 0.842 & 0.865 & 0.864 & 0.911 \\
test & 12,800 & 0.859{\scriptsize[0.842,\,0.876]} & 0.784 & 0.847 & 0.877 & 0.871 & 0.917 \\
test & 19,858 & 0.862{\scriptsize[0.845,\,0.879]} & 0.785 & 0.853 & 0.885 & 0.869 & 0.921 \\
test & 40,000 & 0.869{\scriptsize[0.852,\,0.887]} & 0.794 & 0.857 & 0.889 & 0.879 & 0.927 \\
test & 80,000 & 0.874{\scriptsize[0.856,\,0.891]} & 0.803 & 0.860 & 0.893 & 0.880 & 0.931 \\
test & 124,769 & 0.876{\scriptsize[0.859,\,0.892]} & 0.806 & 0.861 & 0.898 & 0.882 & 0.932 \\
\bottomrule
\end{longtable}}

{\scriptsize\setlength{\tabcolsep}{3pt}\setlength{\LTleft}{0pt plus 1fil}\setlength{\LTright}{0pt plus 1fil}
\begin{longtable}{@{\extracolsep{\fill}}lcccccccc@{}}
\caption{Each entry is the mean over five seeds $\pm$ the sample standard deviation, for one image encoder and one pair count. The text model is Qwen3.5-4B (middle layer). At 0 pairs the map is the zero-pair aligner, and for more pairs it is ridge regression. The upper block is the prompt AUROC over 18 findings; the lower block is image-to-report FOSCTTM of the same map.}\label{tab:supp_encoders_ct}\\
\toprule
\multicolumn{9}{l}{\emph{Mean AUROC, prompt}}\\
\midrule
Image encoder & 0$\uparrow$ & 25$\uparrow$ & 50$\uparrow$ & 100$\uparrow$ & 200$\uparrow$ & 400$\uparrow$ & 800$\uparrow$ & 1,600$\uparrow$ \\
\midrule
VoxelFM & 0.537{\scriptsize$\pm$0.016} & 0.668{\scriptsize$\pm$0.023} & 0.685{\scriptsize$\pm$0.002} & 0.707{\scriptsize$\pm$0.009} & 0.715{\scriptsize$\pm$0.003} & 0.718{\scriptsize$\pm$0.003} & 0.720{\scriptsize$\pm$0.002} & 0.721{\scriptsize$\pm$0.002} \\
VoxelFM patch & 0.522{\scriptsize$\pm$0.028} & 0.642{\scriptsize$\pm$0.023} & 0.663{\scriptsize$\pm$0.023} & 0.700{\scriptsize$\pm$0.009} & 0.713{\scriptsize$\pm$0.003} & 0.716{\scriptsize$\pm$0.003} & 0.720{\scriptsize$\pm$0.003} & 0.722{\scriptsize$\pm$0.000} \\
CT-FM & 0.456{\scriptsize$\pm$0.021} & 0.556{\scriptsize$\pm$0.045} & 0.574{\scriptsize$\pm$0.021} & 0.599{\scriptsize$\pm$0.028} & 0.632{\scriptsize$\pm$0.013} & 0.659{\scriptsize$\pm$0.007} & 0.671{\scriptsize$\pm$0.002} & 0.690{\scriptsize$\pm$0.001} \\
V-JEPA 2-L & 0.587{\scriptsize$\pm$0.018} & 0.604{\scriptsize$\pm$0.037} & 0.634{\scriptsize$\pm$0.014} & 0.654{\scriptsize$\pm$0.019} & 0.678{\scriptsize$\pm$0.015} & 0.689{\scriptsize$\pm$0.007} & 0.703{\scriptsize$\pm$0.003} & 0.708{\scriptsize$\pm$0.001} \\
V-JEPA 2-H & 0.599{\scriptsize$\pm$0.027} & 0.639{\scriptsize$\pm$0.024} & 0.651{\scriptsize$\pm$0.012} & 0.663{\scriptsize$\pm$0.018} & 0.680{\scriptsize$\pm$0.006} & 0.696{\scriptsize$\pm$0.005} & 0.707{\scriptsize$\pm$0.002} & 0.713{\scriptsize$\pm$0.001} \\
\bottomrule
\end{longtable}}

{\footnotesize\setlength{\tabcolsep}{3pt}\setlength{\LTleft}{0pt plus 1fil}\setlength{\LTright}{0pt plus 1fil}
\begin{longtable}{@{\extracolsep{\fill}}lllccc@{}}
\caption{Each row is one encoder--text combination with no paired training: $\eta$ is the share of image variance in the shared directions, and unscaled FOSCTTM is the mean over seeds $\pm$ the sample standard deviation. The $\times 8$ column is that FOSCTTM after scaling the directions eight-fold, with oracle signs where the oracle run exists and GW signs (no pairs) otherwise. A dash means that no run exists for the combination.}\label{tab:supp_locate_cells}\\
\toprule
Dataset & Image encoder & Text model & $\eta$ & unscaled FOSCTTM$\downarrow$ & $\times 8$ FOSCTTM$\downarrow$ \\
\midrule
COCO & DINOv2 & MPNet & 0.277 & 0.463{\scriptsize$\pm$0.091} & -- \\
COCO & DINOv2 & Qwen3.5-4B (middle layer) & 0.195 & 0.475{\scriptsize$\pm$0.035} & -- \\
COCO & DINOv2 mean & MPNet & 0.451 & 0.293{\scriptsize$\pm$0.029} & -- \\
COCO & DINOv2 mean & Qwen3.5-4B (middle layer) & 0.371 & -- & -- \\
COCO & DINOv3 & MPNet & 0.266 & -- & -- \\
COCO & DINOv3 & Qwen3.5-4B (middle layer) & 0.188 & 0.475{\scriptsize$\pm$0.020} & -- \\
\bottomrule
\end{longtable}}
{\footnotesize\setlength{\tabcolsep}{3pt}\setlength{\LTleft}{0pt plus 1fil}\setlength{\LTright}{0pt plus 1fil}
\addtocounter{table}{-1}\begin{longtable}{@{\extracolsep{\fill}}lllccc@{}}
\toprule
Dataset & Image encoder & Text model & $\eta$ & unscaled FOSCTTM$\downarrow$ & $\times 8$ FOSCTTM$\downarrow$ \\
\midrule
Clotho audio & Dasheng & MPNet & 0.093{\scriptsize$\pm$0.000} & 0.530{\scriptsize$\pm$0.014} & 0.237{\scriptsize$\pm$0.093} \\
Clotho audio & Dasheng & Qwen3.5-4B (middle layer) & 0.091{\scriptsize$\pm$0.000} & 0.518{\scriptsize$\pm$0.029} & 0.332{\scriptsize$\pm$0.013} \\
Clotho audio & EAT & MPNet & 0.054{\scriptsize$\pm$0.000} & 0.504{\scriptsize$\pm$0.035} & 0.375{\scriptsize$\pm$0.021} \\
Clotho audio & EAT & Qwen3.5-4B (middle layer) & 0.049{\scriptsize$\pm$0.000} & 0.485{\scriptsize$\pm$0.018} & 0.390{\scriptsize$\pm$0.029} \\
Clotho audio & EAT mean & MPNet & 0.098{\scriptsize$\pm$0.000} & 0.512{\scriptsize$\pm$0.022} & 0.329{\scriptsize$\pm$0.031} \\
Clotho audio & EAT mean & Qwen3.5-4B (middle layer) & 0.104{\scriptsize$\pm$0.000} & 0.525{\scriptsize$\pm$0.025} & 0.329{\scriptsize$\pm$0.021} \\
CT-RATE 3D CT & CT-FM & MPNet & 0.013 & 0.485{\scriptsize$\pm$0.015} & -- \\
\bottomrule
\end{longtable}}
{\footnotesize\setlength{\tabcolsep}{3pt}\setlength{\LTleft}{0pt plus 1fil}\setlength{\LTright}{0pt plus 1fil}
\addtocounter{table}{-1}\begin{longtable}{@{\extracolsep{\fill}}lllccc@{}}
\toprule
Dataset & Image encoder & Text model & $\eta$ & unscaled FOSCTTM$\downarrow$ & $\times 8$ FOSCTTM$\downarrow$ \\
\midrule
CT-RATE 3D CT & DINOv3 2.5D & MPNet & 0.047 & 0.503{\scriptsize$\pm$0.015} & -- \\
CT-RATE 3D CT & DINOv3 2.5D & Qwen3.5-4B (middle layer) & 0.062 & 0.497{\scriptsize$\pm$0.005} & -- \\
CT-RATE 3D CT & DINOv3 2D & MPNet & 0.054 & 0.488{\scriptsize$\pm$0.012} & -- \\
CT-RATE 3D CT & DINOv3 2D & Qwen3.5-4B (middle layer) & 0.068 & 0.509{\scriptsize$\pm$0.022} & -- \\
CT-RATE 3D CT & LTX VAE & MPNet & 0.005 & 0.492{\scriptsize$\pm$0.016} & -- \\
CT-RATE 3D CT & LTX VAE & Qwen3.5-4B (middle layer) & 0.007 & 0.510{\scriptsize$\pm$0.024} & -- \\
CT-RATE 3D CT & VideoMAE-B & MPNet & 0.017 & 0.500{\scriptsize$\pm$0.005} & -- \\
\bottomrule
\end{longtable}}
{\footnotesize\setlength{\tabcolsep}{3pt}\setlength{\LTleft}{0pt plus 1fil}\setlength{\LTright}{0pt plus 1fil}
\addtocounter{table}{-1}\begin{longtable}{@{\extracolsep{\fill}}lllccc@{}}
\toprule
Dataset & Image encoder & Text model & $\eta$ & unscaled FOSCTTM$\downarrow$ & $\times 8$ FOSCTTM$\downarrow$ \\
\midrule
CT-RATE 3D CT & VoxelFM & Gemma 4 31B (middle layer) & -- & 0.468{\scriptsize$\pm$0.007} & -- \\
CT-RATE 3D CT & VoxelFM & Granite 4.2 8B (middle layer) & -- & 0.518{\scriptsize$\pm$0.025} & -- \\
CT-RATE 3D CT & VoxelFM & MedGemma 1.5 4B (middle layer) & -- & 0.501{\scriptsize$\pm$0.014} & -- \\
CT-RATE 3D CT & VoxelFM & Ministral 3 14B (middle layer) & -- & 0.503{\scriptsize$\pm$0.012} & -- \\
CT-RATE 3D CT & VoxelFM & OLMo 3.1 32B (middle layer) & -- & 0.509{\scriptsize$\pm$0.004} & -- \\
CT-RATE 3D CT & VoxelFM & Qwen3-VL-Emb. 8B (middle layer) & -- & 0.453{\scriptsize$\pm$0.010} & -- \\
CT-RATE 3D CT & VoxelFM block 11 & Qwen3.5-9B (middle layer) & 0.083 & 0.490{\scriptsize$\pm$0.010} & 0.322{\scriptsize$\pm$0.018} \\
\bottomrule
\end{longtable}}

{\scriptsize\setlength{\tabcolsep}{3pt}\setlength{\LTleft}{0pt plus 1fil}\setlength{\LTright}{0pt plus 1fil}
\begin{longtable}{@{\extracolsep{\fill}}llcccccc@{}}
\caption{Each row is one seed of one encoder--text combination. FOSCTTM and prompt AUROC are given before the sign selection and after it. The selection evaluates all $2^{10}$ sign patterns of the top ten directions with the Gromov--Wasserstein loss and uses no pairs; flipped is the number of the ten signs it changes (0 keeps the original signs). The upper block is COCO without scaling, where the zero-pair map works on some combinations; the lower block is CT-RATE after the eight-fold scaling of the shared directions.}\label{tab:coco_signs}\\
\toprule
Image encoder & Text model & seed & \multicolumn{2}{c}{FOSCTTM$\downarrow$} & \multicolumn{2}{c}{prompt AUROC$\uparrow$} & flipped \\
\cmidrule(lr){4-5}\cmidrule(lr){6-7}
 & & & before & after & before & after & \\
\midrule
\multicolumn{8}{l}{\emph{COCO, no scaling}}\\
\midrule
DINOv2 class token & MPNet & 734796314 & 0.478 & 0.478 & 0.453 & 0.453 & 0 \\
DINOv2 class token & MPNet & 576165995 & 0.561 & 0.563 & 0.448 & 0.448 & 0 \\
DINOv2 class token & MPNet & 2197670066 & 0.309 & 0.310 & 0.589 & 0.589 & 0 \\
DINOv2 class token & Qwen3.5-4B & 734796314 & 0.417 & 0.418 & 0.477 & 0.477 & 0 \\
\bottomrule
\end{longtable}}

{\footnotesize\setlength{\tabcolsep}{3pt}\setlength{\LTleft}{0pt plus 1fil}\setlength{\LTright}{0pt plus 1fil}
\begin{longtable}{@{\extracolsep{\fill}}rlcccccccc@{}}
\caption{Each CT entry is the weighted F1 or the accuracy at the corner of the receiver operating characteristic (ROC), mean over the 18 findings, for one pair count, one scoring method, and one evaluation subset (VoxelFM block 11 with Qwen3.5-9B, middle layer). Each CheXpert entry is the Matthews correlation coefficient (MCC), positive-class F1, or accuracy (top, with its bootstrap interval) or one finding's MCC, at the validation-set threshold (RAD-DINO with Qwen3.5-4B, middle layer; five seeds, one seed for the row with all pairs).}\label{tab:supp_thresh}\\
\toprule
\multicolumn{10}{l}{\emph{CT-RATE, ROC-corner threshold}}\\
\midrule
$n$ & scoring method & \multicolumn{2}{c}{all (3039)} & \multicolumn{2}{c}{chest (3002)} & \multicolumn{2}{c}{study (1564)} & \multicolumn{2}{c}{patient $a_1$ (1293)} \\
\cmidrule(lr){3-4}\cmidrule(lr){5-6}\cmidrule(lr){7-8}\cmidrule(lr){9-10}
 &  & F1$\uparrow$ & Accuracy$\uparrow$ & F1$\uparrow$ & Accuracy$\uparrow$ & F1$\uparrow$ & Accuracy$\uparrow$ & F1$\uparrow$ & Accuracy$\uparrow$ \\
\midrule
50 & prompt & 0.671 & 0.625 & 0.673 & 0.626 & 0.672 & 0.626 & 0.681 & 0.631 \\
50 & classifier & 0.685 & 0.641 & 0.685 & 0.642 & 0.682 & 0.637 & 0.690 & 0.643 \\
100 & prompt & 0.691 & 0.648 & 0.693 & 0.651 & 0.693 & 0.650 & 0.704 & 0.659 \\
100 & classifier & 0.707 & 0.667 & 0.709 & 0.670 & 0.713 & 0.673 & 0.719 & 0.677 \\
\bottomrule
\end{longtable}}
{\footnotesize\setlength{\tabcolsep}{3pt}\setlength{\LTleft}{0pt plus 1fil}\setlength{\LTright}{0pt plus 1fil}
\addtocounter{table}{-1}\begin{longtable}{@{\extracolsep{\fill}}llccc@{}}
\toprule
\multicolumn{5}{l}{\emph{CheXpert, threshold fit on the validation set}}\\
\midrule
$n$ & scoring method & MCC$\uparrow$ & F1$\uparrow$ & Accuracy$\uparrow$ \\
\midrule
50 & prompt & 0.149{\scriptsize[0.123,\,0.174]} & 0.319{\scriptsize[0.293,\,0.343]} & 0.602{\scriptsize[0.588,\,0.616]} \\
50 & classifier & 0.150{\scriptsize[0.129,\,0.170]} & 0.265{\scriptsize[0.242,\,0.284]} & 0.695{\scriptsize[0.682,\,0.710]} \\
100 & prompt & 0.206{\scriptsize[0.178,\,0.233]} & 0.386{\scriptsize[0.355,\,0.412]} & 0.618{\scriptsize[0.602,\,0.635]} \\
100 & classifier & 0.191{\scriptsize[0.169,\,0.213]} & 0.344{\scriptsize[0.319,\,0.365]} & 0.691{\scriptsize[0.680,\,0.703]} \\
200 & prompt & 0.243{\scriptsize[0.211,\,0.277]} & 0.406{\scriptsize[0.374,\,0.434]} & 0.636{\scriptsize[0.616,\,0.657]} \\
200 & classifier & 0.247{\scriptsize[0.221,\,0.271]} & 0.390{\scriptsize[0.360,\,0.414]} & 0.693{\scriptsize[0.680,\,0.706]} \\
400 & prompt & 0.272{\scriptsize[0.239,\,0.307]} & 0.433{\scriptsize[0.401,\,0.460]} & 0.641{\scriptsize[0.618,\,0.663]} \\
400 & classifier & 0.293{\scriptsize[0.264,\,0.320]} & 0.418{\scriptsize[0.388,\,0.443]} & 0.705{\scriptsize[0.691,\,0.719]} \\
1,600 & prompt & 0.328{\scriptsize[0.290,\,0.366]} & 0.467{\scriptsize[0.434,\,0.497]} & 0.693{\scriptsize[0.669,\,0.716]} \\
1,600 & classifier & 0.371{\scriptsize[0.339,\,0.399]} & 0.478{\scriptsize[0.446,\,0.504]} & 0.734{\scriptsize[0.716,\,0.751]} \\
all (19,858) & prompt & 0.346{\scriptsize[0.305,\,0.389]} & 0.481{\scriptsize[0.444,\,0.512]} & 0.699{\scriptsize[0.674,\,0.726]} \\
all (19,858) & classifier & 0.431{\scriptsize[0.400,\,0.464]} & 0.513{\scriptsize[0.483,\,0.542]} & 0.743{\scriptsize[0.724,\,0.763]} \\
\bottomrule
\end{longtable}}
{\footnotesize\setlength{\tabcolsep}{3pt}\setlength{\LTleft}{0pt plus 1fil}\setlength{\LTright}{0pt plus 1fil}
\addtocounter{table}{-1}\begin{longtable}{@{\extracolsep{\fill}}llccccc@{}}
\toprule
\multicolumn{7}{l}{\emph{CheXpert, per-finding MCC}}\\
\midrule
$n$ & scoring method & Atelectasis$\uparrow$ & Cardiomegaly$\uparrow$ & Consolidation$\uparrow$ & Edema$\uparrow$ & \shortstack{Pleural\\Effusion}$\uparrow$ \\
\midrule
50 & prompt & 0.180 & 0.078 & 0.120 & 0.177 & 0.190 \\
50 & classifier & 0.067 & 0.157 & 0.069 & 0.176 & 0.282 \\
100 & prompt & 0.251 & 0.185 & 0.141 & 0.221 & 0.232 \\
100 & classifier & 0.084 & 0.201 & 0.112 & 0.198 & 0.359 \\
200 & prompt & 0.300 & 0.212 & 0.153 & 0.270 & 0.280 \\
200 & classifier & 0.114 & 0.285 & 0.183 & 0.280 & 0.372 \\
400 & prompt & 0.355 & 0.248 & 0.173 & 0.266 & 0.318 \\
400 & classifier & 0.158 & 0.365 & 0.180 & 0.330 & 0.431 \\
1,600 & prompt & 0.435 & 0.294 & 0.196 & 0.342 & 0.375 \\
1,600 & classifier & 0.266 & 0.399 & 0.256 & 0.416 & 0.516 \\
all (19,858) & prompt & 0.387 & 0.289 & 0.244 & 0.420 & 0.392 \\
all (19,858) & classifier & 0.394 & 0.431 & 0.254 & 0.458 & 0.617 \\
\bottomrule
\end{longtable}}

{\scriptsize\setlength{\tabcolsep}{3pt}\setlength{\LTleft}{0pt plus 1fil}\setlength{\LTright}{0pt plus 1fil}
\begin{longtable}{@{\extracolsep{\fill}}lcccccccc@{}}
\caption{Each entry is the mean over five seeds $\pm$ the sample standard deviation for one VoxelFM block and one pair count. The text model is Qwen3.5-4B (middle layer). At 0 pairs the map is the zero-pair aligner, and for more pairs it is ridge regression. The row blocks are prompt AUROC, FOSCTTM, and AUROC after removing the axis common to all findings.}\label{tab:supp_encdepth}\\
\toprule
\multicolumn{9}{l}{\emph{Prompt AUROC}}\\
\midrule
Block & 0$\uparrow$ & 25$\uparrow$ & 50$\uparrow$ & 100$\uparrow$ & 200$\uparrow$ & 400$\uparrow$ & 800$\uparrow$ & 1,600$\uparrow$ \\
\midrule
2 & 0.481{\scriptsize$\pm$0.004} & 0.510{\scriptsize$\pm$0.015} & 0.518{\scriptsize$\pm$0.011} & 0.528{\scriptsize$\pm$0.011} & 0.544{\scriptsize$\pm$0.020} & 0.616{\scriptsize$\pm$0.012} & 0.638{\scriptsize$\pm$0.005} & 0.666{\scriptsize$\pm$0.002} \\
4 & 0.486{\scriptsize$\pm$0.009} & 0.517{\scriptsize$\pm$0.024} & 0.526{\scriptsize$\pm$0.010} & 0.547{\scriptsize$\pm$0.040} & 0.633{\scriptsize$\pm$0.045} & 0.670{\scriptsize$\pm$0.008} & 0.692{\scriptsize$\pm$0.003} & 0.697{\scriptsize$\pm$0.001} \\
6 & 0.490{\scriptsize$\pm$0.021} & 0.566{\scriptsize$\pm$0.039} & 0.579{\scriptsize$\pm$0.026} & 0.620{\scriptsize$\pm$0.030} & 0.679{\scriptsize$\pm$0.009} & 0.687{\scriptsize$\pm$0.006} & 0.699{\scriptsize$\pm$0.001} & 0.705{\scriptsize$\pm$0.001} \\
8 & 0.501{\scriptsize$\pm$0.015} & 0.603{\scriptsize$\pm$0.036} & 0.616{\scriptsize$\pm$0.026} & 0.675{\scriptsize$\pm$0.008} & 0.697{\scriptsize$\pm$0.004} & 0.705{\scriptsize$\pm$0.005} & 0.710{\scriptsize$\pm$0.003} & 0.718{\scriptsize$\pm$0.001} \\
10 & 0.506{\scriptsize$\pm$0.020} & 0.647{\scriptsize$\pm$0.029} & 0.676{\scriptsize$\pm$0.009} & 0.703{\scriptsize$\pm$0.007} & 0.715{\scriptsize$\pm$0.002} & 0.717{\scriptsize$\pm$0.002} & 0.721{\scriptsize$\pm$0.003} & 0.723{\scriptsize$\pm$0.001} \\
\bottomrule
\end{longtable}}

{\footnotesize\setlength{\tabcolsep}{3pt}\setlength{\LTleft}{0pt plus 1fil}\setlength{\LTright}{0pt plus 1fil}
\begin{longtable}{@{\extracolsep{\fill}}lccccccc@{}}
\caption{At 1,600 pairs, each CT entry is the prompt AUROC of one image encoder and one text model. Each entry for the other datasets is the mean over three seeds $\pm$ the sample standard deviation of FOSCTTM, or of the report-classifier AUROC on a labeled dataset (COCO: presence of the 80 instance categories). The threshold row is the median absolute error of the largest squared canonical correlation.}\label{tab:supp_textmodels}\\
\toprule
\multicolumn{8}{l}{\emph{CT prompt AUROC at $n=1,600$}}\\
\midrule
Image encoder & CXR-BERT$\uparrow$ & Gemma 4 31B$\uparrow$ & Granite 4.2 8B$\uparrow$ & \shortstack{MedGemma\\1.5 4B}$\uparrow$ & \shortstack{Ministral\\3 14B}$\uparrow$ & OLMo 3.1 32B$\uparrow$ & \shortstack{Qwen3-VL-Emb.\\8B}$\uparrow$ \\
\midrule
VoxelFM & 0.710 & 0.685 & 0.578 & 0.718 & 0.506 & 0.741 & 0.664 \\
VoxelFM patch & 0.708 & 0.686 & 0.572 & 0.721 & 0.508 & 0.740 & 0.665 \\
CT-FM & 0.674 & 0.645 & 0.570 & 0.687 & 0.526 & 0.697 & 0.634 \\
V-JEPA 2-L & 0.698 & 0.673 & 0.571 & 0.705 & 0.508 & 0.717 & 0.653 \\
V-JEPA 2-H & 0.701 & 0.682 & 0.578 & 0.711 & 0.508 & 0.724 & 0.660 \\
V-JEPA 2-g & 0.702 & 0.680 & 0.584 & 0.710 & 0.515 & 0.723 & 0.667 \\
Wan VAE & 0.680 & 0.644 & 0.572 & 0.684 & 0.513 & 0.695 & 0.638 \\
Wan DiT & 0.688 & 0.655 & 0.569 & 0.693 & 0.509 & 0.707 & 0.639 \\
LTX VAE & 0.599 & 0.591 & 0.543 & 0.609 & 0.515 & 0.611 & 0.576 \\
DINOv3 2D & 0.703 & 0.682 & 0.576 & 0.711 & 0.509 & 0.725 & 0.658 \\
DINOv3 2.5D & 0.704 & 0.681 & 0.566 & 0.709 & 0.503 & 0.723 & 0.654 \\
DALECT-0 & 0.698 & 0.667 & 0.569 & 0.705 & 0.509 & 0.720 & 0.653 \\
DALECT-1s & 0.696 & 0.664 & 0.566 & 0.705 & 0.510 & 0.718 & 0.648 \\
VideoMAE-B & 0.689 & 0.646 & 0.569 & 0.689 & 0.513 & 0.702 & 0.650 \\
DINOv2 2D & 0.705 & 0.678 & 0.577 & 0.709 & 0.512 & 0.722 & 0.659 \\
\bottomrule
\end{longtable}}
{\footnotesize\setlength{\tabcolsep}{3pt}\setlength{\LTleft}{0pt plus 1fil}\setlength{\LTright}{0pt plus 1fil}
\addtocounter{table}{-1}\begin{longtable}{@{\extracolsep{\fill}}lc@{}}
\toprule
\multicolumn{2}{l}{\emph{CT detection-threshold error, largest squared canonical correlation}}\\
\midrule
Text model & median $|$obs$-$pred$|$$\downarrow$ \\
\midrule
CXR-BERT & 0.010 \\
Gemma 4 31B & 0.010 \\
Granite 4.2 8B & 0.010 \\
MedGemma 1.5 4B & 0.009 \\
Ministral 3 14B & 0.010 \\
OLMo 3.1 32B & 0.010 \\
Qwen3-VL-Emb. 8B & 0.008 \\
\bottomrule
\end{longtable}}
{\footnotesize\setlength{\tabcolsep}{3pt}\setlength{\LTleft}{0pt plus 1fil}\setlength{\LTright}{0pt plus 1fil}
\addtocounter{table}{-1}\begin{longtable}{@{\extracolsep{\fill}}llrcccccc@{}}
\toprule
\multicolumn{9}{l}{\emph{Other datasets, FOSCTTM of ridge regression}}\\
\midrule
Dataset & Image encoder & $n$ & MPNet$\downarrow$ & \shortstack{Qwen3.5-0.8B\\(middle layer)}$\downarrow$ & \shortstack{Qwen3.5-27B\\(middle layer)}$\downarrow$ & \shortstack{Qwen3.5-2B\\(middle layer)}$\downarrow$ & \shortstack{Qwen3.5-4B\\(middle layer)}$\downarrow$ & \shortstack{Qwen3.5-9B\\(middle layer)}$\downarrow$ \\
\midrule
COCO & DINOv2 & 1,600 & 0.011{\scriptsize$\pm$0.000} & 0.056{\scriptsize$\pm$0.002} & 0.036{\scriptsize$\pm$0.002} & 0.055{\scriptsize$\pm$0.001} & 0.052{\scriptsize$\pm$0.002} & 0.051{\scriptsize$\pm$0.002} \\
COCO & DINOv2 mean & 1,600 & 0.010{\scriptsize$\pm$0.000} & 0.040{\scriptsize$\pm$0.001} & 0.028{\scriptsize$\pm$0.000} & 0.040{\scriptsize$\pm$0.001} & 0.039{\scriptsize$\pm$0.000} & 0.038{\scriptsize$\pm$0.000} \\
COCO & DINOv3 & 1,600 & 0.010{\scriptsize$\pm$0.000} & 0.054{\scriptsize$\pm$0.004} & 0.037{\scriptsize$\pm$0.002} & 0.053{\scriptsize$\pm$0.004} & 0.052{\scriptsize$\pm$0.004} & 0.051{\scriptsize$\pm$0.004} \\
Flickr30k & DINOv2 & 1,600 & 0.015{\scriptsize$\pm$0.001} & 0.094{\scriptsize$\pm$0.003} & 0.061{\scriptsize$\pm$0.003} & 0.093{\scriptsize$\pm$0.005} & 0.081{\scriptsize$\pm$0.003} & 0.080{\scriptsize$\pm$0.003} \\
Flickr30k & DINOv2 mean & 1,600 & 0.011{\scriptsize$\pm$0.001} & 0.083{\scriptsize$\pm$0.000} & 0.055{\scriptsize$\pm$0.002} & 0.085{\scriptsize$\pm$0.000} & 0.074{\scriptsize$\pm$0.002} & 0.070{\scriptsize$\pm$0.001} \\
Flickr30k & DINOv3 & 1,600 & 0.015{\scriptsize$\pm$0.001} & 0.089{\scriptsize$\pm$0.001} & 0.059{\scriptsize$\pm$0.001} & 0.089{\scriptsize$\pm$0.002} & 0.078{\scriptsize$\pm$0.001} & 0.076{\scriptsize$\pm$0.001} \\
\bottomrule
\end{longtable}}

{\footnotesize\setlength{\tabcolsep}{3pt}\setlength{\LTleft}{0pt plus 1fil}\setlength{\LTright}{0pt plus 1fil}
\begin{longtable}{@{\extracolsep{\fill}}lcccccccc@{}}
\caption{Each entry is the lower end of the 95\,\% bootstrap interval of the report-classifier AUROC minus the prompt AUROC, for one image encoder and one text model (1,600 pairs, ridge regression, mean AUROC over 18 findings on the 3,002 chest volumes of the official validation set, seed mean over the five seeds). The bootstrap resamples patients (1,000 resamples), and the same resamples serve both scoring methods. Every entry is above zero: the smallest is 0.016. The smallest lower end over all four evaluation subsets, both the 18-finding and the 16-finding means, and both pair pools is 0.008.}\label{tab:reader120}\\
\toprule
Image encoder & \shortstack{Qwen3.5-4B} & \shortstack{Gemma 4 31B} & \shortstack{Ministral\\3 14B} & \shortstack{OLMo 3.1 32B} & \shortstack{Granite\\4.2 8B} & \shortstack{MedGemma\\1.5 4B} & \shortstack{Qwen3-VL-Emb.\\8B} & \shortstack{CXR-BERT} \\
\midrule
VoxelFM & 0.081 & 0.109 & 0.295 & 0.058 & 0.222 & 0.080 & 0.143 & 0.084 \\
VoxelFM patch & 0.065 & 0.078 & 0.277 & 0.049 & 0.206 & 0.063 & 0.128 & 0.075 \\
CT-FM & 0.025 & 0.054 & 0.179 & 0.016 & 0.146 & 0.027 & 0.078 & 0.034 \\
V-JEPA 2-L & 0.039 & 0.060 & 0.238 & 0.030 & 0.164 & 0.037 & 0.096 & 0.042 \\
V-JEPA 2-H & 0.043 & 0.060 & 0.244 & 0.033 & 0.171 & 0.044 & 0.098 & 0.049 \\
V-JEPA 2-g & 0.042 & 0.061 & 0.242 & 0.031 & 0.167 & 0.042 & 0.090 & 0.046 \\
\bottomrule
\end{longtable}}

{\footnotesize\setlength{\tabcolsep}{3pt}\setlength{\LTleft}{0pt plus 1fil}\setlength{\LTright}{0pt plus 1fil}
\begin{longtable}{@{\extracolsep{\fill}}lcccccc@{}}
\caption{Each entry of the first two blocks is the mean over five seeds $\pm$ the sample standard deviation for VoxelFM patch, one way of choosing directions with no pairs and one scale (unscaled, or the factor shown). The directions come from TotalSegmentator measurements on the image and zero-shot prompt scores on the report (measurements), from the anchor findings only, or at random. Those blocks are FOSCTTM and prompt AUROC; the last block is FOSCTTM after whitening, independent component analysis (ICA), or skew-ICA, where the run exists. Skew-ICA $K/D$: ICA with $D$ components, keeping the $K$ components with the largest absolute skewness.}\label{tab:supp_pairfree}\\
\toprule
\multicolumn{7}{l}{\emph{FOSCTTM}}\\
\midrule
Directions & unscaled$\downarrow$ & 2$\downarrow$ & 4$\downarrow$ & 8$\downarrow$ & 16$\downarrow$ & 32$\downarrow$ \\
\midrule
measurements & 0.504{\scriptsize$\pm$0.014} & 0.507{\scriptsize$\pm$0.020} & 0.511{\scriptsize$\pm$0.014} & 0.559{\scriptsize$\pm$0.078} & 0.607{\scriptsize$\pm$0.016} & 0.512{\scriptsize$\pm$0.123} \\
anchor findings & 0.504{\scriptsize$\pm$0.014} & 0.506{\scriptsize$\pm$0.018} & 0.495{\scriptsize$\pm$0.029} & 0.533{\scriptsize$\pm$0.083} & 0.554{\scriptsize$\pm$0.082} & 0.543{\scriptsize$\pm$0.078} \\
random directions & 0.504{\scriptsize$\pm$0.014} & 0.504{\scriptsize$\pm$0.007} & 0.506{\scriptsize$\pm$0.006} & 0.501{\scriptsize$\pm$0.013} & 0.500{\scriptsize$\pm$0.003} & 0.497{\scriptsize$\pm$0.017} \\
\midrule
\multicolumn{7}{l}{\emph{Prompt AUROC}}\\
\midrule
Directions & unscaled$\uparrow$ & 2$\uparrow$ & 4$\uparrow$ & 8$\uparrow$ & 16$\uparrow$ & 32$\uparrow$ \\
\bottomrule
\end{longtable}}
{\footnotesize\setlength{\tabcolsep}{3pt}\setlength{\LTleft}{0pt plus 1fil}\setlength{\LTright}{0pt plus 1fil}
\addtocounter{table}{-1}\begin{longtable}{@{\extracolsep{\fill}}cccccc@{}}
\toprule
\multicolumn{6}{l}{\emph{Whitening, ICA, and skew-ICA}}\\
\midrule
ICA 16$\downarrow$ & ICA 64$\downarrow$ & \shortstack{skew-ICA\\16/128}$\downarrow$ & whitening 16$\downarrow$ & whitening 64$\downarrow$ & \shortstack{skew-ICA\\10/64}$\downarrow$ \\
\midrule
\multicolumn{6}{l}{\emph{COCO, DINOv2 mean, MPNet}} \\
0.318{\scriptsize$\pm$0.105} & 0.503{\scriptsize$\pm$0.009} & 0.485{\scriptsize$\pm$0.015} & 0.370{\scriptsize$\pm$0.075} & 0.474{\scriptsize$\pm$0.028} & -- \\
\multicolumn{6}{l}{\emph{CT-RATE 3D CT, DINOv2 2D, Qwen3.5-4B (middle layer)}} \\
0.476{\scriptsize$\pm$0.023} & 0.499{\scriptsize$\pm$0.009} & 0.499{\scriptsize$\pm$0.002} & 0.490{\scriptsize$\pm$0.007} & 0.491{\scriptsize$\pm$0.009} & 0.489{\scriptsize$\pm$0.009} \\
\multicolumn{6}{l}{\emph{CT-RATE 3D CT, VoxelFM block 11, Qwen3.5-9B (middle layer)}} \\
0.460{\scriptsize$\pm$0.017} & 0.496{\scriptsize$\pm$0.006} & 0.495{\scriptsize$\pm$0.005} & 0.453{\scriptsize$\pm$0.004} & 0.501{\scriptsize$\pm$0.012} & 0.500{\scriptsize$\pm$0.007} \\
\multicolumn{6}{l}{\emph{CT-RATE 3D CT, VoxelFM patch, Qwen3.5-4B (middle layer)}} \\
0.477{\scriptsize$\pm$0.011} & 0.485{\scriptsize$\pm$0.004} & 0.502{\scriptsize$\pm$0.012} & 0.440{\scriptsize$\pm$0.015} & 0.494{\scriptsize$\pm$0.005} & 0.498{\scriptsize$\pm$0.002} \\
\bottomrule
\end{longtable}}

{\scriptsize\setlength{\tabcolsep}{3pt}\setlength{\LTleft}{0pt plus 1fil}\setlength{\LTright}{0pt plus 1fil}
\begin{longtable}{@{\extracolsep{\fill}}lccccc@{}}
\caption{Each row is one image encoder on the BIMCV-R CT dataset (Qwen3.5-4B, middle and last layer, $d\in\{16,32\}$, $n\in\{100,150\}$ where $2d<n$). Settings is the number of such settings. $|$obs$-$pred$|$ is the median absolute error of the top three squared canonical correlations between the detection threshold and the BIMCV-R sample. Predicted $k$ is the number of directions the threshold predicts as detected; ``within 1'' counts settings where the observed mean count of detected directions is within one of it. The last two columns take the count $k$ predicted from the CT-RATE fit of the same encoder, text layer, $d$, and $n$ (the CT-RATE grid has $n=100$, not 150), and count the BIMCV-R settings where it equals the predicted BIMCV-R count or is within one of the observed one. On CT-RATE the same threshold has median error 0.016 over 2,550 settings and $k$ within 1 in 0.713 of settings.}\label{tab:bimcv_transfer}\\
\toprule
Image encoder & settings & \shortstack{median\\$|$obs$-$pred$|$$\downarrow$} & \shortstack{BIMCV-R $k$\\within 1} & \shortstack{CT $k$ =\\BIMCV-R $k$} & \shortstack{CT $k$ within\\1 of obs.} \\
\midrule
VoxelFM & 8 & 0.022 & 1/8 & 0/4 & 4/4 \\
VoxelFM patch & 8 & 0.016 & 1/8 & 0/4 & 4/4 \\
CT-FM & 8 & 0.032 & 7/8 & 0/4 & 4/4 \\
V-JEPA 2-L & 8 & 0.032 & 6/8 & 2/4 & 4/4 \\
V-JEPA 2-H & 8 & 0.021 & 5/8 & 2/4 & 4/4 \\
V-JEPA 2-g & 8 & 0.024 & 4/8 & 2/4 & 4/4 \\
Wan VAE & 8 & 0.028 & 8/8 & 2/4 & 4/4 \\
Wan DiT & 8 & 0.020 & 8/8 & 2/4 & 4/4 \\
LTX VAE & 8 & 0.027 & 8/8 & 3/4 & 4/4 \\
DINOv3 2D & 8 & 0.017 & 5/8 & 2/4 & 4/4 \\
DINOv3 2.5D & 8 & 0.019 & 3/8 & 2/4 & 4/4 \\
DALECT-0 & 8 & 0.023 & 3/8 & 1/4 & 4/4 \\
DALECT-1s & 8 & 0.018 & 4/8 & 2/4 & 4/4 \\
VideoMAE-B & 8 & 0.029 & 8/8 & 2/4 & 4/4 \\
DINOv2 2D & 8 & 0.023 & 5/8 & 2/4 & 4/4 \\
\midrule
All 15 encoders & 120 & 0.024 & 76/120 & 24/60 & 60/60 \\
\bottomrule
\end{longtable}}

{\scriptsize\setlength{\tabcolsep}{3pt}\setlength{\LTleft}{0pt plus 1fil}\setlength{\LTright}{0pt plus 1fil}
\begin{longtable}{@{\extracolsep{\fill}}lccc@{}}
\caption{Each row is one language-model layer or one VoxelFM block. Settings is the number of settings (one image encoder, one $d$, one pair count $n$). The two error columns are the median and the maximum, over the settings of the row and the top three squared canonical correlations, of the absolute difference between the observed sample value and the value of the detection threshold. Sizes and layers (upper block) use $d\in\{16,32,64,128\}$ and $n$ from 100 to 1,600; the layer block uses $d=64$ and $n\in\{200,400,800,1600\}$; the block rows use $n\in\{200,400,800,1600\}$.}\label{tab:law_depth}\\
\toprule
Layer or block & settings & \shortstack{median\\$|$obs$-$pred$|$$\downarrow$} & \shortstack{max\\$|$obs$-$pred$|$$\downarrow$} \\
\midrule
\multicolumn{4}{l}{\emph{Qwen3.5 size and layer, 15 encoders}}\\
\midrule
Qwen3.5-0.8B, middle layer & 255 & 0.015 & 0.146 \\
Qwen3.5-0.8B, last layer & 255 & 0.013 & 0.153 \\
Qwen3.5-2B, middle layer & 255 & 0.015 & 0.140 \\
Qwen3.5-2B, last layer & 255 & 0.015 & 0.160 \\
Qwen3.5-4B, middle layer & 255 & 0.015 & 0.142 \\
\bottomrule
\end{longtable}}

{\footnotesize\setlength{\tabcolsep}{3pt}\setlength{\LTleft}{0pt plus 1fil}\setlength{\LTright}{0pt plus 1fil}
\begin{longtable}{@{\extracolsep{\fill}}lllrrrrrrr@{}}
\caption{Each row is one encoder--text combination on one dataset, with no pairs. It gives the mutual $k$-nearest-neighbor rate (m$k$NN) or linear CKA (centered kernel alignment), the same quantity after calibration against a permutation null, and the null $p$-value. The null shuffles the text rows (200 draws), and $n$ is the number of test items.}\label{tab:supp_loc}\\
\toprule
Dataset & Image encoder & Text model & $n$ & m$k$NN$\uparrow$ & calibrated$\uparrow$ & $p$ & CKA$\uparrow$ & calibrated$\uparrow$ & $p$ \\
\midrule
COCO & DINOv2 & MPNet & 2,020 & 0.257 & 0.253 & 0.005 & 0.445 & 0.445 & 0.005 \\
COCO & DINOv2 & Qwen3.5-4B (middle layer) & 2,020 & 0.097 & 0.092 & 0.005 & 0.212 & 0.212 & 0.005 \\
COCO & DINOv2 mean & MPNet & 2,020 & 0.224 & 0.220 & 0.005 & 0.567 & 0.567 & 0.005 \\
COCO & DINOv2 mean & Qwen3.5-4B (middle layer) & 2,020 & 0.094 & 0.090 & 0.005 & 0.323 & 0.323 & 0.005 \\
COCO & DINOv3 & MPNet & 2,020 & 0.258 & 0.254 & 0.005 & 0.434 & 0.434 & 0.005 \\
COCO & DINOv3 & Qwen3.5-4B (middle layer) & 2,020 & 0.096 & 0.092 & 0.005 & 0.211 & 0.211 & 0.005 \\
\bottomrule
\end{longtable}}
{\footnotesize\setlength{\tabcolsep}{3pt}\setlength{\LTleft}{0pt plus 1fil}\setlength{\LTright}{0pt plus 1fil}
\addtocounter{table}{-1}\begin{longtable}{@{\extracolsep{\fill}}lllrrrrrrr@{}}
\toprule
Dataset & Image encoder & Text model & $n$ & m$k$NN$\uparrow$ & calibrated$\uparrow$ & $p$ & CKA$\uparrow$ & calibrated$\uparrow$ & $p$ \\
\midrule
CT-RATE 3D CT & DINOv3 2.5D & Qwen3.5-4B (middle layer) & 2,025 & 0.011 & 0.007 & 0.005 & 0.051 & 0.051 & 0.005 \\
CT-RATE 3D CT & DINOv3 2D & MPNet & 2,025 & 0.011 & 0.006 & 0.005 & 0.036 & 0.036 & 0.005 \\
CT-RATE 3D CT & DINOv3 2D & Qwen3.5-4B (middle layer) & 2,025 & 0.011 & 0.006 & 0.005 & 0.065 & 0.065 & 0.005 \\
CT-RATE 3D CT & LTX VAE & MPNet & 2,025 & 0.007 & 0.002 & 0.005 & 0.014 & 0.014 & 0.005 \\
CT-RATE 3D CT & LTX VAE & Qwen3.5-4B (middle layer) & 2,025 & 0.008 & 0.003 & 0.005 & 0.015 & 0.015 & 0.005 \\
CT-RATE 3D CT & VideoMAE-B & MPNet & 2,025 & 0.010 & 0.005 & 0.005 & 0.022 & 0.022 & 0.005 \\
CT-RATE 3D CT & VideoMAE-B & Qwen3.5-4B (middle layer) & 2,025 & 0.010 & 0.005 & 0.005 & 0.040 & 0.040 & 0.005 \\
\bottomrule
\end{longtable}}

{\footnotesize\setlength{\tabcolsep}{3pt}\setlength{\LTleft}{0pt plus 1fil}\setlength{\LTright}{0pt plus 1fil}
\begin{longtable}{@{\extracolsep{\fill}}lcccc@{}}
\caption{Each entry is the mean over five seeds $\pm$ the sample standard deviation at one layer of the language model Qwen3.5-4B and one image encoder, for $n=100$ and $n=1{,}600$ pairs, with the zero-shot prompt.}\label{tab:supp_layers}\\
\toprule
Image encoder, layer & \multicolumn{2}{c}{prompt AUROC} & \multicolumn{2}{c}{FOSCTTM} \\
\cmidrule(lr){2-3}\cmidrule(lr){4-5}
 & 100$\uparrow$ & 1,600$\uparrow$ & 100$\downarrow$ & 1,600$\downarrow$ \\
\midrule
VoxelFM, layer 0 & 0.665{\scriptsize$\pm$0.010} & 0.685{\scriptsize$\pm$0.002} & 0.331{\scriptsize$\pm$0.012} & 0.243{\scriptsize$\pm$0.002} \\
VoxelFM, layer 4 & 0.667{\scriptsize$\pm$0.012} & 0.699{\scriptsize$\pm$0.002} & 0.300{\scriptsize$\pm$0.012} & 0.223{\scriptsize$\pm$0.001} \\
VoxelFM, layer 8 & 0.696{\scriptsize$\pm$0.009} & 0.712{\scriptsize$\pm$0.001} & 0.289{\scriptsize$\pm$0.007} & 0.215{\scriptsize$\pm$0.003} \\
VoxelFM, layer 12 & 0.698{\scriptsize$\pm$0.011} & 0.716{\scriptsize$\pm$0.001} & 0.286{\scriptsize$\pm$0.005} & 0.219{\scriptsize$\pm$0.002} \\
VoxelFM, layer 16 & 0.707{\scriptsize$\pm$0.009} & 0.720{\scriptsize$\pm$0.001} & 0.297{\scriptsize$\pm$0.007} & 0.237{\scriptsize$\pm$0.002} \\
\bottomrule
\end{longtable}}
{\footnotesize\setlength{\tabcolsep}{3pt}\setlength{\LTleft}{0pt plus 1fil}\setlength{\LTright}{0pt plus 1fil}
\addtocounter{table}{-1}\begin{longtable}{@{\extracolsep{\fill}}lcccc@{}}
\toprule
Image encoder, layer & \multicolumn{2}{c}{prompt AUROC} & \multicolumn{2}{c}{FOSCTTM} \\
\cmidrule(lr){2-3}\cmidrule(lr){4-5}
 & 100$\uparrow$ & 1,600$\uparrow$ & 100$\downarrow$ & 1,600$\downarrow$ \\
\midrule
LTX VAE, layer 0 & 0.523{\scriptsize$\pm$0.020} & 0.581{\scriptsize$\pm$0.002} & 0.462{\scriptsize$\pm$0.006} & 0.375{\scriptsize$\pm$0.004} \\
LTX VAE, layer 4 & 0.530{\scriptsize$\pm$0.010} & 0.595{\scriptsize$\pm$0.005} & 0.454{\scriptsize$\pm$0.007} & 0.373{\scriptsize$\pm$0.003} \\
LTX VAE, layer 8 & 0.545{\scriptsize$\pm$0.005} & 0.600{\scriptsize$\pm$0.006} & 0.453{\scriptsize$\pm$0.006} & 0.370{\scriptsize$\pm$0.003} \\
LTX VAE, layer 12 & 0.545{\scriptsize$\pm$0.003} & 0.607{\scriptsize$\pm$0.009} & 0.451{\scriptsize$\pm$0.004} & 0.369{\scriptsize$\pm$0.004} \\
LTX VAE, layer 16 & 0.548{\scriptsize$\pm$0.004} & 0.616{\scriptsize$\pm$0.009} & 0.454{\scriptsize$\pm$0.002} & 0.375{\scriptsize$\pm$0.005} \\
LTX VAE, layer 20 & 0.545{\scriptsize$\pm$0.004} & 0.613{\scriptsize$\pm$0.008} & 0.461{\scriptsize$\pm$0.005} & 0.378{\scriptsize$\pm$0.005} \\
\bottomrule
\end{longtable}}

{\scriptsize\setlength{\tabcolsep}{3pt}\setlength{\LTleft}{0pt plus 1fil}\setlength{\LTright}{0pt plus 1fil}
\begin{longtable}{@{\extracolsep{\fill}}llllcccc@{}}
\caption{Each row is one seed: FOSCTTM after enumerating the sign patterns of the top 10 directions, after the greedy search over 40 directions, after the closed-form rule from the skewness of the two sides (40 directions), and after the best of the enumerated sign patterns chosen with test pairs (needs pairs). CT rows: a second run of the aligner with the same seeds (Section~\ref{app:impl}).}\label{tab:supp_sign_methods}\\
\toprule
Dataset & Image encoder & Text model & seed & enumerate 10$\downarrow$ & greedy 40$\downarrow$ & skew 40$\downarrow$ & \shortstack{best pattern\\(needs pairs)}$\downarrow$ \\
\midrule
Flickr30k & DINOv2 & MPNet & 2197670066 & 0.179 & 0.180 & 0.183 & 0.169 \\
Flickr30k & DINOv2 & MPNet & 576165995 & 0.123 & 0.129 & 0.123 & 0.123 \\
Flickr30k & DINOv2 & MPNet & 734796314 & 0.152 & 0.161 & 0.149 & 0.150 \\
Flickr30k & DINOv2 mean & MPNet & 2197670066 & 0.099 & 0.102 & 0.099 & 0.095 \\
Flickr30k & DINOv2 mean & MPNet & 576165995 & 0.091 & 0.098 & 0.091 & 0.091 \\
Flickr30k & DINOv2 mean & MPNet & 734796314 & 0.028 & 0.032 & 0.028 & 0.028 \\
\bottomrule
\end{longtable}}
{\scriptsize\setlength{\tabcolsep}{3pt}\setlength{\LTleft}{0pt plus 1fil}\setlength{\LTright}{0pt plus 1fil}
\addtocounter{table}{-1}\begin{longtable}{@{\extracolsep{\fill}}llllcccc@{}}
\toprule
Dataset & Image encoder & Text model & seed & enumerate 10$\downarrow$ & greedy 40$\downarrow$ & skew 40$\downarrow$ & \shortstack{best pattern\\(needs pairs)}$\downarrow$ \\
\midrule
Clotho audio & AST & MPNet & 576165995 & 0.066 & 0.066 & 0.066 & 0.066 \\
Clotho audio & AST & MPNet & 734796314 & 0.066 & 0.066 & 0.066 & 0.066 \\
Clotho audio & AST & Qwen3.5-4B (middle layer) & 2197670066 & 0.306 & 0.301 & 0.303 & 0.292 \\
Clotho audio & AST & Qwen3.5-4B (middle layer) & 576165995 & 0.097 & 0.101 & 0.097 & 0.097 \\
Clotho audio & AST & Qwen3.5-4B (middle layer) & 734796314 & 0.148 & 0.154 & 0.148 & 0.148 \\
Clotho audio & Dasheng & MPNet & 2197670066 & 0.255 & 0.268 & 0.460 & 0.217 \\
Clotho audio & Dasheng & MPNet & 576165995 & 0.147 & 0.148 & 0.147 & 0.139 \\
\bottomrule
\end{longtable}}

\clearpage
\begin{table}[tp]
\centering
\caption{Image classifier without a map against \textit{LightAlign} on CT-RATE. Entries: mean AUROC over 18 findings on the test set (3,002 scans), averaged over five seeds (5,994 pairs: one fit). Brackets: 95\,\% paired bootstrap interval of the difference from CT-CLIP, which reaches 0.733. Untuned: logistic regression with $C=1$ on the $n$ paired images. Tuned: $C$ chosen for each finding by cross-validation on the $n$ pairs. Frozen VoxelFM block 11 features; \textit{LightAlign} uses Qwen3.5-9B (middle layer) and the report classifier. --: not part of this comparison.}
\label{tab:supp_probe_cv}
\setlength{\tabcolsep}{3pt}
\footnotesize
\begin{tabular}{@{}lccc@{}}
\toprule
Pairs $n$ & Untuned image classifier & Tuned image classifier & \textit{LightAlign} \\
\midrule
100 & 0.689{\scriptsize[-0.051,\,-0.036]} & 0.698{\scriptsize[-0.043,\,-0.028]} & 0.727{\scriptsize[-0.013,\,0.002]} \\
200 & -- & 0.735{\scriptsize[-0.006,\,0.008]} & 0.754{\scriptsize[0.014,\,0.028]} \\
400 & 0.713{\scriptsize[-0.028,\,-0.012]} & 0.767{\scriptsize[0.027,\,0.041]} & 0.782{\scriptsize[0.041,\,0.056]} \\
1,600 & 0.712{\scriptsize[-0.029,\,-0.012]} & 0.806{\scriptsize[0.065,\,0.080]} & 0.809{\scriptsize[0.068,\,0.084]} \\
5,994 & -- & 0.824{\scriptsize[0.083,\,0.099]} & 0.824{\scriptsize[0.083,\,0.100]} \\
\bottomrule
\end{tabular}
\end{table}

\begin{table*}[tp]
\centering
\caption{\textit{LightAlign} on CT-RATE with and without the choice of the VoxelFM block and the language-model layer. Entries: mean AUROC over 18 findings on the test set (3,002 scans), averaged over five seeds (5,994 pairs: one fit), with ridge regression on $n$ image--report pairs and the report classifier; the text model is Qwen3.5-9B. Brackets: 95\,\% paired bootstrap interval of the difference from CT-CLIP, which reaches 0.733. Block 12 (last) is the final output of VoxelFM. Shaded: the final configuration, chosen on the validation patients. The last row uses no block choice and no layer choice.}
\label{tab:supp_no_selection}
\setlength{\tabcolsep}{3pt}
\footnotesize
\begin{tabular*}{\linewidth}{@{\extracolsep{\fill}}llcccc@{}}
\toprule
Image encoder & Qwen3.5-9B layer & 100$\uparrow$ & 200$\uparrow$ & 1,600$\uparrow$ & 5,994$\uparrow$ \\
\midrule
\rowcolor{oursbg} VoxelFM block 11 & middle & 0.727{\scriptsize[-0.013,\,0.002]} & 0.754{\scriptsize[0.014,\,0.028]} & 0.809{\scriptsize[0.068,\,0.084]} & 0.824{\scriptsize[0.083,\,0.100]} \\
VoxelFM block 12 (last) & middle & 0.728{\scriptsize[-0.012,\,0.002]} & 0.753{\scriptsize[0.013,\,0.027]} & 0.807{\scriptsize[0.066,\,0.082]} & 0.824{\scriptsize[0.083,\,0.099]} \\
VoxelFM block 11 & last & 0.723{\scriptsize[-0.017,\,-0.003]} & 0.748{\scriptsize[0.008,\,0.022]} & 0.808{\scriptsize[0.067,\,0.082]} & 0.825{\scriptsize[0.084,\,0.100]} \\
VoxelFM block 12 (last) & last & 0.725{\scriptsize[-0.015,\,-0.001]} & 0.748{\scriptsize[0.008,\,0.021]} & 0.806{\scriptsize[0.066,\,0.081]} & 0.824{\scriptsize[0.084,\,0.099]} \\
\bottomrule
\end{tabular*}
\end{table*}

\begin{table}[tp]
\centering
\caption{Mean AUROC on CT-RATE (3,002 chest scans) after removing the axis common to all findings. For each finding, we regress its scores on the mean score across the 18 findings and retain the residuals. Raw: AUROC of the scores; Resid.: AUROC of the residuals. All scores are logits: for CT-CLIP, we use the logit of its two-prompt probability. CT-CLIP reaches 0.733 raw and 0.595 after removal, and its scores have an effective rank of 1.63. Brackets: 95\,\% paired bootstrap intervals for the residual AUROC difference of \textit{LightAlign} relative to CT-CLIP and to the tuned image classifier in Table~\ref{tab:supp_probe_cv}. Eff.\ rank: effective rank of the score matrix after centring each finding over the scans (Section~\ref{app:payoff}), median over seeds. Frozen VoxelFM block 11 features; \textit{LightAlign} uses Qwen3.5-9B (middle layer) and the report classifier. Five seeds (5,994 pairs: one fit).}
\label{tab:supp_resid}
\setlength{\tabcolsep}{2.5pt}
\footnotesize
\begin{tabular*}{\linewidth}{@{\extracolsep{\fill}}lcccccccc@{}}
\toprule
 & \multicolumn{2}{c}{\textit{LightAlign}} & \multicolumn{2}{c}{Tuned image classifier} & \multicolumn{2}{c}{Resid.\ difference of \textit{LightAlign}} & \multicolumn{2}{c}{Eff.\ rank} \\
\cmidrule(lr){2-3}\cmidrule(lr){4-5}\cmidrule(lr){6-7}\cmidrule(l){8-9}
Pairs $n$ & Raw & Resid. & Raw & Resid. & from CT-CLIP & from image classifier & \textit{LightAlign} & Image cl. \\
\midrule
100 & 0.727 & 0.613 & 0.698 & 0.619 & [0.008,\,0.029] & [-0.011,\,-0.001] & 4.93 & 5.16 \\
200 & 0.754 & 0.641 & 0.735 & 0.640 & [0.036,\,0.058] & [-0.004,\,0.006] & 4.26 & 3.71 \\
400 & 0.782 & 0.676 & 0.767 & 0.663 & [0.071,\,0.093] & [0.008,\,0.019] & 4.40 & 4.31 \\
1,600 & 0.809 & 0.712 & 0.806 & 0.696 & [0.106,\,0.130] & [0.011,\,0.020] & 4.73 & 4.50 \\
5,994 & 0.824 & 0.732 & 0.824 & 0.722 & [0.126,\,0.150] & [0.006,\,0.014] & 4.94 & 4.43 \\
\bottomrule
\end{tabular*}
\end{table}

\end{document}